# Automated Copper Alloy Grain Size Evaluation Using a Deep-learning CNN

George S. Baggs                        gbaggs@moog.com

Paul Guerrier                          pguerrier@moog.com

Andrew Loeb                            aloeb@moog.com

Jason C. Jones                         jjones8@moog.com

## Abstract

Moog Inc. has automated the evaluation of copper (Cu) alloy grain size using a deep-learning convolutional neural network (CNN). The proof-of-concept automated image acquisition and batch-wise image processing offers the potential for significantly reduced labor, improved accuracy of grain evaluation, and decreased overall turnaround times for approving Cu alloy bar stock for use in flight critical aircraft hardware. A classification accuracy of 91.1% on individual sub-images of the Cu alloy coupons was achieved. Process development included minimizing the variation in acquired image color, brightness, and resolution to create a dataset with 12300 sub-images, and then optimizing the CNN hyperparameters on this dataset using statistical design of experiments (DoE).

Over the development of the automated Cu alloy grain size evaluation, a degree of "explainability" in the artificial intelligence (XAI) output was realized, based on the decomposition of the large raw images into many smaller dataset sub-images, through the ability to explain the CNN ensemble image output via inspection of the classification results from the individual smaller sub-images.

## 1. Introduction

The evaluation of copper (Cu) alloy bar-stock has historically been executed by trained metallurgical technicians. Removed bar stock ends are delivered to the Moog Inc. Global Materials & Process Engineering (M&PE) department in East Aurora NY for evaluation. The month-to-month bar-stock batch sizes were highly variable and typically ranged between 10 and 150. The batch size variation, when combined with differences between technician evaluations, resulted in inspection uncertainty, unpredictable process cycle times and an evaluation process that was both tedious and monotonous.

In response to the above, automated analytical instrumentation combined with advanced image-recognition techniques were evaluated for implementation; additionally, the supply chain was engaged to level the monthly demand, and now bar-stock batch sizes range between 40 and 45.

The proposed deep learning convolutional neural network (CNN) was developed as a proof-of-concept demonstration that will be evaluated against the improved inspection process described above. An automated inspection process using the CNN has the potential to:

- Further reduce inspection variations to negligible levels
- Improve the quality of the sample evaluations beyond the current level
- Reduce process cycle time by eliminating human bias and errors in evaluation

# Automated Copper Alloy Grain Size Evaluation Using a Deep-learning CNN

- Free skilled employees to focus on other value-added tasks that still require human cognizant decision making.

An overview of the development process that was used for the creation of the trained Cu-alloy CNN described in this writing is depicted in Figure 1. The numbers in the boxes refer to the applicable sections of this paper.

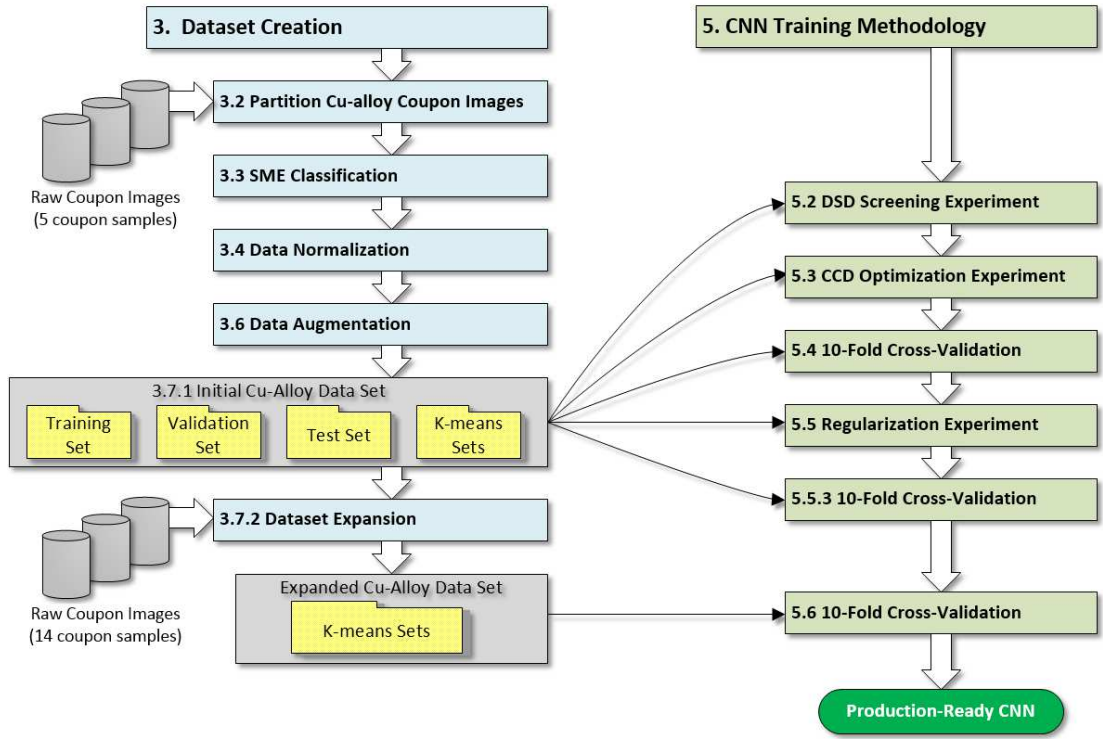

Figure 1 – Development Overview of Cu-Alloy Deep Learning CNN

The dataset creation is described in Section 3, which involved physical sample preparation, high-resolution imaging of the physical samples, image pre-processing, sample labeling and dataset normalization. The CNN model architecture is described in Section 4. The CNN training methodology is provided in Section 5, and includes CNN hyperparameter screening and optimization experiments; a CNN weight-regularization experiment is also described. Ten-fold cross-validation is used to verify CNN performance after the optimization and weight-regularization experiments. Another 10-fold cross-validation on an expanded dataset is used to baseline the final performance of the CNN. A summary of experimental results is provided in Section 6 followed by final conclusions in Section 7. Additional information about the hardware used for physical sample preparation, the software applications employed, computing hardware, details about the funding for this effort, as well as a glossary of terms are provided in Section 8.

The primary contribution of this work is to demonstrate of the use of statistical Design-of-Experiments (DoE) [1] and other forms of applied industrial statistics to set CNN hyperparameters to achieve the highest possible classification accuracy with the lowest run-to-run training variation. Statistical testing is used as the basis for either accepting the null hypothesis ($H_0: \mu_1 = \mu_2$) that CNN performance has not changed, or the alternate hypothesis ($H_A: \mu_1 \neq \mu_2$) that CNN performance has changed.

---

[1] DoE and other applied statistical methods, first developed in the United Kingdom in the early 1920s for agricultural use, have been used routinely to maximize the efficiency of industrial product and process design since the mid-20th Century for the chemical and process industries, and then spreading to many other industries between the 1970s and 1990s [33].

# 2. Literature Review

This literature review is approached in three stages: first, the general application of machine learning and specific use of CNNs for inspection and/or classification of metals for process and quality control are reviewed; second, the best practices and research into the configuration and training of CNNs are reviewed; and third, the background of applied industrial statistics, especially the application of DoE for screening and optimization, are reviewed.

## 2.1. Previous Applications

Improving the efficiency and accuracy of visually-oriented industrial inspection and quality control methodologies has always been a goal for industrial process development. Advances in machine vision technology coupled with machine-learning techniques have proven useful in moving industrial processes towards the aforementioned goal [1]. In recent years, a class of deep learning algorithms in the form of CNNs have made significant progress in enabling the recognition of spatial patterns in image data, with classification accuracies that equal or exceed those produced by human subject matter experts (SME) [2].

A CNN was taught to recognize good metal additive manufacturing (AM) output by training it with labeled images of top surface printed CoCrMo (Cobalt Chromium Molybdenum) alloy output that showed even hatch lines and appropriate overlaps to indicate good quality welding [2]. The CoCrMo alloy AM dataset provided by the aforementioned CNN effort was used to demonstrate that semi-supervised training may offer the potential to provide promising classification accuracy performance, using a more limited labeled dataset along with lower quality noisy and blurry images [3]. To improve the accuracy and speed of the industrial inspection for metal gears, a CNN was used to classify defects, with a tradeoff demonstrated between processing time with direct classification of images, and accuracy, when a 'fine-grained' image preprocessing was used prior to classification [4]. A CNN was adapted for the automated visual inspection of surface defects of countersunk drilled holes in steel, which required considerable image preprocessing and data augmentation [5]. CNNs were applied to the detection of micro-sized defects on the surfaces of metal screws, with image preprocessing to extract the screw images for the CNN [6].

All the above sources provide deterministic performance metrics for the accuracies achieved by the CNNs.

## 2.2. CNN Best Practices

Prior research and academic sources have established theoretical and empirically-based best practices for setting the hyperparameters of CNNs. The primary objectives of these best practices are to reducing overfitting and improve classification accuracy.

During CNN training, the *batch* size represents the number of training examples used in one forward and backward pass, which is often referred to as the *mini-batch* when the *batch* size is a fraction of the total training set size. "*Small mini-batch sizes provide more up-to-date gradient calculations, which yields more stable and reliable training*" [7], while larger mini-batch sizes require more memory space. Excessive mini-batch size may also make a CNN tend "*to converge to sharp minimizers of the training and testing functions…sharp minima lead to poorer generalization*" [8].

## Automated Copper Alloy Grain Size Evaluation Using a Deep-learning CNN

The *Optimizer* function is used for the gradient descent while training, which occurs during back propagation. Mini-batch gradient descent was used for our application, which combines the advantages of both batch-gradient descent[2] and stochastic gradient descent.[3] We explored the use of three different optimizers: *Adam* [9], *AdaMax* [9] and *Nadam* [10] because each provide different levels of adaptive moment [11].

The *Kernel Constraint* is a method to prevent overfitting and improve generalization by limiting the magnitude of neuron weights in the fully-connected dense classification layers during training to the range of 3 to 4 [12]. *Dropout* is a regularization method to prevent CNN overfitting during training and improve generalization and works by stochastically dropping the specified percentage of neurons for the associated layer for each training mini-batch [12]. This counters the tendency for co-adaptation of CNN layers [12] by essentially presenting a different network architecture (i.e. a different model) for each training iteration. Srivastava, et al. demonstrated that dropout reduced classification error for the MNIST dataset, and that dropout values around 20% for input layers and 50% for hidden layers were optimal. Combining both kernel constraint and dropouts "*provides a significant boost over just using dropout*" [12].

*Pooling* layers were used to down-sample the representations of input features created by the convolutional layers, which reduces the sensitivity of the CNN to changes in the position of features (i.e. improves translation invariance). *Max-Pooling* was used instead of average pooling because it is "*more informative to look at the maximal presence of different features than at their average presence*" [13].

The filter kernels are used in the convolutional-operation[4] in the convolutional layers of the CNN [14]. Smaller filter kernels reduce the receptor field to capture highly-localized image features, while filter kernels with larger receptor fields capture less localized more generalized image features [13]. Larger filter kernels also require more computational processing.

*Stride* is a form of down-sampling (or information reduction) when stride length $> 1$, and is applied to the filter kernel convolution in our CNN.

Three different *Activation* functions *TanH, SELU* and *ReLU* were selected to provide different levels of resistance to excessively small, or 'vanishing' gradients during back-propagation while training the CNN. "*One of many difficulties is that the norm of the back-propagated error gradient can grow or decay exponentially*" [15]. On one extreme, *TanH* or Hyperbolic Tangent, is a logistic function with an output that ranges between -1.0 and +1.0; however, the output can saturate with large inputs and only has sensitivity around 0…the mid-point of the input [13]. On the other extreme, the *ReLU* or Rectified Linear Unit activation function provides a linear positive output for positive inputs and no output for any negative input. "*Because rectified linear units are nearly linear, they preserve many of the properties that make linear models easy to optimize with gradient-based methods. They also preserve many of the properties that make linear models generalize well*" [13]. The *SELU* or Scaled Exponential Linear Unit is a self-normalizing function, as "*activations close to zero mean and unit variance that are propagated through many network layers will converge towards zero mean and unit variance—even under the presence of noise and perturbations*" [16]. This property permits robust learning within deep networks, and makes vanishing gradients impossible.

---

[2] The batch size = total training set size, which consumes computer memory and takes longer to compute, but provides a stable non-noisy gradient descent trajectory [11].

[3] When 'batch size' = 1, only one update at a time is performed, which reduces computer memory requirements and dramatically speeds up computations, but results in a very unstable and noisy (i.e. stochastic) gradient descent trajectory [11].

[4] The term 'convolution' is used to describe the operation of CNNs, but the mathematical process actually being used is 'cross-correlation' because there is no inversion of the filter kernel performed prior to its application to the input [13].

## Automated Copper Alloy Grain Size Evaluation Using a Deep-learning CNN

We explored weight regularization as a way to reduce the variance observed in the 10-fold cross-validation performed as described in Section 5.4.1. Larger weights can make a CNN less stable, because minor variations or noise in the inputs can produce larger changes at the outputs [17]. Weight regularization adds a vector normalization penalty to the training loss function to encourage the formation of smaller weights.

## 2.3. Applied Industrial Statistics

### _Screening Experiments_

Industrial optimization projects often begin with the screening experiment, which is designed to identify non-random effects of $n$ responses within a stochastic environment across a wide range of $m$ variables. Because of information limitations within the configuration of experimental matrices, screening experiments with large numbers of factors were traditionally Resolution III designs (main-effects aliased with two-factor interactions) or at best Resolution IV designs (two-factor interactions aliased with other two-factor interactions) and additional experimentation $n \gg m$ would be required to separate the confounded effects; furthermore, traditional screening designs only have the capability to study first-order effects [18].

The _Definitive Screening Design_ (DSD) is a new class of recently developed three-level screening experiments, which can _"provide estimates of main effects that are unbiased by any second-order effect, require only one more than twice as many runs as there are factors, and avoid confounding of any pair of second-order effects"_ [18], where $n = 2m + 1$, and $y_i, \dots, y_n$ is the response vector of the normal linear regression model, _"where the parameters $\beta_o, \dots, \beta_{mm}$ are the vector of unknown regression coefficients, where $x_{ij}$ denotes the jkth two-factor interaction column, the ith entry of which is $x_{i,j}x_{i,k}$, while $x_{jj}$ denotes the jth pure-quadratic effect column, the ith entry of which is $x_{i,j}^2$"_ [18], and $\{\varepsilon_i\}$ is the unobservable vector of error terms $iid \sim N(\mu_i, \sigma^2)$ where $\mu_i = 0$, and $n > m$. The Equation (1) defines the normal linear regression response vector $y_i$ [18]

$$y_i = \beta_o + \sum_{j=1}^{m} \beta_j x_{i,j} + \sum_{j=1}^{m-1} \sum_{k=j+1}^{m} \beta_{jk} x_{i,j} x_{i,k} + \sum_{j=1}^{m} \beta_{jj} x_{i,j}^2 + \varepsilon_i \quad \text{where } i = 1 \text{ to } n. \quad (1)$$

The DSD can also be augmented with two-level categorical factors where the experiment can estimate quadratic effects, the estimation of which decreases as more categorical factors are added. Augmenting the DSD with categorical factors in this way increases the required runs $n$ to $2m + 2$ [19].

### _Optimization Experiments_

A class of _Response Surface Method_ (RSM) experiments called _Central Composite Designs_ (CCD) were introduced in the mid-twentieth century [20]. The CCD embeds factorial design _Cube Points_ with _Center-Points_ and _Star-Points_ to better estimate curvature in the response; a CCD contains twice as many star-points as experimental design factors. The CCD can be rotated about its center-points when all points are equidistant from the design center and when the experiment is designed to provide constant prediction variance [21]. The experimental structure of a CCD is illustrated below, and shows how three factors for study can be arranged into DoE matrices as depicted by cubes in 3-dimensional space [22], where each experimental factor (A, B and C) is represented by one of the coordinate axes, the two-factor interactions (AB, AC and BC) are represented by each coordinate plane, and the three-factor interaction (ABC) is represented by the cube volume.

- The Figure 2 cube is a full-factorial DoE with three factors at two levels yielding eight different treatment combinations (shown as the red cube points). With just two levels per factor, only the linear effects can be estimated. This configuration is termed a full-factorial design because all the components of variation resulting from the three main effects, the three two-factor interactions, and the single three-factor interaction, can be separated in the analysis.

## Automated Copper Alloy Grain Size Evaluation Using a Deep-learning CNN

- The Figure 3 cube has a center-point (grey) added so that the presence of curvature may be detected; however, the specific source or sources of curvature cannot be ascertained because the component or components of variation from quadratic effects will be pooled with the effects from all three factors. This DOE is no longer a full-factorial because the center-point is shared with the eight cube points.

- The Figure 4 cube shows six face-centered star-points (grey) added to create a CCD RSM DoE. The star-points allow the sources of curvature to be better estimated, and since the star-points are centered on the faces of the cube, the matrix shown is termed a *Central-Composite Face-centered* (CCF) RSM, where the star-point distances from the center point are ±1 (a normalized unit of the factor level settings notated as +α or -α). For the CCF |α| = 1.0, and depending on the number of factors studied and the experimental strategy, the value of α can be modified to change characteristics of the RSM. The CCF is not a full-factorial matrix, because star-points are shared with the two associated factors in the cube face planes and the center-point is shared with the six star-points.

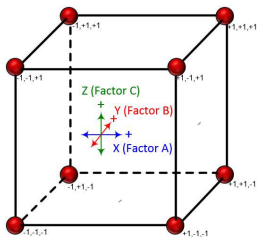 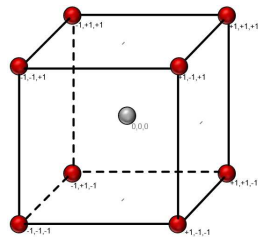 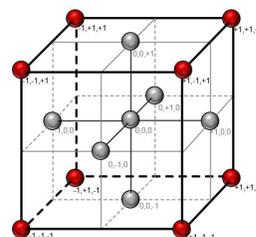 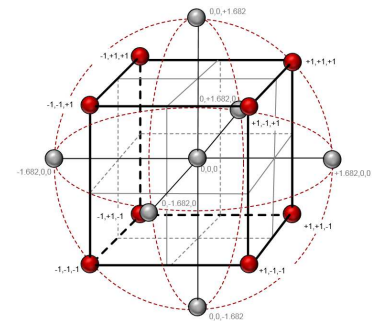

Figure 2 – 3-Factor Factorial Cube

Figure 3 – 3-Factor Factorial Cube with Center-Point

Figure 4 – 3-Factor Factorial Cube with Center-Point and 6 CCF Star-Points

Figure 5 – 3-Factor Factorial Cube with Center-Point and 6 CCC Star-Points

- The Figure 5 cube shows the star-points extended outward to create a *Central-Composite Circumscribed* (CCC) RSM, with α = 1.682, which allows the matrix to be rotated around the center-point in any orientation (each cube-point and star-point are on the surface of a sphere) [20] [21]. Conversely, if extending the factor extremes is not desirable or perhaps impossible (e.g. to prevent exceeding a design or process safe limit), the red cube points can be made into the star-points and the factor levels contracted by the reciprocal of 1.682, creating a *Central-Composite Inscribed* (CCI) RSM. The response surface may be mapped using sequential experimentation, where the RSM CCD can be moved over the experimental space by extending the matrix in any direction and then reusing part of the existing data for each move. Many small moves can be made sequentially along the path of steepest gradient descent using a method called *Evolutionary Operations* (EVOP) [14]. For n factors > 3, the experimental surface becomes a hypersurface of dimension n-1, within n-dimensional space.

To maintain the ability to rotate the experimental CCD matrix, the value of α will change as a function of the number of experimental factorial runs [21]:

$$\alpha = \left[n_f\right]^{1/4} \text{ where } n_f = \text{the number of factorial runs.} \qquad (2)$$

If the experiment is a full-factorial two-level design:

$$\alpha = \left[2^k\right]^{1/4} \text{ where } 2^k = n_f \text{ and } k = \text{the number of experimental factors.} \qquad (3)$$

The configuration of the CCD star-points impacts the ability to rotate the matrix around the center-points.

**Automated Copper Alloy Grain Size Evaluation Using a Deep-learning CNN**

# 3. Creating the Data Set

Five Cu-alloy bar-stock end sample coupons were used to create the initial training and test dataset. The generation of this dataset represented a significant portion of the overall effort for this work.

## 3.1. Preparing Samples from Metal Coupons

The Cu alloy coupons were prepared for inspection using standard methods.

- In preparation for inspection by microscope, sample coupons (Figure 6) from each piece of Cu-alloy bar-stock are mounted in an autopolisher carrier for polishing and etching (Figure 7, Figure 8).

- During the microscope inspection of the prepared sample coupons, grain size and grain size distribution are evaluated by a technician who inspects one small area at a time using metallograph images of magnified areas of the polished upper surface (Figure 9).

- Using a fixture that holds each coupon in a precise location in the microscope's XY focal plane (Figure 10), the polished upper surfaces of these same sample coupons are also fully imaged (Figure 11) under high-magnification using a robotic microscope that has been programmed to process arrays of sample coupons automatically.

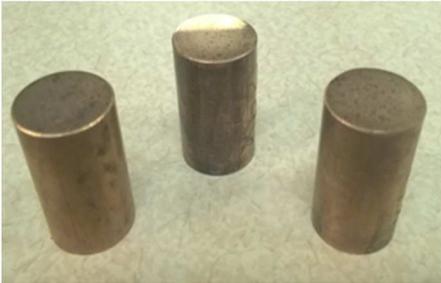
Figure 6 – Typical Cu-alloy polished coupons (3/4 and 5/8 in. diameter)

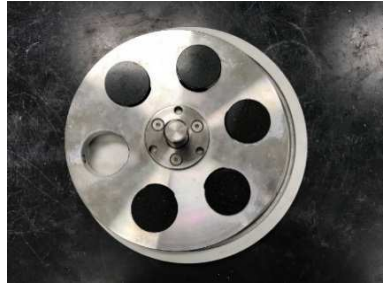
Figure 7 – Coupons in autopolisher carrier

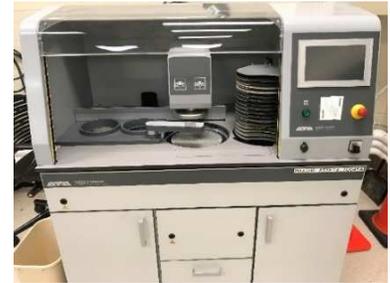
Figure 8 – Autopolisher

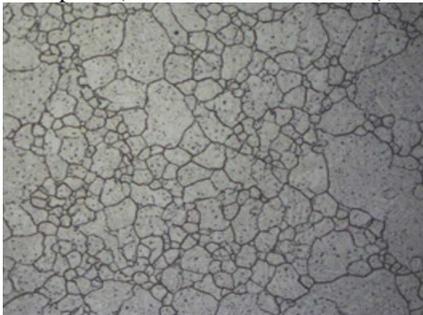
Figure 9 – Typical metallograph image (400X mag.)[5]

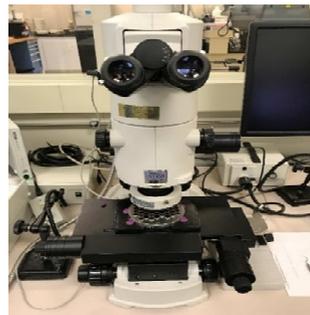
Figure 10 – Coupon tray in robotic microscope

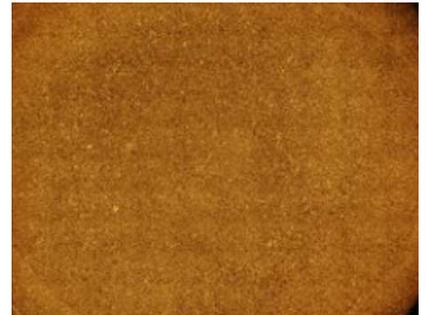
Figure 11 – Typical robotic microscope image (63X mag.)[6]

## 3.2. Pre-Processing into Sub-Images

The sample coupons images created by the robotic microscope were saved as 22000 x 16600 pixel TIFFs with an approximate size of 1.2 GB for each image file. To prepare these large images for creation of the deep learning dataset, they were partitioned into smaller sub-image tiles and algorithmically compressed into a different file format.

---

[5] Scale indicator intentionally omitted (see Section 8.1)

[6] Ibid

Using the *Image Slicer* API (see Section 8.3), the large images were sliced into 2025 490 x 368 pixels sub-image tiles, and then the tiles were compressed via interpolation using the *Scikit-image* API (see Section 8.3) and converted to 189 x 142 pixel RGB JPEGs, and ranging in file size from approximately 10 kB to 12 kB.

## 3.3. Classification of Sub-Images

The grain structure of each sub-image was classified as either 'good', 'bad' or 'neutral' using an internal process standard based on *ASTM E1282-97* [23] already used to determine if the metallograph-imaged grain structures are acceptable. The intent was to label data for a binary classification CNN in order to classify grain structures as either acceptable or not acceptable. Figure 12 notionally depicts this process, where the average of the largest orthogonal diameters ($d_1$ and $d_2$) are used to estimate the grain size[7]. The image samples with smaller gain sizes that would be obviously acceptable per the process standard were classified as 'good', while conversely, image samples with excessively large grain sizes that would be obviously rejected per the process standard were classified as 'bad'. The 'neutral' classification were grain structures that could be regarded as either marginally acceptable or non-acceptable.

The metallograph images used in the existing manual process only represented a small fraction of the total grains visible in a coupon, and by the existing process standard, the coupon would be rejected if any of the limited number of metallograph images sampled from the coupon were deemed unacceptable; therefore, it was hypothesized that to be successful, the CNN would only need to recognize either obviously good or bad grain structures.

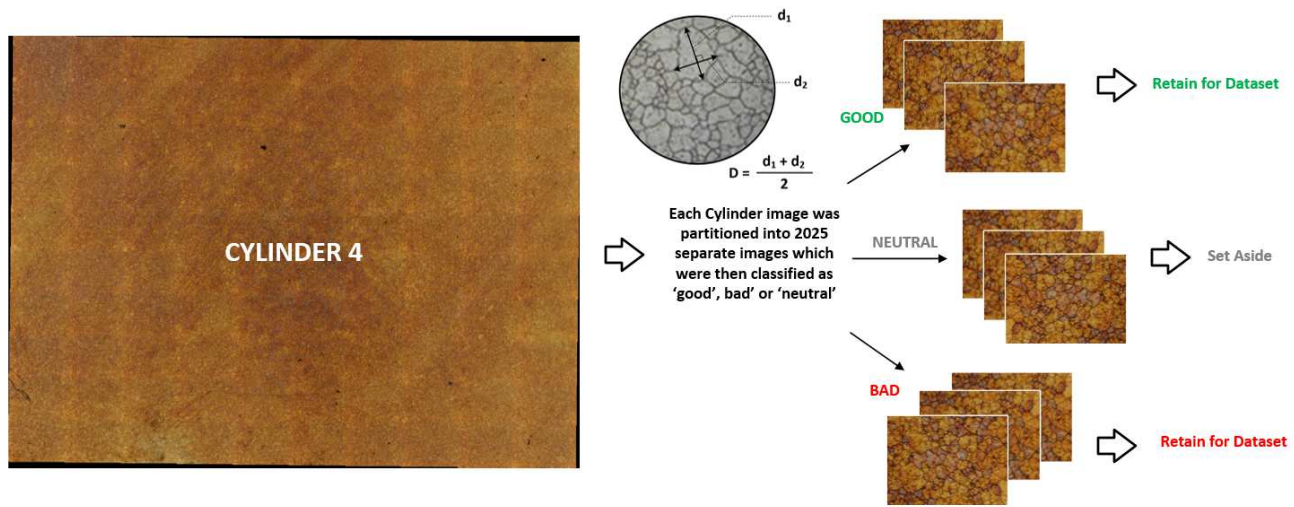

Figure 12 – Partitioning and classifying the sub-images into 'good, 'neutral' and 'bad' grain structure

This approach produced approximately twice as many examples of 'good' grain structures as 'bad' examples.

---

[7] Note: Lines $d_1$ and $d_2$ are two orthogonal lines, drawn in a manner to maximize the summation of their lengths

## 3.4.  Sample Images Normalization

Due to the large variations between coupon images in lighting, feature-contrast and color, the sub-images were RGB-weighted gray-scaled and histogram equalized [24] using the *Scikit-image* API. This transformation resulted in a loss of information and emphasized spatial features at the expense of color content; however:

- The metallograph-based grain structure inspection process standard uses only spatial features to accept or reject samples
- The SME-based classification of the sub-images into 'good' and 'bad' grain structure relied only on the spatial features
- The CNN was only intended to replicate the metallograph process.

Because of the above bullet-points, we were confident that the transformation process would not compromise the effectiveness of the CNN.

Any coupon boundary images were discarded because the black background severely reduced image contrast when the histogram equalization transformation was used.

The transformation is shown for several coupon sub-section samples in Figure 13 (original colored images) and Figure 14 (original images gray-scaled and histogram equalized). The transformation produces what is essentially feature-wise standardization for the dataset because all samples have an identical pixel-intensity mean and standard deviation.

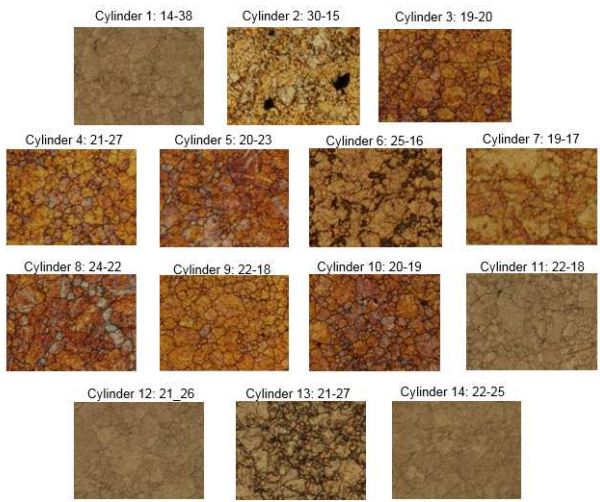

Figure 13 – Sub-images in original RGB color with variations in color and contrast

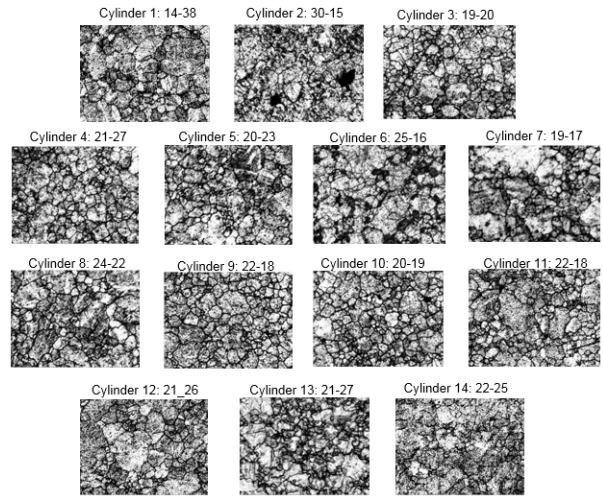

Figure 14 – Sub-images after RGB-weighted gray scaling and histogram equalization

## 3.5.  Ensemble Image Classification

After the gray-scaled histogram equalized tiles were classified as either 'good' or 'bad' by the CNN, the corresponding color tiles were then red color-tinted using a luminesce-preserving algorithm that varied the red intensity based on the classification probability (P) of the CNN, where P = 0 indicated a 0% probability and P = 1 indicated a 100% probability that the tile had good grain structure. Both the non-tinted and tint-classified tiles were reassembled[8] into ensemble images of the original Cu-alloy coupons that could be inspected by an operator. The ensemble images allowed a very accurate basis for either accepting or rejecting Cu alloy coupons.

The ensemble image also provided a forensic capability to understand why a coupon was either acceptable or not acceptable, because each sub-image tile classification could be checked both visually by tint and numerically by classification probability. Two examples are shown in Figure 15 for two different sample coupon images. The forensic capability for the coupon ensemble image provides some transparency into the machine-based classification process since the decision of the CNN could be understood down to the level of individual tile partitions. This transparency therefore represents a degree of explainable AI (XAI) [25].

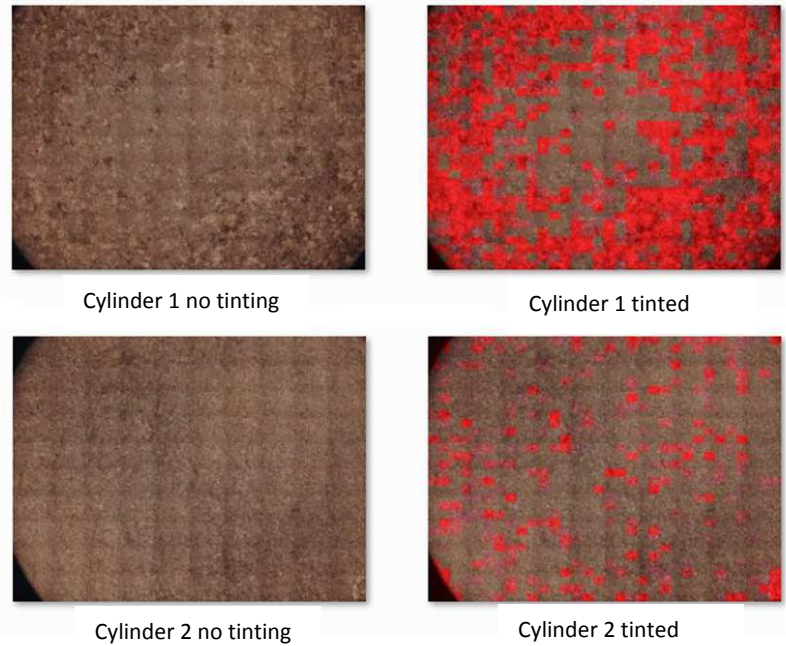

Figure 15 – Reconstructed ensemble tinted and non-tinted coupon images.

## 3.6. Data Augmentation

After sub-image classification was completed using image samples obtained from the first five Cu-alloy coupons, there were approximately twice as many 'good' examples of grain structure as there were 'bad' examples. This was expected since normally more Cu-alloy coupons are accepted by the existing process than are rejected. To eliminate possible model bias towards 'good' classifications, the CNN training dataset was balanced; therefore, the 'bad' examples were subjected to methods [26] of data augmentation using the *Keras* API (see Section 8.3) [27], which involved random rotations, vertical and horizontal flips and shifts, and shears[9], while employing a wrapping fill-mode for the displaced areas of the augmented images. Using this data augmentation, the quantity of 'bad' grain structure examples was doubled.

---

[8] The blurry rectangular patterns visible in the reconstructed imaged in Figure 15 are artifacts from how the robotic microscope saves the coupon images it creates; the edges of rectangular sub-images are overlapped by 1% with neighboring sub-images when the coupon image is reconstructed. These blurry artifacts are entrained in the CNN dataset.

[9] Random zooms were not used because this would have changed the apparent size of the grains.

## 3.7. Datasets

### 3.7.1. Initial Dataset

The initial dataset had 7420 total images (5020 for training, 1600 for training validation, and 800 for test verification), equally split between good and bad classifications. Approximately 50% of the bad grain structure examples, which represented 25% of the initial dataset, were created using the data augmentation methods described above.

### 3.7.2. Expanding the Dataset

Later, a larger dataset was constructed by adding examples from 14 additional Cu-alloy coupons to the initial dataset. No data augmentation was used with this additional data and instead the number of added 'good' examples was limited to the number of added 'bad' examples. The resulting combined larger dataset had 12300 total images (8700 for training, 2400 for training validation, and 1200 for test verification), equally split between good and bad classifications. Approximately 30% of the bad grain structure examples, which represented 15% of the larger dataset, were augmented images that came from the initial dataset.

## 4. Model Architecture

Simple CNN architecture was selected to resist overfitting. This is because a model that is physically limited in its data storage capacity will be forced to focus on the most significant features within the data, which translates to better generalization [28]. Therefore, the CNN described in the *Keras* blog article [27] was adapted to our application as shown in Figure 16 below. It uses only three convolutional layers with a limited number of filters in each layer for feature extraction, and only two fully-connected hidden dense layers for classification. Additionally, to disrupt random correlations within the data to further reduce the chances of overfitting, stochastic dropouts within layers and layer output pooling were also used. The *Appendix A: CNN Structure* provides additional details.

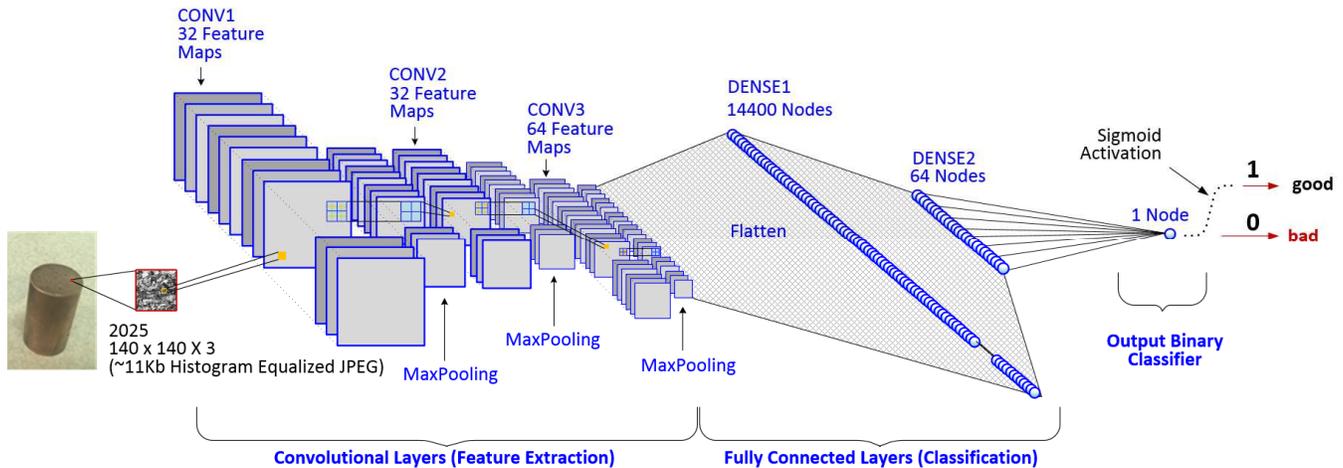

Figure 16 – Binary Classifier CNN Architecture

## 5. Training Methodology

Besides the dataset creation, the CNN training represented another substantial investment of effort. Therefore, to improve the efficiency of CNN training and optimization, we explored using DoE. If this approach was successful, it could be applied with other machine-learning and deep-learning projects.

Accordingly, the hyperparameters for both CNN layers and training were simultaneously studied as model factors using DoE. Initially, screening experiments were used to statistically screen a large number of hyperparameter factors to assess which factors should be studied in greater detail. The screening experiment was then followed by optimization experiments.

The following hyperparameters were selected for initial screening.

- *Training batch size (a subset of the training component of the dataset)*
- *Stride length and padding in the input convolutional layer*
- *Filter kernel sizes in the convolutional layers*
- *Activation functions on the layer outputs*
- *Max-normal kernel constraint in the densely-connected classification layers during training*
- *Stochastic dropouts in the convolutional and densely-connected layers during training*
- *Max-pooling after the convolutional layers*
- *Various optimization functions for training*

Note that these hyperparameters are already known to affect the performance of CNNs, and often these are set following best practices already established by researchers and practitioners of deep-learning [29]. As alluded to previously, we decided to test the efficacy of the DoE approach for tuning CNN performance by confirming the established best practices and other findings of machine/deep-learning research and theory to establish a proven optimization methodology for use on future projects.

The processing methodology developed for this project is notionally depicted in

Figure 17. Blue items were automated, while green items required manual intervention. The *Minitab* application was used to design and create the experimental matrices, which were then converted to CSV-format files that could be read by the main application written in the *Python* programming language. *Minitab* was also used to analyze the experimental output. The *Keras* API was used to create the model shown in Figure 16, while the *Tensorflow* API backend ran the model. Other specialized Python-based APIs were also used; refer to Section 8.3 for additional information. This approach was initially developed and tested on the MNIST public domain dataset using a different deep learning CNN model, which achieved 99.4% validation accuracy during training.

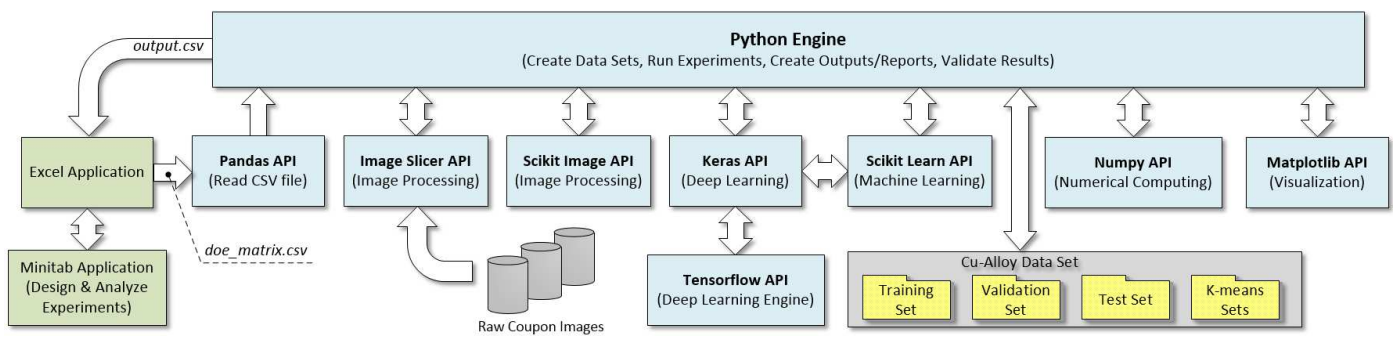

Figure 17 – Processing Methodology

## 5.1. Experimental Responses

Several response variables were used to analyze the results of experimental outputs (underlined responses were used for the experimental analysis).

*Group-A Responses*

- <u>Time</u>: the time required in minutes for the CNN to run through all the training epochs for each experimental treatment combination.
- <u>Training Accuracy (TRNACC)</u>: the average training accuracy of the last 5 epochs for each experimental treatment combination.
- <u>Validation Accuracy (VALACC)</u>: the average validation accuracy of the last 5 epochs for each experimental treatment combination.
- Confusion Matrix Outputs: metrics obtained by testing the classification performance of the trained CNN on the test dataset for each experimental treatment combination [30]. The confusion matrix is shown in Table 1.

| **Predicted Class Labels** | | | |
|---|---|---|---|
| **Class** | Bad Grains | Good Grains | True Rates |
| **Actual Class Labels** | | | |
| Bad Grains | True Negative (TN) | False Positive (FP) | Specificity = TN / (TN + FP) |
| Good Grains | False negative (FN) | True Positive (TP) | Sensitivity = TP / (TP + FN) |
| | False Positive Rate = FP / (FP + TN) | Precision = TP / (TP + FP) | Accuracy = (TN + TP) / (TN + FN + FP + TN) |

Table 1 – Confusion Matrix for the Classification CNN

- <u>Test Accuracy (TSTACC)</u>: the number of all correct predictions divided by the total number of the dataset.

*Group-B Responses*

- <u>Specificity</u> (True Negative Rate - TNR): the number of correct negative predictions divided by the total number of negatives.
- <u>Sensitivity</u> (Recall or True Positive Rate - TPR): the number of correct positive predictions divided by the total number of positives.
- False Positive Rate: the number of incorrect positive predictions divided by the total number of negatives (or 1 – Specificity).
- Precision (Positive Predictive Rate - PPR): the number of correct positive predictions divided by the total number of positive predictions.

Note that 'False Positive Rate' may be studied by analyzing 'Specificity'.

For the initial dataset:

- The test set size N = 800
- Actual bad grains = 400
- Actual good grains = 400

## 5.2. Screening Experiments

We defined the *Optimizer* function as a continuous experimental factor because the three different optimizer functions provide different levels of adaptive moment. The approach we used was to treat the differences in adaptive

moment—as represented by these optimizer functions, as the level settings. Note that coded units (i.e. 1, 2, and 3) were used in the actual DSD matrix instead of the categorical values.

The *Drops* for the three convolutional layers, as well as the first dense classification layer were varied as shown in Table 2 (*Conv1, Conv2,* and *Conv3,* as represented by the feature extraction and *Dense1* as represented by the classification parts, respectively, of the CNN shown in Figure 16). The mid-levels used in the DSD experiment were set to 20% for the convolutional layers, and 30% for the dense layer.

*Pooling* between convolutional layers was either not-applied ("none") or applied ("two") as a categorical experimental factor.

The two-dimensional[10] *Filter* kernel sizes for the convolutional layers were varied between *3×3, 5×5* and *7×7* pixels. *Stride* was applied to the convolution of the *Filter* kernel with the input image for the convolutional layer 1 ( CONV1*)*; it was set to either a *Stride* length of "one" or "two" as a categorical experimental factor. When set to "two," the *Filter* kernel would be stepped across two pixels in the input image for each convolution, skipping a line of pixels on the input image [31].

*Padding* in the convolutional layers was either applied ("same") or not applied ("valid") as a categorical experimental factor. If applied, the level of *Padding* for all convolutional layers was the "same" as the *Stride* factor setting for CONV1, CONV2 and CONV3. When *Padding* = "same" and S*tride* = "one" or "two," rows of zeros would be added (i.e. padded) to the far input image borders to accommodate the *Stride* length and *Filter* kernel sizes. When *Padding* = "valid," no rows of zeros would be added to the input image borders, regardless of the *Stride* setting and *Filter* kernel sizes [31].

The *Activation* function was defined as a continuous experimental factor, although categorical values were selected for the levels based on resistance to vanishing gradients. Note that coded units (i.e. 1, 2 and 3) were used in the actual DSD matrix instead of the categorical values.

## 5.2.1. DSD Factors and Levels

A total of 16 factors, which were a combination of CNN layer and training hyperparameters, were selected for study using a DSD with 34 treatment-combinations[11]. The presence of five categorical factors increased the number of treatment combinations to $2m + 2$. The DSD factors, class (training or CNN layer), type (continuous or categorical) and levels (1, 2, and 3) are shown in Table 2. Note the use of coded units for the *Optimizer* and *Activation* factors, which had three categorical levels for each of these factors. Conversely, several factors such as *Max-Pooling* that could be called continuous have limited settings, so these were treated as categorical factors.

| | Factor Name | Factor Class | Factor Type | Level 1 | Level 2 | Level 3 |
|---|---|---|---|---|---|---|
| **1.** | Batch Size | Training | Continuous | 80 | 160 | 240 |
| **2.** | Kernel Constraint | Training | Continuous | 3 | 5 | 7 |
| **3.** | Optimizer | Training | Continuous[12] | Adam | AdaMax | Nadam |
| **4.** | Drops Conv1 | Training | Continuous | 0% | 10% | 20% |
| **5.** | Drops Conv2 | Training | Continuous | 0% | 10% | 20% |
| **6.** | Drops Conv3 | Training | Continuous | 0% | 10% | 20% |
| **7.** | Drops Dense1 | Training | Continuous | 10% | 30% | 50% |
| **8.** | Max-Pooling Conv1 | CNN Layer | Categorical | None | Two | |
| **9.** | Max-Pooling Conv2 | CNN Layer | Categorical | None | Two | |

---

[10] The images processed were RGB grayscale, with three color channels, so the kernel filters were actually 3 × 3  3, 5 × 5 × 3 and 7 × 7 × 3.

[11] See the glossary in Table 35.

[12] This is a categorical factor turned quasi-continuous. The matrix used coded units where 1 = Adam, 2 = AdaMax and 3 = Nadam.

| 10. | Max-Pooling Conv3 | CNN Layer | Categorical | None | Two | |
|---|---|---|---|---|---|---|
| 11. | Filter Conv1 | CNN Layer | Continuous | 3 | 5 | 7 |
| 12. | Filter Conv2 | CNN Layer | Continuous | 3 | 5 | 7 |
| 13. | Filter Conv3 | CNN Layer | Continuous | 3 | 5 | 7 |
| 14. | Padding | CNN Layers | Categorical | Same | Valid | |
| 15. | Stride Conv1 | CNN Layer | Categorical | One | Two | |
| 16. | Activation | CNN Layers | Continuous[13] | TanH | SELU | ReLU |

Table 2 – DSD Experiment Factor Information

## 5.2.2. DSD Experimental Matrix

The DSD experimental matrix[14] is shown in Table 3, and is given in standard design order; the actual run order was randomized upon execution. The blue rows indicate the treatment combinations where the CNN model was able to successfully learn the data set.

| TC | Batch | Kern Const | Optm | Drop C1 | Drop C2 | Drop C3 | Drop D1 | MaxP C1 | MaxP C2 | MaxP C3 | Filt C1 | Filt C2 | Filt C3 | Padd | Stride C1 | Active |
|---|---|---|---|---|---|---|---|---|---|---|---|---|---|---|---|---|
| 1 | 160 | 7 | 3 | 0.2 | 0.2 | 0.2 | 0.5 | 2 | 2 | 2 | 7 | 7 | 7 | S | 2 | 3 |
| 2 | 160 | 3 | 1 | 0 | 0 | 0 | 0.1 | 0 | 0 | 0 | 3 | 3 | 3 | V | 1 | 1 |
| 3 | 240 | 5 | 1 | 0 | 0.2 | 0 | 0.5 | 2 | 0 | 2 | 3 | 3 | 7 | V | 2 | 3 |
| 4 | 80 | 5 | 3 | 0.2 | 0 | 0.2 | 0.1 | 0 | 2 | 0 | 7 | 7 | 3 | S | 1 | 1 |
| 5 | 240 | 7 | 2 | 0 | 0 | 0.2 | 0.1 | 2 | 0 | 2 | 7 | 3 | 3 | S | 1 | 3 |
| 6 | 80 | 3 | 2 | 0.2 | 0.2 | 0 | 0.5 | 0 | 2 | 0 | 3 | 7 | 7 | V | 2 | 1 |
| 7 | 240 | 7 | 3 | 0.1 | 0 | 0 | 0.5 | 0 | 0 | 2 | 7 | 7 | 3 | V | 2 | 1 |
| 8 | 80 | 3 | 1 | 0.1 | 0.2 | 0.2 | 0.1 | 2 | 2 | 0 | 3 | 3 | 7 | S | 1 | 3 |
| 9 | 240 | 3 | 3 | 0.2 | 0.1 | 0 | 0.1 | 2 | 0 | 0 | 7 | 7 | 7 | V | 1 | 3 |
| 10 | 80 | 7 | 1 | 0 | 0.1 | 0.2 | 0.5 | 0 | 2 | 2 | 3 | 3 | 3 | S | 2 | 1 |
| 11 | 240 | 7 | 1 | 0.2 | 0.2 | 0.1 | 0.1 | 0 | 0 | 2 | 3 | 5 | 5 | S | 1 | 1 |
| 12 | 80 | 3 | 3 | 0 | 0 | 0.1 | 0.5 | 2 | 2 | 0 | 7 | 3 | 3 | V | 2 | 3 |
| 13 | 240 | 3 | 3 | 0 | 0.2 | 0.2 | 0.3 | 0 | 0 | 0 | 7 | 3 | 7 | S | 2 | 1 |
| 14 | 80 | 7 | 1 | 0.2 | 0 | 0 | 0.3 | 2 | 2 | 2 | 3 | 7 | 3 | V | 1 | 3 |
| 15 | 240 | 3 | 1 | 0.2 | 0 | 0.2 | 0.5 | 0 | 0 | 0 | 3 | 7 | 3 | S | 2 | 3 |
| 16 | 80 | 7 | 3 | 0 | 0.2 | 0 | 0.1 | 2 | 2 | 2 | 7 | 3 | 7 | V | 1 | 1 |
| 17 | 240 | 7 | 3 | 0.2 | 0.2 | 0.2 | 0.5 | 2 | 0 | 0 | 3 | 3 | 3 | V | 1 | 1 |
| 18 | 80 | 3 | 1 | 0 | 0 | 0 | 0.1 | 0 | 2 | 2 | 7 | 7 | 7 | S | 2 | 3 |
| 19 | 240 | 3 | 1 | 0 | 0 | 0 | 0.5 | 2 | 2 | 0 | 7 | 7 | 3 | S | 1 | 1 |
| 20 | 80 | 7 | 3 | 0.2 | 0 | 0.2 | 0.1 | 0 | 0 | 2 | 3 | 3 | 7 | V | 2 | 3 |
| 21 | 240 | 7 | 1 | 0 | 0 | 0.2 | 0.1 | 2 | 2 | 0 | 5 | 7 | 7 | V | 2 | 1 |
| 22 | 80 | 3 | 3 | 0.2 | 0.2 | 0 | 0.5 | 0 | 0 | 2 | 5 | 3 | 3 | S | 1 | 3 |
| 23 | 240 | 7 | 3 | 0 | 0 | 0 | 0.5 | 0 | 2 | 2 | 3 | 5 | 7 | S | 1 | 3 |
| 24 | 80 | 3 | 1 | 0.2 | 0.2 | 0.2 | 0.1 | 2 | 0 | 2 | 7 | 5 | 3 | V | 2 | 1 |
| 25 | 240 | 3 | 3 | 0.2 | 0 | 0 | 0.1 | 2 | 2 | 2 | 3 | 3 | 5 | S | 2 | 1 |
| 26 | 80 | 7 | 1 | 0 | 0.2 | 0.2 | 0.5 | 0 | 0 | 0 | 7 | 7 | 5 | V | 1 | 3 |
| 27 | 240 | 7 | 1 | 0.2 | 0.2 | 0 | 0.1 | 0 | 2 | 0 | 7 | 3 | 3 | V | 2 | 3 |
| 28 | 80 | 3 | 3 | 0 | 0 | 0.2 | 0.5 | 2 | 0 | 2 | 3 | 7 | 7 | S | 1 | 1 |
| 29 | 240 | 3 | 3 | 0 | 0.2 | 0.2 | 0.1 | 0 | 2 | 2 | 3 | 7 | 3 | V | 1 | 3 |
| 30 | 80 | 7 | 1 | 0.2 | 0 | 0 | 0.5 | 2 | 0 | 0 | 3 | 7 | 7 | S | 2 | 1 |
| 31 | 240 | 3 | 1 | 0.2 | 0 | 0.2 | 0.5 | 0 | 2 | 2 | 7 | 3 | 7 | V | 1 | 2 |
| 32 | 80 | 7 | 3 | 0 | 0.2 | 0 | 0.1 | 0 | 0 | 0 | 3 | 7 | 3 | S | 2 | 2 |
| 33 | 160 | 5 | 2 | 0.1 | 0.1 | 0.1 | 0.3 | 0 | 0 | 0 | 5 | 5 | 5 | V | 1 | 2 |
| 34 | 160 | 5 | 2 | 0.1 | 0.1 | 0.1 | 0.3 | 2 | 2 | 2 | 5 | 5 | 5 | S | 2 | 2 |

---

[13] This is another (the second) categorical factor turned quasi-continuous. The matrix used coded units where 1 = TanH, 2 = SELU and 3 = ReLU.

[14] TC = Treatment Combination, Batch = Batch Size, KCon = Kernel Const, Optm (1,2,3) = Optimizer (Adam, AdaMax, Nadam), Drop C1/C2/C3/D1 = Dropout CONV1/CONV2/CONV3/DENSE1, MaxP C1/C2/C3 = Max Pooling CONV1/CONV2/CONV3, Filt C1/C2/C3 = Filter Size CONV1/CONV2/CONV3, Padd = Padding on CONV1/CONV2/CONV3, StrideC1 = Stride CONV1, Active (1,2,3) = Activation (TanH, SELU, ReLU).

**Automated Copper Alloy Grain Size Evaluation Using a Deep-learning CNN**

Table 3 – DSD Experimental Matrix

## 5.2.3.        DSD Experimental Results

Each treatment combination in the matrix was iterated through 35 training epochs, where for each epoch the CNN model is subjected to one forward pass and one backward pass for all training images, with the number of iterations in the epoch determined by the *Batch* size. During the forward pass, the CNN model attempts to fit the data in the *Batch* using the neuron weights between layers (initially randomized at the start of training), and then the predicted outputs are evaluated against the expected (known) outputs of the training examples. During the backward pass (backpropagation), neuron weights between layers are adjusted to minimize the penalty as calculated by a binary cross-entropy loss function.  The number of iterations for each epoch is equal to the number of training samples divided by the batch size: for *Batch* sizes of 80, 160 and 240, there were 63, 32 and 21 iterations respectively. At the end of each epoch, the CNN model attempts to classify the validation image samples (not seen during the training iterations) and based on the loss calculation on the validation set, the weights are readjusted by the *Optimizer* to achieve the steepest gradient descent of the hyperplane of the loss function.  Once all 35 epochs were completed, the trained CNN model weights with the highest validation accuracy[15] from the 35 epochs was shown the test image samples. The test image samples were not part of either the training or validation sets used during the epoch training iterations. The test image evaluation produces a confusion matrix output for each experimental treatment combination as described in Section 5.1.

The responses for the 34 treatments combinations are given in Table 4. The training and validation accuracies represent the average values of the last 5 epochs (31 to 35) and test accuracy represented the overall test accuracy achieved on the test data set.  Only 5 (14.7%) of the treatment combinations (highlighted blue) learned the training data[16].

| TC | EPOCH | First Experiment (Group-A) | | | | | | | Replicated Experiment (Group-B) | | | | | | |
|---|---|---|---|---|---|---|---|---|---|---|---|---|---|---|---|
| | | Time | TRN ACC | VAL ACC | TST ACC | TPR | TNR | PPR | Time | TRN ACC | VAL ACC | TST ACC | TPR | TNR | PPR |
| 1 | 35 | 10.78 | 0.5043 | 0.4999 | 0.5000 | 0.0000 | 1.0000 | 0.0000 | 10.93 | 0.4990 | 0.4994 | 0.5000 | 0.0000 | 1.0000 | 0.000 |
| 2 | 35 | 13.32 | 0.5000 | 0.5023 | 0.5000 | 1.0000 | 0.0000 | 0.5000 | 14.00 | 0.4931 | 0.4991 | 0.5000 | 0.0000 | 1.0000 | 0.0000 |
| 3 | 35 | 10.18 | 0.7817 | 0.7897 | 0.8013 | 0.8375 | 0.7650 | 0.7809 | 10.13 | 0.7736 | 0.7876 | 0.7975 | 0.8525 | 0.7425 | 0.0000 |
| 4 | 35 | 11.53 | 0.5007 | 0.4993 | 0.5000 | 1.0000 | 0.0000 | 0.5000 | 11.47 | 0.5004 | 0.5007 | 0.5000 | 0.0000 | 1.0000 | 0.7680 |
| 5 | 35 | 10.38 | 0.7177 | 0.7471 | 0.7575 | 0.9450 | 0.5700 | 0.6873 | 10.32 | 0.7810 | 0.7995 | 0.8138 | 0.9100 | 0.7175 | 0.5000 |
| 6 | 35 | 10.65 | 0.4912 | 0.4984 | 0.5000 | 1.0000 | 0.0000 | 0.5000 | 10.58 | 0.4953 | 0.5045 | 0.5000 | 0.0000 | 1.0000 | 0.7631 |
| 7 | 35 | 10.22 | 0.5001 | 0.4990 | 0.5000 | 0.0000 | 1.0000 | 0.0000 | 10.27 | 0.5001 | 0.4994 | 0.5000 | 1.0000 | 0.0000 | 0.5000 |
| 8 | 35 | 10.68 | 0.4924 | 0.5022 | 0.5000 | 0.0000 | 1.0000 | 0.0000 | 10.83 | 0.4955 | 0.4999 | 0.5000 | 1.0000 | 0.0000 | 0.5000 |
| 9 | 35 | 10.57 | 0.5003 | 0.4971 | 0.5000 | 1.0000 | 0.0000 | 0.5000 | 10.60 | 0.4988 | 0.4991 | 0.5000 | 1.0000 | 0.0000 | 0.5000 |
| 10 | 35 | 10.50 | 0.4959 | 0.5013 | 0.5000 | 0.0000 | 1.0000 | 0.0000 | 10.57 | 0.4946 | 0.4980 | 0.5000 | 1.0000 | 0.0000 | 0.5000 |
| 11[17] | 35 | 10.88 | 0.5362 | 0.5418 | 0.5438 | 0.6480 | 0.4408 | 0.5368 | 10.91 | 0.5402 | 0.5418 | 0.5438 | 0.6070 | 0.4805 | 0.5000 |
| 12 | 35 | 10.47 | 0.4933 | 0.5082 | 0.5000 | 0.0000 | 1.0000 | 0.0000 | 10.50 | 0.4965 | 0.4996 | 0.5000 | 0.0000 | 1.0000 | 0.5388 |
| 13 | 35 | 10.87 | 0.5004 | 0.5003 | 0.5000 | 0.0000 | 1.0000 | 0.0000 | 10.93 | 0.5007 | 0.4999 | 0.5000 | 0.0000 | 1.0000 | 0.0000 |
| 14 | 35 | 10.57 | 0.7876 | 0.8112 | 0.8075 | 0.9025 | 0.7125 | 0.7584 | 10.57 | 0.7844 | 0.8135 | 0.8138 | 0.9850 | 0.6425 | 0.0000 |
| 15 | 35 | 10.87 | 0.4948 | 0.4991 | 0.5000 | 0.0000 | 1.0000 | 0.0000 | 10.93 | 0.4998 | 0.4998 | 0.5000 | 0.0000 | 1.0000 | 0.7337 |
| 16 | 35 | 10.52 | 0.4999 | 0.4996 | 0.5000 | 0.0000 | 1.0000 | 0.0000 | 10.57 | 0.5000 | 0.5002 | 0.5000 | 1.0000 | 0.0000 | 0.5000 |
| 17 | 35 | 10.78 | 0.4916 | 0.5013 | 0.5000 | 1.0000 | 0.0000 | 0.5000 | 10.77 | 0.5000 | 0.4999 | 0.5000 | 0.0000 | 1.0000 | 0.5000 |
| 18 | 35 | 10.55 | 0.7077 | 0.7111 | 0.7738 | 0.8475 | 0.7000 | 0.7386 | 10.57 | 0.7362 | 0.7427 | 0.8013 | 0.7175 | 0.8850 | 0.0000 |
| 19 | 35 | 10.33 | 0.4911 | 0.5030 | 0.5000 | 1.0000 | 0.0000 | 0.5000 | 10.42 | 0.4958 | 0.4997 | 0.5000 | 0.0000 | 1.0000 | 0.8619 |

[15] The Keras API 'checkpoint' feature was used to save the highest validation accuracy weights during training, and the best-performing model weights from the 35 epochs within each treatment was used to perform the test image classification after each treatment was completed. See Section 8.3 for additional information about Keras.

[16] In our experience with industrial screening experiments, the reverse normally occurs, where 75% or more of the treatments yield positive results (during screening, the factor levels are usually extended beyond normal limits to encourage 'bad' responses.

[17] Treatment 11 would not iterate due to a memory error in the GPU; this error was repeatable between the first and replicated experimental runs.

## Automated Copper Alloy Grain Size Evaluation Using a Deep-learning CNN

| TC | EPOCH | First Experiment (Group-A) | | | | | | | Replicated Experiment (Group-B) | | | | | | |
|----|----|------|---------|---------|---------|--------|--------|--------|------|---------|---------|---------|--------|--------|--------|
| | | Time | TRN ACC | VAL ACC | TST ACC | TPR | TNR | PPR | Time | TRN ACC | VAL ACC | TST ACC | TPR | TNR | PPR |
| 20 | 35 | 10.62 | 0.5005 | 0.4993 | 0.5000 | 1.0000 | 0.0000 | 0.5000 | 10.67 | 0.4908 | 0.5009 | 0.5000 | 1.0000 | 0.0000 | 0.0000 |
| 21 | 35 | 10.08 | 0.4949 | 0.4979 | 0.5000 | 1.0000 | 0.0000 | 0.5000 | 10.12 | 0.5002 | 0.5008 | 0.5000 | 1.0000 | 0.0000 | 0.5000 |
| 22 | 35 | 11.37 | 0.4915 | 0.4820 | 0.5000 | 1.0000 | 0.0000 | 0.5000 | 11.57 | 0.4962 | 0.5008 | 0.5000 | 1.0000 | 0.0000 | 0.5000 |
| 23 | 35 | 11.33 | 0.5024 | 0.5141 | 0.5025 | 0.9900 | 0.0150 | 0.5013 | 10.97 | 0.4924 | 0.5016 | 0.5000 | 0.0000 | 1.0000 | 0.5000 |
| 24 | 35 | 10.50 | 0.4944 | 0.5045 | 0.5000 | 1.0000 | 0.0000 | 0.5000 | 10.47 | 0.4962 | 0.5698 | 0.4825 | 0.0550 | 0.9100 | 0.5000 |
| 25 | 35 | 10.08 | 0.4980 | 0.5030 | 0.5000 | 1.0000 | 0.0000 | 0.5000 | 10.15 | 0.4967 | 0.4994 | 0.5000 | 1.0000 | 0.0000 | 0.3793 |
| 26 | 35 | 14.25 | 0.4886 | 0.5002 | 0.5000 | 1.0000 | 0.0000 | 0.5000 | 14.28 | 0.4921 | 0.4992 | 0.5000 | 1.0000 | 0.0000 | 0.5000 |
| 27 | 35 | 10.23 | 0.4975 | 0.4981 | 0.5000 | 1.0000 | 0.0000 | 0.5000 | 10.17 | 0.4968 | 0.4988 | 0.5000 | 1.0000 | 0.0000 | 0.5000 |
| 28 | 35 | 10.72 | 0.4992 | 0.4997 | 0.5000 | 1.0000 | 0.0000 | 0.5000 | 10.65 | 0.5003 | 0.5016 | 0.5000 | 1.0000 | 0.0000 | 0.5000 |
| 29 | 35 | 10.33 | 0.5008 | 0.5003 | 0.5000 | 0.0000 | 1.0000 | 0.5000 | 10.28 | 0.5017 | 0.4980 | 0.5000 | 1.0000 | 0.0000 | 0.5000 |
| 30 | 35 | 10.65 | 0.4909 | 0.5030 | 0.5000 | 1.0000 | 0.0000 | 0.5000 | 10.55 | 0.4905 | 0.5004 | 0.5000 | 0.0000 | 1.0000 | 0.5000 |
| 31 | 35 | 10.28 | 0.4988 | 0.5017 | 0.5000 | 1.0000 | 0.0000 | 0.5000 | 10.27 | 0.4989 | 0.5032 | 0.5000 | 1.0000 | 0.0000 | 0.5000 |
| 32 | 35 | 10.65 | 0.4994 | 0.5019 | 0.5000 | 1.0000 | 0.0000 | 0.5000 | 10.62 | 0.4996 | 0.5002 | 0.5000 | 1.0000 | 0.0000 | 0.5000 |
| 33 | 35 | 13.92 | 0.4993 | 0.4975 | 0.5000 | 1.0000 | 0.0000 | 0.5000 | 14.05 | 0.4996 | 0.5011 | 0.5000 | 0.0000 | 1.0000 | 0.5000 |
| 34 | 35 | 10.25 | 0.7869 | 0.8055 | 0.8025 | 0.8600 | 0.7850 | 0.8000 | 10.18 | 0.7829 | 0.7694 | 0.7350 | 0.5100 | 0.9600 | 0.0000 |

Table 4 – DSD Response Matrix

The experimental run-time required approximately 6 hours and 10 minutes to conduct all 35 epochs for each treatment; the experimental run-time in minutes is equal to the sum of the 'Time' column responses.

To validate the results of the first experiment and also improve the precision of the responses, the experiment was repeated (replicated), which is not normally used in screening experiments, but was used in this case because of the stochastic nature of CNN training. The replicated experiment provided nearly identical results to the first experiment, including Treatment 11, which would not run in the GPU[18].

The treatments (3, 5, 14, 18 and 34) where the CNN model was able to learn the dataset are shown in Table 5. Normally, for CNN training, the number of training epochs would be raised until the training accuracy stops increasing, but in the case for this screening experiment, the training epochs were kept to a relatively low number because the goal was to capture the influence on accuracy from significant factors and reduce the time needed to run the experiment.

| TC | First Experiment | | Replicated Experiment | |
|----|----|----|----|----|
| | Model Accuracy Plots[19] | Test Confusion Matrix | Model Accuracy Plots[18] | Test Confusion Matrix |
| 3 | | | | |

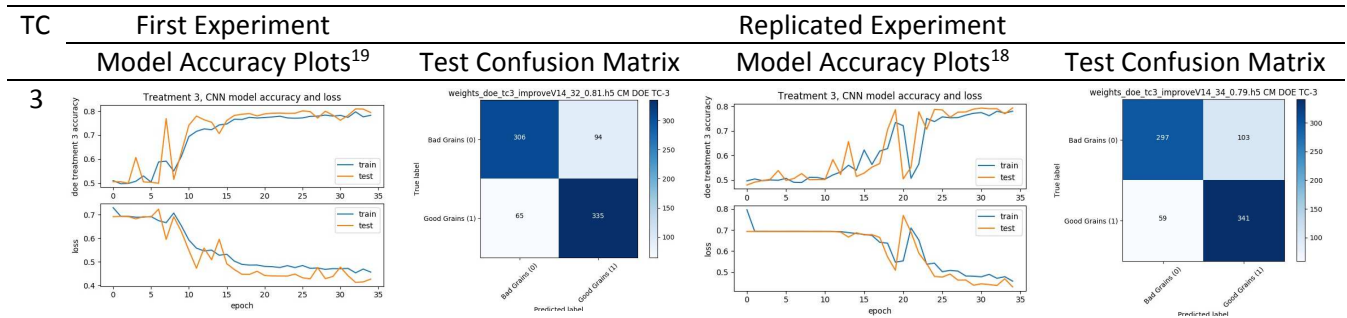

---

                5/19/2020

# Automated Copper Alloy Grain Size Evaluation Using a Deep-learning CNN

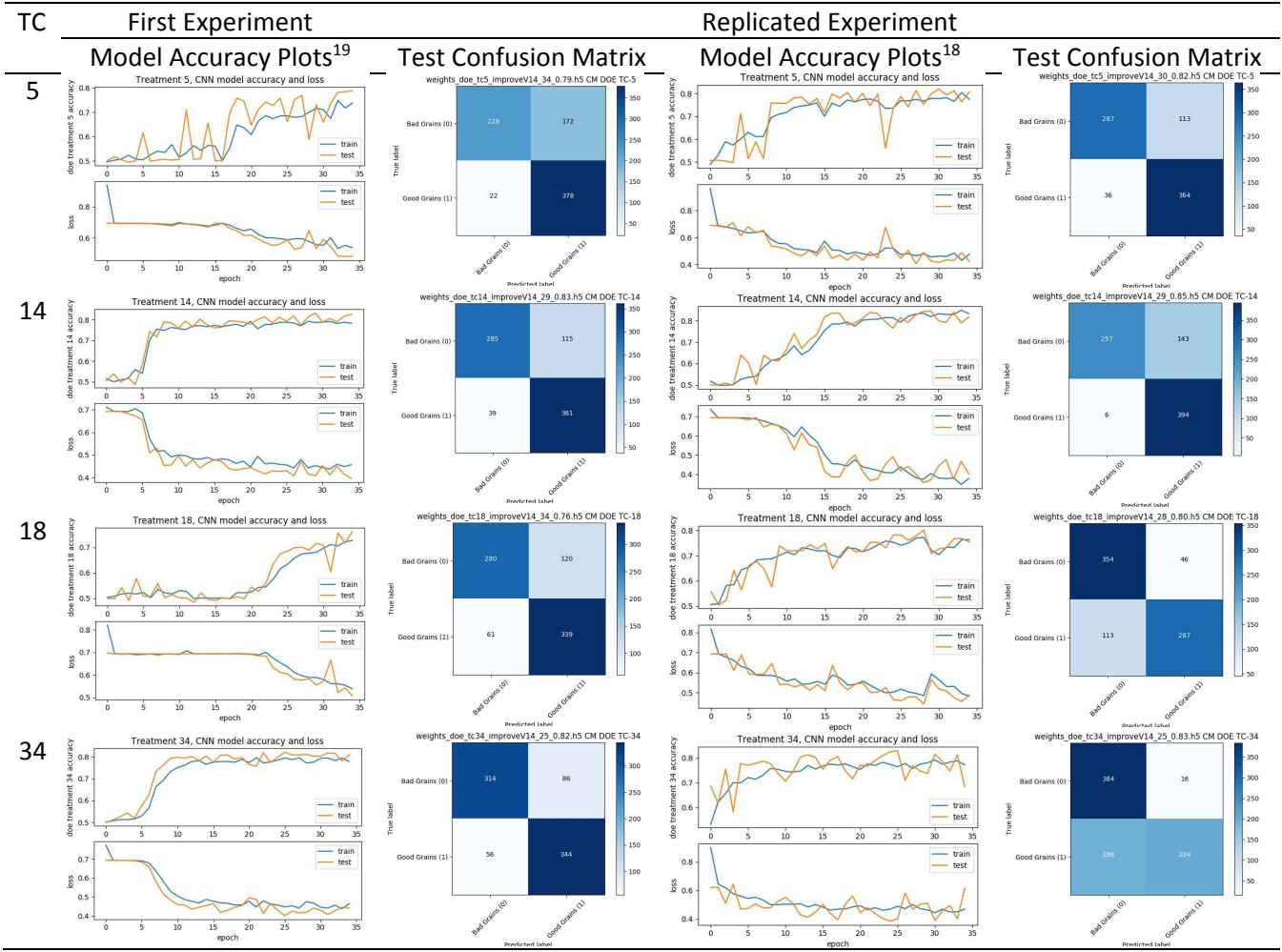

Table 5 – DSD Experimental Outputs (for Converging Treatments)

Notice that the orange test accuracy plots[18] are generally higher than the blue training accuracy plots, which provided good indication that the CNN was not overfitting.

The stochastic nature of the training used for the CNN model produced different results using identical datasets[20]. In some cases, the output training accuracy plots and test confusion matrices exhibited similar patterns between the first and replicated experiments. In other cases, such as Treatment 34, the replicated CNN model exhibited poor generalization performance, as evidenced by the misclassification of good grains in the test dataset as shown in the confusion matrix. Experimental replication helps to overcome these variations for the analysis, and provides better precision when attempting to estimate the Equation (1) model $\beta$ coefficients.

Because of the low number of treatments that were able to successfully learn the dataset, there was an initial temptation to revisit the experimental design and redo the entire experimental screening phase. Instead, we decided to continue the project with the screening results we had.

The full ANOVA results for the statistical model of the DSD experiment test accuracy are given in Table 6.

---

[20] The CNN model weights are initialized randomly at the start of each training epoch, a percentage of the weights are stochastically dropped during training and a stochastic gradient descent optimizer is used while training.

## Automated Copper Alloy Grain Size Evaluation Using a Deep-learning CNN

**Analysis of Variance (ANOVA)**

**Test Accuracy**

| Source | DF | Seq. SS | Contribution | Adj. SS | Adj. MS | F-Value | P-Value | VIF |
|---|---|---|---|---|---|---|---|---|
| Model | 27 | 0.63124 | 87.27% | 0.63124 | 0.02338 | 10.1 | 0.000 | |
| Linear | 16 | 0.40527 | 56.03% | 0.41126 | 0.02570 | 11.1 | 0.000 | |
| batch | 1 | 0.00011 | 0.02% | 0.00440 | 0.00440 | 1.9 | 0.174 | 1.05 |
| constrain | 1 | 0.00876 | 1.21% | 0.00631 | 0.00631 | 2.7 | 0.106 | 1.07 |
| optimize | 1 | 0.05782 | 7.99% | 0.08666 | 0.08666 | 37.6 | 0.000 | 1.08 |
| dropC1 | 1 | 0.01859 | 2.57% | 0.01483 | 0.01483 | 6.4 | 0.015 | 1.06 |
| dropC2 | 1 | 0.02021 | 2.79% | 0.01074 | 0.01074 | 4.6 | 0.037 | 1.07 |
| dropC3 | 1 | 0.02578 | 3.56% | 0.03279 | 0.03279 | 14.2 | 0.001 | 1.03 |
| dropD1 | 1 | 0.00631 | 0.87% | 0.00515 | 0.00515 | 2.2 | 0.143 | 1.07 |
| maxpC1 | 1 | 0.03903 | 5.40% | 0.02949 | 0.02949 | 12.8 | 0.001 | 1.10 |
| maxpC2 | 1 | 0.00007 | 0.01% | 0.00298 | 0.00298 | 1.2 | 0.262 | 1.10 |
| maxpC3 | 1 | 0.13673 | 18.90% | 0.12961 | 0.12961 | 56.3 | 0.000 | 1.11 |
| filtC1 | 1 | 0.00011 | 0.01% | 0.00042 | 0.00042 | 0.1 | 0.673 | 1.04 |
| filtC2 | 1 | 0.00410 | 0.57% | 0.00374 | 0.00374 | 1.6 | 0.210 | 1.09 |
| filtC3 | 1 | 0.00007 | 0.01% | 0.00005 | 0.00005 | 0.0 | 0.887 | 1.10 |
| padding | 1 | 0.00318 | 0.44% | 0.00930 | 0.00930 | 4.0 | 0.051 | 1.11 |
| strideC1 | 1 | 0.00005 | 0.01% | 0.00005 | 0.00005 | 0.0 | 0.882 | 1.10 |
| activation | 1 | 0.08435 | 11.66% | 0.06999 | 0.06999 | 30.4 | 0.000 | 1.07 |
| Square | 11 | 0.22597 | 31.24% | 0.22597 | 0.02054 | 8.9 | 0.000 | |
| batch*batch | 1 | 0.00492 | 0.68% | 0.00608 | 0.00608 | 2.6 | 0.112 | 1.60 |
| constrain*constrain | 1 | 0.08655 | 11.97% | 0.05593 | 0.05593 | 24. | 0.000 | 1.60 |
| optimize*optimize | 1 | 0.04133 | 5.71% | 0.04931 | 0.04931 | 21.4 | 0.000 | 1.60 |
| dropC1*dropC1 | 1 | 0.00938 | 1.30% | 0.00608 | 0.00608 | 2.6 | 0.112 | 1.60 |
| dropC2*dropC2 | 1 | 0.00640 | 0.89% | 0.00608 | 0.00608 | 2.6 | 0.112 | 1.60 |
| dropC3*dropC3 | 1 | 0.00062 | 0.09% | 0.00145 | 0.00145 | 0.6 | 0.433 | 2.83 |
| dropD1*dropD1 | 1 | 0.05111 | 7.07% | 0.06166 | 0.06166 | 26.7 | 0.000 | 1.60 |
| filtC1*filtC1 | 1 | 0.00866 | 1.20% | 0.00608 | 0.00608 | 2.6 | 0.112 | 1.60 |
| filtC2*filtC2 | 1 | 0.00598 | 0.83% | 0.00557 | 0.00557 | 2.4 | 0.128 | 2.15 |
| filtC3*filtC3 | 1 | 0.00494 | 0.68% | 0.00451 | 0.00451 | 1.9 | 0.169 | 2.15 |
| activation*activation | 1 | 0.00608 | 0.84% | 0.00608 | 0.00608 | 2.6 | 0.112 | 1.60 |
| Error | 40 | 0.09206 | 12.73% | 0.09206 | 0.00230 | | | |
| Total | 67 | 0.72330 | 100.00% | | | | | |

**Model Summary**

| S | R-sq | R-sq(adj) | PRESS | R-sq(pred) |
|---|---|---|---|---|
| 0.04797 | 0.8727 | 0.7868 | 0.187374 | 0.7409 |

Table 6 – DSD Analysis of Variance (ANOVA) for Test Accuracy

The statistically significant model effects are highlighted in blue, where the $\alpha$ risk for determining statistical significance was 0.05 (applied to the ANOVA P-values for the F-values of the F-distributions). The contribution shows the contribution to variation for each statistical model component using the sequential sums-of-squares. The factor F-values (the ratio of the adjusted sum-of-squares of the effect $\beta$ over the error $\epsilon$ adjusted sums-of-squares) provide an estimate of a factor's power with respect to the total variation observed in the experimental results. Error contributed 12.7% to the total observed variation in the experimental response vector.

The model residual sample standard deviation (S) was 0.05 (or 5.0% test accuracy), which appears large except that the predicted fit of the model (R-sq-(pred)) is approximately 74%, indicating that this selection of model terms is a good predictor of test accuracy. The variance inflation factor (VIF) of the model's linear terms are all near a value

# Automated Copper Alloy Grain Size Evaluation Using a Deep-learning CNN

of 1 while the VIF of model's square terms range between 1.60 and 2.83, showing this model for test accuracy has moderate multicollinearity[21], or correlation, between predictor terms [32].

An abbreviated summary of the DSD ANOVA results for the Group-A[22] experimental responses, including 'Test Accuracy', is provided in Table 7. The statistically significant model factors with $\alpha < 0.05$ are shaded blue, while model factors with $\alpha < 0.10$ are shaded orange. The $\alpha$-risk indicates if we can reject the null-hypothesis with either a 95% (blue) or 90% (orange) confidence. The 90% confidence factors are highlighted for reference purposes.

The time in minutes (to run 35 epochs) response had a very low component of error (3.1%) and 13 of the 16 factors had statistically significant effects. Four continuous factors (*batch*, *dropC1*, *filtC3* and *activation*) had both linear and quadratic components that were significant, while others had either only linear (*dropC2*, *dropD1*, *filtC1*) or only quadratic (*optimize* and *dropC3*) effects that were significant. All five of the categorical factors (*maxpC1*, *maxpC2*, *maxpC3*, *padding* and *strideC1*) effects were statistically significant. The *Batch* factor contributed about 26% (linear + quadratic) of the total variation seen in the time response. The R-squared value was 0.97. The regression model had a predicted sums-of-squares (PRESS)[23] of 0.19 with a residual sample standard deviation (S) of 0.23 minutes.

The training, validation, and test accuracy responses had similar results with $\alpha < 0.05$, with 9 of 16 factors having statistically significant effects. Continuous factors had either linear (*dropC1*, *dropC2*, *dropC3* and *activation*) or quadratic (*constrain*, *dropD1*) effects that were significant, and only one factor had both linear and quadratic effect (*optimize*), while only two of the categorical factors (*maxpC1* and *maxpC3*) had effects that were statistically significant. The continuous factor with the largest contribution to the accuracy responses was *constrain*, which ranged between approximately 12% and 14% for the three responses. The categorical factor with the largest contribution to the accuracy responses was *maxpC3*, which was approximately 19% for all three responses. The component of experimental error for training, validation and test accuracies were 11.40%, 11.37% and 12.73% respectively. The R-squared values for training, validation and test accuracies were 0.886, 0.886 and 0.873 respectively. The regression models had PRESS values of 0.16, 0.17 and 0.19 respectively, with a residual sample standard deviations (S) of 4.27%, 4.34% and 4.60% training, validation and test accuracies respectively.

An abbreviated summary of the DSD ANOVA results for the Group-B[24] experimental responses, including the test accuracy, is provided in Table 8. As before, statistically significant model factors with $\alpha < 0.05$ are shaded blue, while model factors with $\alpha < 0.10$ are shaded orange. The $\alpha$-risk indicates if we can reject the null-hypothesis with either a 95% (blue) or 90% (orange) confidence. The 90% confidence factors are highlighted for reference purposes. The VIF values are identical to those listed in Table 6.

| DSD Experiment ANOVAs: Response Group-A | | | | Blue = $\alpha < 0.05$ (95% confidence), Orange = $\alpha < 0.10$ (90% confidence) | | | | | | | | |
|---|---|---|---|---|---|---|---|---|---|---|---|---|
| | | Time (35 epochs) | | | Training Accuracy (average last 5 epochs) | | | Validation Accuracy (average last 5 epochs) | | | Test Accuracy (after 35 epochs) | | |
| Source | DF | Contribution | F-Value | P-Value | Contribution | F-Value | P-Value | Contribution | F-Value | P-Value | Contribution | F-Value | P-Value |
| Model | 27 | 96.90% | 46.35 | 0.000 | 88.60% | 11.52 | 0.000 | 88.63% | 11.55 | 0.000 | 87.27% | 10.16 | 0.000 |
| Linear | 16 | 62.02% | 50.80 | 0.000 | 54.16% | 12.05 | 0.000 | 55.07% | 12.19 | 0.000 | 56.03% | 11.17 | 0.000 |
| batch | 1 | 4.64% | 154.59 | 0.000 | 0.05% | 2.64 | 0.112 | 0.01% | 1.65 | 0.206 | 0.02% | 1.91 | 0.174 |
| constrain | 1 | 0.00% | 0.21 | 0.653 | 1.51% | 3.60 | 0.065 | 1.38% | 2.47 | 0.124 | 1.21% | 2.74 | 0.106 |
| optimize | 1 | 1.12% | 1.73 | 0.196 | 6.94% | 37.09 | 0.000 | 8.31% | 40.74 | 0.000 | 7.99% | 37.65 | 0.000 |
| dropC1 | 1 | 1.69% | 42.46 | 0.000 | 1.76% | 4.78 | 0.035 | 1.73% | 5.47 | 0.024 | 2.57% | 6.44 | 0.015 |
| dropC2 | 1 | 0.19% | 4.13 | 0.049 | 2.36% | 4.49 | 0.04 | 2.18% | 4.96 | 0.032 | 2.79% | 4.67 | 0.037 |
| dropC3 | 1 | 0.33% | 1.54 | 0.221 | 3.21% | 13.88 | 0.001 | 2.70% | 11.50 | 0.002 | 3.56% | 14.25 | 0.001 |
| dropD1 | 1 | 0.40% | 9.02 | 0.005 | 0.72% | 1.69 | 0.202 | 0.80% | 1.46 | 0.234 | 0.87% | 2.24 | 0.143 |

---

[21] VIF = 1: not correlated, 1 < VIF < 5: moderately correlated; VIF > 5: highly correlated.

[22] See Section 5.1 of this report.

[23] A measure of the deviation between the fitted values and the observed values, where smaller values of PRESS indicate a better predictive model

[24] See Section 5.1.

# Automated Copper Alloy Grain Size Evaluation Using a Deep-learning CNN

**DSD Experiment ANOVAs: Response Group-A**  Blue = α < 0.05 (95% confidence), Orange = α < 0.10 (90% confidence)

| Source | DF | Time (35 epochs) Contribution | F-Value | P-Value | Training Accuracy (average last 5 epochs) Contribution | F-Value | P-Value | Validation Accuracy (average last 5 epochs) Contribution | F-Value | P-Value | Test Accuracy (after 35 epochs) Contribution | F-Value | P-Value |
|---|---|---|---|---|---|---|---|---|---|---|---|---|---|
| maxpC1 | 1 | 17.24% | 107.54 | 0.000 | 6.78% | 19.15 | 0.000 | 7.82% | 23.86 | 0.000 | 5.40% | 12.81 | 0.001 |
| maxpC2 | 1 | 12.13% | 85.32 | 0.000 | 0.05% | 0.72 | 0.400 | 0.00% | 1.11 | 0.299 | 0.01% | 1.29 | 0.262 |
| maxpC3 | 1 | 10.31% | 150.07 | 0.000 | 19.17% | 63.64 | 0.000 | 19.22% | 61.56 | 0.000 | 18.90% | 56.32 | 0.000 |
| filtC1 | 1 | 0.04% | 5.96 | 0.019 | 0.14% | 0.81 | 0.372 | 0.06% | 0.32 | 0.578 | 0.01% | 0.18 | 0.673 |
| filtC2 | 1 | 0.09% | 1.57 | 0.217 | 0.67% | 2.01 | 0.164 | 0.47% | 1.10 | 0.301 | 0.57% | 1.63 | 0.210 |
| filtC3 | 1 | 0.09% | 4.51 | 0.04 | 0.09% | 0.30 | 0.586 | 0.17% | 0.81 | 0.372 | 0.01% | 0.02 | 0.887 |
| padding | 1 | 1.36% | 29.82 | 0.000 | 0.22% | 2.91 | 0.096 | 0.17% | 2.04 | 0.161 | 0.44% | 4.04 | 0.051 |
| strideC1 | 1 | 12.33% | 109.02 | 0.000 | 0.01% | 0.03 | 0.868 | 0.00% | 0.01 | 0.908 | 0.01% | 0.02 | 0.882 |
| activation | 1 | 0.06% | 8.03 | 0.007 | 10.50% | 30.71 | 0.000 | 10.06% | 31.14 | 0.000 | 11.66% | 30.41 | 0.000 |
| Square | 11 | 34.88% | 40.95 | 0.000 | 34.45% | 10.99 | 0.000 | 33.56% | 10.74 | 0.000 | 31.24% | 8.93 | 0.000 |
| batch*batch | 1 | 21.26% | 189.78 | 0.000 | 1.48% | 2.34 | 0.134 | 1.10% | 2.95 | 0.093 | 0.68% | 2.64 | 0.112 |
| constrain*constrain | 1 | 0.00% | 0.85 | 0.362 | 13.62% | 28.72 | 0.000 | 13.08% | 27.66 | 0.000 | 11.97% | 24.3 | 0.000 |
| optimize*optimize | 1 | 0.46% | 4.79 | 0.035 | 5.34% | 20.09 | 0.000 | 6.24% | 24.24 | 0.000 | 5.71% | 21.42 | 0.000 |
| dropC1*dropC1 | 1 | 0.25% | 4.16 | 0.048 | 0.95% | 2.66 | 0.110 | 1.13% | 2.96 | 0.093 | 1.30% | 2.64 | 0.112 |
| dropC2*dropC2 | 1 | 0.08% | 2.29 | 0.138 | 0.64% | 2.61 | 0.114 | 0.83% | 3.17 | 0.083 | 0.89% | 2.64 | 0.112 |
| dropC3*dropC3 | 1 | 0.00% | 36.11 | 0.000 | 0.05% | 0.54 | 0.468 | 0.05% | 0.09 | 0.772 | 0.09% | 0.63 | 0.433 |
| dropD1*dropD1 | 1 | 0.02% | 0.00 | 0.952 | 9.28% | 37.84 | 0.000 | 8.28% | 34.06 | 0.000 | 7.07% | 26.79 | 0.000 |
| filtC1*filtC1 | 1 | 0.05% | 0.26 | 0.613 | 1.14% | 2.88 | 0.098 | 1.41% | 3.84 | 0.057 | 1.20% | 2.64 | 0.112 |
| filtC2*filtC2 | 1 | 0.04% | 0.86 | 0.358 | 0.59% | 1.92 | 0.174 | 0.06% | 0.16 | 0.694 | 0.83% | 2.42 | 0.128 |
| filtC3*filtC3 | 1 | 12.24% | 160.21 | 0.000 | 0.70% | 2.25 | 0.141 | 0.62% | 1.98 | 0.167 | 0.68% | 1.96 | 0.169 |
| activation*activation | 1 | 0.47% | 6.01 | 0.019 | 0.66% | 2.33 | 0.135 | 0.76% | 2.69 | 0.109 | 0.84% | 2.64 | 0.112 |
| Error | 40 | 3.10% | | | 11.40% | | | 11.37% | | | 12.73% | | |
| Total | 67 | 100.00% | | | 100.00% | | | 100.00% | | | 100.00% | | |

| **Model Summaries** | | S | R-sq | PRESS | S | R-sq | PRESS | S | R-sq | PRESS | S | R-sq | PRESS |
|---|---|---|---|---|---|---|---|---|---|---|---|---|---|---|
| | | 0.23359 | 0.9690 | 5.617 | 0.04269 | 0.8860 | 0.15832 | 0.043411 | 0.886 | 0.1648 | 0.04597 | 0.873 | 0.1874 |

Table 7 – DSD Summary of Significant Factors for Group-A Response ANOVAs

The Group-B responses are more detailed evaluations of the information within the confusion matrix, as explained in Section 5.1. The primary observation for this group of response was that the factors selected in the experiment had little influence on the observed experimental results.

For sensitivity (recall or true positive rate), only the quadratic component of the *Batch* factor had statistical significance with α < 0.05. The *stride* and *filtC2* factors had effects with α < 0.10, with P-values of 0.095 and 0.070 respectively. The component of experimental error was 53.4%. The R-squared value for the model was only 0.47. The regression model had a PRESS value of 20.4, with a residual sample standard deviation (S) of 43.5% for true positive rate.

Specificity (true negative rate) had only the quadratic component of the *dropD1* factor with statistical significance with α < 0.05. The *stride*, *batch*, *filtC1*, *filtC3* and *activation* factors had effects with α < 0.10, with P-values of 0.094, 0.079, 0.067, 0.071 and 0.067 respectively. The component of experimental error was 50.5%. The R-squared value for the model was only 0.495. The regression model had a PRESS value of 19.7, with a residual sample standard deviation (S) of 42.8% for true negative rate. The VIF values are identical to those listed in Table 6.

The precision response values are included for reference purposes only. See Section 5.1 for additional information.

**DSD Experiment ANOVAs: Response Group-B**  Blue = α < 0.05 (95% confidence), Orange = α < 0.10 (90% confidence)

| Source | DF | Sensitivity (Recall or True Positive Rate - TPR) Contribution | F-Value | P-Value | Specificity (True Negative Rate - TNR) Contribution | F-Value | P-Value | Precision (Positive Predictive Rate - PPR) Contribution | F-Value | P-Value |
|---|---|---|---|---|---|---|---|---|---|---|
| Model | 27 | 46.58% | 1.29 | 0.227 | 49.53% | 1.45 | 0.138 | 53.23% | 1.69 | 0.065 |
| Linear | 16 | 18.63% | 0.83 | 0.649 | 14.14% | 0.64 | 0.835 | 35.41% | 1.80 | 0.067 |
| batch | 1 | 0.06% | 0.02 | 0.877 | 0.09% | 0.22 | 0.640 | 0.05% | 0.01 | 0.942 |
| constrain | 1 | 2.24% | 1.85 | 0.182 | 0.98% | 1.02 | 0.318 | 2.01% | 1.34 | 0.253 |
| optimize | 1 | 0.03% | 0.16 | 0.690 | 1.22% | 0.93 | 0.340 | 2.98% | 3.92 | 0.055 |
| dropC1 | 1 | 0.14% | 0.19 | 0.662 | 1.18% | 1.03 | 0.315 | 0.01% | 0.00 | 0.947 |
| dropC2 | 1 | 2.17% | 1.44 | 0.237 | 0.50% | 0.54 | 0.465 | 2.01% | 1.24 | 0.272 |

## Automated Copper Alloy Grain Size Evaluation Using a Deep-learning CNN

**DSD Experiment ANOVAs: Response Group-B**  Blue = α < 0.05 (95% confidence), Orange = α < 0.10 (90% confidence)

| Source | DF | Sensitivity (Recall or True Positive Rate - TPR) | | | Specificity (True Negative Rate - TNR) | | | Precision (Positive Predictive Rate - PPR) | | |
|---|---|---|---|---|---|---|---|---|---|---|
| | | Contribution | F-Value | P-Value | Contribution | F-Value | P-Value | Contribution | F-Value | P-Value |
| dropC3 | 1 | 2.10% | 2.02 | 0.163 | 0.35% | 0.36 | 0.552 | 3.41% | 3.84 | 0.057 |
| dropD1 | 1 | 4.31% | 2.66 | 0.111 | 2.66% | 1.75 | 0.194 | 6.17% | 4.14 | 0.049 |
| maxpC1 | 1 | 0.02% | 0.02 | 0.889 | 0.84% | 0.45 | 0.507 | 1.21% | 0.85 | 0.361 |
| maxpC2 | 1 | 0.01% | 0.00 | 0.974 | 0.00% | 0.04 | 0.833 | 0.00% | 0.14 | 0.708 |
| maxpC3 | 1 | 1.60% | 1.28 | 0.265 | 0.49% | 0.29 | 0.592 | 11.75% | 9.32 | 0.004 |
| filtC1 | 1 | 0.54% | 0.33 | 0.570 | 0.45% | 0.24 | 0.629 | 0.24% | 0.15 | 0.698 |
| filtC2 | 1 | 0.41% | 0.11 | 0.737 | 0.09% | 0.00 | 0.953 | 1.05% | 0.58 | 0.450 |
| filtC3 | 1 | 0.47% | 0.29 | 0.592 | 0.52% | 0.34 | 0.562 | 0.41% | 0.26 | 0.615 |
| padding | 1 | 0.01% | 0.01 | 0.908 | 0.16% | 0.12 | 0.736 | 0.31% | 0.47 | 0.498 |
| strideC1 | 1 | 4.08% | 2.92 | 0.095 | 3.87% | 2.93 | 0.094 | 1.46% | 1.03 | 0.315 |
| activation | 1 | 0.45% | 0.28 | 0.597 | 0.74% | 0.48 | 0.491 | 2.33% | 1.88 | 0.178 |
| Square | 11 | 27.96% | 1.90 | 0.068 | 35.39% | 2.55 | 0.015 | 17.82% | 1.39 | 0.217 |
| batch*batch | 1 | 7.11% | 4.52 | 0.040 | 9.22% | 3.24 | 0.079 | 1.76% | 4.73 | 0.036 |
| constrain*constrain | 1 | 1.71% | 1.26 | 0.268 | 0.08% | 0.00 | 0.976 | 8.19% | 4.51 | 0.040 |
| optimize*optimize | 1 | 0.00% | 0.01 | 0.910 | 1.20% | 0.86 | 0.360 | 0.92% | 0.75 | 0.391 |
| dropC1*dropC1 | 1 | 1.07% | 0.84 | 0.364 | 0.28% | 0.33 | 0.569 | 1.19% | 1.27 | 0.267 |
| dropC2*dropC2 | 1 | 0.13% | 0.08 | 0.774 | 0.59% | 0.43 | 0.518 | 0.01% | 0.01 | 0.938 |
| dropC3*dropC3 | 1 | 2.85% | 1.29 | 0.263 | 2.39% | 1.78 | 0.190 | 0.34% | 0.28 | 0.597 |
| dropD1*dropD1 | 1 | 0.70% | 1.11 | 0.298 | 4.08% | 5.00 | 0.031 | 0.02% | 0.01 | 0.925 |
| filtC1*filtC1 | 1 | 3.65% | 2.24 | 0.142 | 5.61% | 3.53 | 0.067 | 1.36% | 0.94 | 0.337 |
| filtC2*filtC2 | 1 | 4.40% | 3.47 | 0.070 | 2.79% | 2.40 | 0.130 | 1.56% | 1.41 | 0.243 |
| filtC3*filtC3 | 1 | 3.34% | 2.32 | 0.136 | 4.70% | 3.44 | 0.071 | 1.36% | 1.08 | 0.304 |
| activation*activation | 1 | 2.99% | 2.24 | 0.142 | 4.46% | 3.53 | 0.067 | 1.10% | 0.94 | 0.337 |
| Error | 40 | 53.42% | | | 50.47% | | | 46.77% | | |
| Total | 67 | 100.00% | | | 100.00% | | | 100.00% | | |

| Model Summaries | | S | R-sq | PRESS | S | R-sq | PRESS | S | R-sq | PRESS |
|---|---|---|---|---|---|---|---|---|---|---|
| | | 0.43494 | 0.4658 | 20.357 | 0.42815 | 0.4953 | 19.676 | 0.25034 | 0.5323 | 6.6103 |

Table 8 – DSD Summary of Significant Factors for Group-B Response ANOVAs

The factorial plots for the four DSD Group-A responses are provided in Figure 18, Figure 19, Figure 20 and Figure 21. The main effects from the Table 7 statistical models are plotted.

Inspection of the factorial plots allows the factor trends with level settings to be evaluated, which can be used for follow-on parameter experimentation. The plots as shown were rendered by the *Minitab* application. See Section 8.3 for additional information about *Minitab*.

The time (to run 35 epochs) response was affected by the *batch*, *dropC3* and *filtC3* factor quadratic effects. The center *batch* size setting of 160-samples increased the mean training time by over 1 minute more than the 80-samples *batch* size, and by almost 2 minutes over the 240-sample *batch* size. The drops on the 10% center level of convolutional layer-3 reduced the average training time by around 1 minute when compared to the 0% and 20% extreme settings. A convolutional layer-3 *filter* size center setting of '5' increased mean training time by nearly 1½ minutes over the extreme settings of '3' and '7'. *Max-pooling* for all convolution layers reduced mean training time by about 1 minute for each factor. *Padding* when set on all convolutional layers decreased training time by an average of about 30 seconds. Setting the *stride* for convolutional input layers 1 to 2 reduced mean training time by approximately 1 minute. The time response metric would be useful for minimizing resources expended while training the CNN.

**Automated Copper Alloy Grain Size Evaluation Using a Deep-learning CNN**

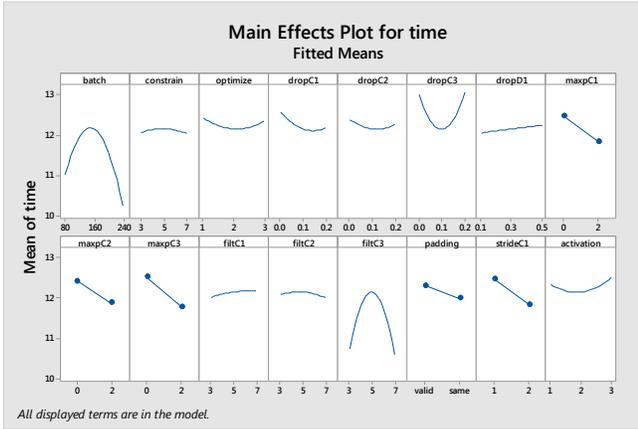

Figure 18 – Factorial Plots: Time to Run 35 Epochs

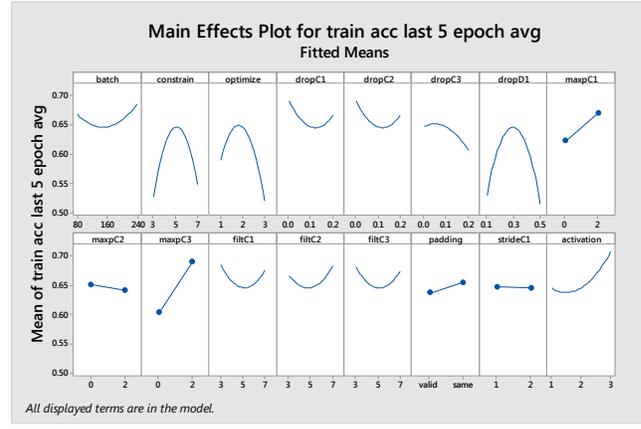

Figure 19 – Factorial Plots: Training Accuracy

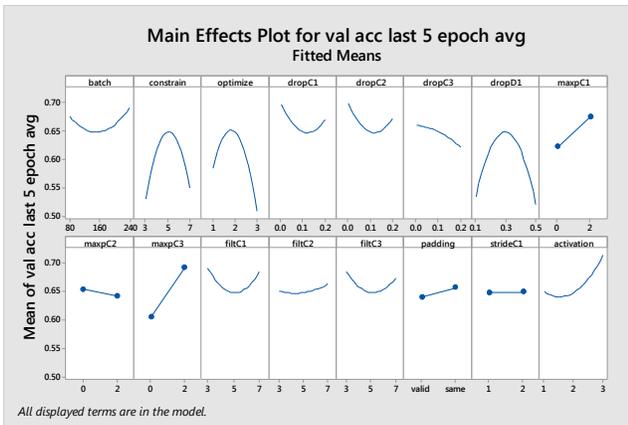

Figure 20 – Factorial Plots: Validation Accuracy

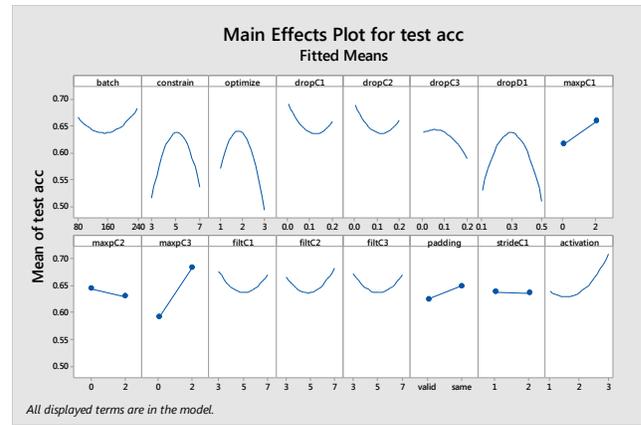

Figure 21 – Factorial Plots: Test Accuracy

All three accuracy responses exhibited nearly identical trends, with the center level settings for the *constraint*, *optimizer* and *drops* on dense layer-1 lifting accuracies approximately 10% over the extreme level settings. *Drops* of 20% on the convolutional layer-3 reduced all three accuracies by about 5% when compared to no *drops*. The center *filter* setting of 5 for convolutional layer-1 reduced accuracies by about 5% when compared to the 0% setting. The *drops* for convolutional layer-2 exhibited a downward linear reduction in test accuracy, but also with a quadratic component that was inverted (concave upward). *Max-pooling* of 2 on both convolutional layers 1 and 3 lifted accuracies around 5% for layer-1 and 10% for layer-3 on the average. The ReLU *activation* tended to increase accuracies from 5% to 10% when compared to the other activators. The quadratic component for *batch* size was inverted (concave upward) with the center setting of 160-samples dropping accuracies by around 3% when compared to the extreme settings.

The factorial plots for the three DSD Group-B responses are provided in Figure 22, Figure 23 and Figure 24. The main effects from the Table 8 statistical models are plotted. These plots are provided for reference purposes because of the low predictability of the Table 8 statistical models. Some comments about these response Group-B factorial plots follow:

- The sensitivity (true positive rate) and precision (positive predictive rate) exhibit the same general trends.
- For sensitivity, only the *batch* factor was statistically significant.
- For specificity, only the *dropD1* factor was statistically significant.

- Trends for specificity (true negative rate) are opposite from those for sensitivity (true positive rate) and precision (positive predictive rate).
- Specificity has trends that exceed 1.0, which is another indicator that the model has little or no predictive value.

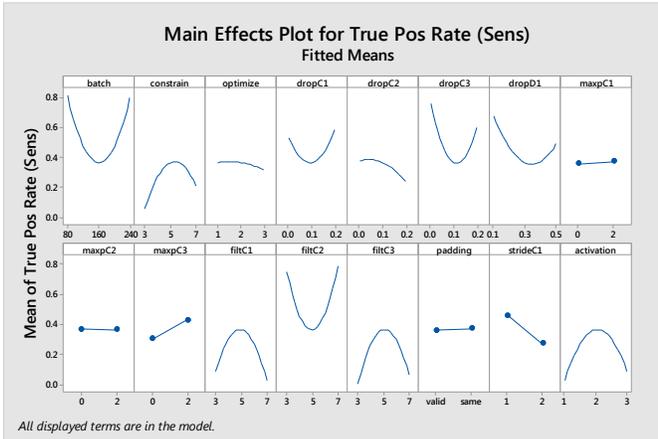

Figure 22 – Factorial Plots: Sensitivity (TPR)

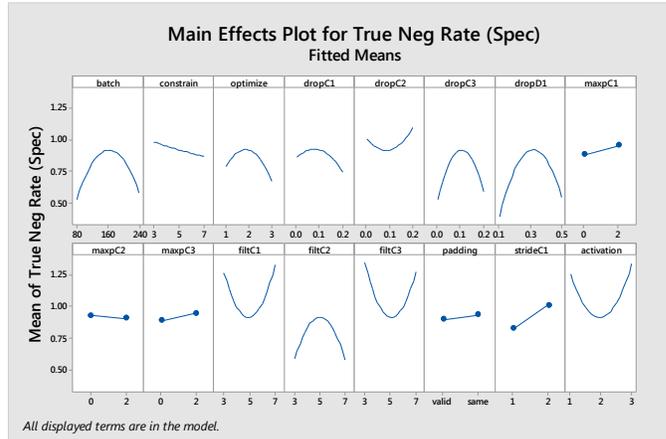

Figure 23 – Factorial Plots: Specificity (TNR)

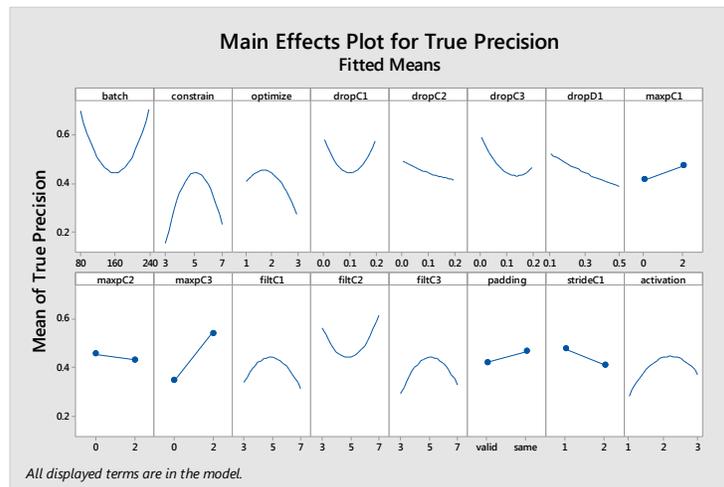

Figure 24 – Factorial Plots: Precision (PPR)

### 5.2.4. DSD Verification Check

The best CNN configuration for test accuracy from the DSD experiments was configured and run through 250 epochs of training on the existing data set. The dropouts for all three convolutional layers were set to zero.

| batch | constrain | optimizer | dropC1 | dropC2 | dropC3 | dropD1 | maxpC1 |
|-------|-----------|-----------|--------|--------|--------|--------|--------|
| 240 | 5 | 2 = Adamax | 0 | 0 | 0 | 0.3 | 2 |
| **maxpC2** | **maxpC3** | **filtC1** | **filtC2** | **filtC3** | **padding** | **strideC1** | **activation** |
| 0 | 2 | 3 | 3 | 3 | same | 2 | 3 = ReLU |

Table 9 – CNN configuration for DSD Verification Check

## Automated Copper Alloy Grain Size Evaluation Using a Deep-learning CNN

| Run | Training Plots | Confusion Matrix Metrics | | | |
|-----|---------------|--------------------------|---|---|---|

**Run 1**

Treatment 1, CNN model accuracy and loss

| | **Predicted Class Labels** | | |
|---|---|---|---|
| **Class** | Bad Grains | Good Grains | True Rates |
| **Actual Class Labels** | | | |
| Bad Grains | TN = 296 | FP = 104 | Specificity = 74.00% |
| Good Grains | FN =   9 | TP = 391 | Sensitivity = 97.75% |
| | False Positive Rate = 26.00% | Precision = 78.99% | Accuracy = 85.88% |
| Train. Acc. (L5): 88.21% | | | |
| Val. Acc. (L5): 90.76% | | | Time: 56.50 min. |

**Run 2**

Treatment 2, CNN model accuracy and loss

| | **Predicted Class Labels** | | |
|---|---|---|---|
| **Class** | Bad Grains | Good Grains | True Rates |
| **Actual Class Labels** | | | |
| Bad Grains | TN = 374 | FP =   26 | Specificity = 93.50% |
| Good Grains | FN =   46 | TP = 354 | Sensitivity = 88.50% |
| | False Positive Rate = 6.50% | Precision = 93.16% | Accuracy = 91.00% |
| Train. Acc. (L5): 88.82% | | | |
| Val. Acc. (L5): 90.74% | | | Time: 56.50 min. |

**Run 3**

Treatment 3, CNN model accuracy and loss

| | **Predicted Class Labels** | | |
|---|---|---|---|
| **Class** | Bad Grains | Good Grains | True Rates |
| **Actual Class Labels** | | | |
| Bad Grains | TN = 332 | FP =   68 | Specificity = 83.00% |
| Good Grains | FN =   14 | TP = 386 | Sensitivity = 96.50% |
| | False Positive Rate = 17.00% | Precision = 85.02% | Accuracy = 89.75% |
| Train. Acc. (L5): 89.32% | | | |
| Val. Acc. (L5): 91.42% | | | Time: 56.55 min. |

**Run 4**

Treatment 4, CNN model accuracy and loss

| | **Predicted Class Labels** | | |
|---|---|---|---|
| **Class** | Bad Grains | Good Grains | True Rates |
| **Actual Class Labels** | | | |
| Bad Grains | TN = 291 | FP = 109 | Specificity = 72.75% |
| Good Grains | FN =   7 | TP = 393 | Sensitivity = 98.25% |
| | False Positive Rate = 27.25% | Precision = 78.29% | Accuracy = 85.50% |
| Train. Acc. (L5): 88.35% | | | |
| Val. Acc. (L5): 90.01% | | | Time: 56.55 min. |

**Run 5**

Treatment 5, CNN model accuracy and loss

| | **Predicted Class Labels** | | |
|---|---|---|---|
| **Class** | Bad Grains | Good Grains | True Rates |
| **Actual Class Labels** | | | |
| Bad Grains | TN = 331 | FP =   69 | Specificity = 82.75% |
| Good Grains | FN =   15 | TP = 385 | Sensitivity = 96.25% |
| | False Positive Rate = 17.25% | Precision = 84.80% | Accuracy = 89.50% |
| Train. Acc. (L5): 90.02% | | | |
| Val. Acc. (L5): 92.29% | | | Time: 56.57 min. |

Table 10 – DSD Verification Tests Results

The DSD-configured CNN successfully trained for all 5 test runs, with an average test accuracy of 88.3% after 250 epochs of training, with no evidence of overfitting (orange validation plots consistently higher than the blue training plots, both accuracy plots constantly increasing, and both loss plots constantly decreasing throughout all the training epochs). The average specificity at 81.2% was significantly lower than average sensitivity at 95.5%, indicating the CNN favored the classification of good over bad grain structure (the average false positive rate was 18.8%).

**Automated Copper Alloy Grain Size Evaluation Using a Deep-learning CNN**

## 5.3.    Optimization Experiments

### 5.3.1.        CCD Factors and Levels

CCD optimization experiments are much more resource-hungry than DSD screening experiments; therefore, only a limited number of factors may be studied before incurring significant computational penalties in the form of experimental treatment combinations. Two continuous factors (kernel *Constraint* and *Drops Dense 1*) and one semi-continuous factor (*Optimizer*) with significant response effects from the DSD screening experiments were selected for further study. The CCD factors, class (training or CNN layer), type (continuous or categorical) and levels (-α, 1, 2, 3 and +α)[25] are shown in Table 11.

The other factors were primarily set to the DSD configuration for highest test accuracy; achieving lower computational time was considered for factors that had exhibited no real impact on accuracy, but had effects on the time. The *Batch* size was the only factor that met the secondary criteria, with the 240 *Batch* size setting offering a reduction of 1 to 2 minutes per DSD experimental treatment combination epoch.

| Factor Name | Factor Class | Factor Type | - Star (α) | Level 1 | Level 2 | Level 3 | + Star (α) |
|---|---|---|---|---|---|---|---|
| 1. Kernel Constraint | Training | Continuous | 2.17157 | 3 | 5 | 7 | 7.82843 |
| 2. Optimizer | Training | Categorical | | Adam | AdaMax | | |
| 3. Drops Dense1 | Training | Continuous | 1.716% | 10% | 30% | 50% | 58.284% |

Table 11 – CCD Experiment Factor Information

Together, *Kernel Constraint*, *DropD1(drops dense 1),* and *Optimizer* contributed 37.4%, 38.1%, and 34.8% respectively to the variation observed in the training, validation and test accuracies during the DSD experiments. The star-points would open the experimental range for the two continuous factors up by over 60%, and because it was not possible to set a star-point level for the *Optimizer* factor momentum, only the *Adam* and *Adamax* settings were explored, and the *Optimizer* was configured as a categorical CCD factor. The *Kernel Constraint* and *DropD1* would allow any interactions between these two factors to be studied in greater detail, and because both factors can affect the level of CNN over-fit the training data, there was considerable interest in exploring this potential interaction.

### 5.3.2.        CCD Experimental Matrix

The CCD experiment required 26 treatment combinations as shown in a matrix in Table 12.  CCD factor selections and level settings were based on the outcome of the DSD screening experiment. CCC-configured star-points were configured to provide capability for rotation of the experimental matrix. The star-points for the *Kernel Constraint* and *DropD1* factors appear as fractional values for these factors. *Optimizer* was treated as a categorical factor for the CCD. The factors that were held at a fixed level are greyed.

| TC | Batch | Kernel Constraint | Optimizer | Drop C1 | Drop C2 | Drop C3 | Drop D1 | MaxP C1 | MaxP C2 | MaxP C3 | Filt C1 | Filt C2 | Filt C3 | Padd | Stride C1 | Active C1 |
|---|---|---|---|---|---|---|---|---|---|---|---|---|---|---|---|---|
| 1 | 240 | 3 | Adam | 0 | 0 | 0 | 0.1 | 2 | 2 | 2 | 3 | 3 | 3 | Same | 2 | ReLU |
| 2 | 240 | 7 | Adam | 0 | 0 | 0 | 0.1 | 2 | 2 | 2 | 3 | 3 | 3 | Same | 2 | ReLU |
| 3 | 240 | 3 | Adam | 0 | 0 | 0 | 0.5 | 2 | 2 | 2 | 3 | 3 | 3 | Same | 2 | ReLU |
| 4 | 240 | 7 | Adam | 0 | 0 | 0 | 0.5 | 2 | 2 | 2 | 3 | 3 | 3 | Same | 2 | ReLU |
| 5 | 240 | 2.17157 | Adam | 0 | 0 | 0 | 0.3 | 2 | 2 | 2 | 3 | 3 | 3 | Same | 2 | ReLU |
| 6 | 240 | 7.82843 | Adam | 0 | 0 | 0 | 0.3 | 2 | 2 | 2 | 3 | 3 | 3 | Same | 2 | ReLU |

---

[25] Note that the CCD ±α star-points have no relationship to the α (alpha) risk used in the ANOVA

**Automated Copper Alloy Grain Size Evaluation Using a Deep-learning CNN**

| TC | Batch | Kernel Constraint | Optimizer | Drop C1 | Drop C2 | Drop C3 | Drop D1 | MaxP C1 | MaxP C2 | MaxP C3 | Filt C1 | Filt C2 | Filt C3 | Padd | Stride C1 | Active |
|----|-------|------------------|-----------|---------|---------|---------|---------|---------|---------|---------|---------|---------|---------|------|-----------|--------|
| 7 | 240 | 5 | Adam | 0 | 0 | 0 | 0.01716 | 2 | 2 | 2 | 3 | 3 | 3 | Same | 2 | ReLU |
| 8 | 240 | 5 | Adam | 0 | 0 | 0 | 0.58284 | 2 | 2 | 2 | 3 | 3 | 3 | Same | 2 | ReLU |
| 9 | 240 | 5 | Adam | 0 | 0 | 0 | 0.3 | 2 | 2 | 2 | 3 | 3 | 3 | Same | 2 | ReLU |
| 10 | 240 | 5 | Adam | 0 | 0 | 0 | 0.3 | 2 | 2 | 2 | 3 | 3 | 3 | Same | 2 | ReLU |
| 11 | 240 | 5 | Adam | 0 | 0 | 0 | 0.3 | 2 | 2 | 2 | 3 | 3 | 3 | Same | 2 | ReLU |
| 12 | 240 | 5 | Adam | 0 | 0 | 0 | 0.3 | 2 | 2 | 2 | 3 | 3 | 3 | Same | 2 | ReLU |
| 13 | 240 | 5 | Adam | 0 | 0 | 0 | 0.3 | 2 | 2 | 2 | 3 | 3 | 3 | Same | 2 | ReLU |
| 14 | 240 | 3 | Adamax | 0 | 0 | 0 | 0.1 | 2 | 2 | 2 | 3 | 3 | 3 | Same | 2 | ReLU |
| 15 | 240 | 7 | Adamax | 0 | 0 | 0 | 0.1 | 2 | 2 | 2 | 3 | 3 | 3 | Same | 2 | ReLU |
| 16 | 240 | 3 | Adamax | 0 | 0 | 0 | 0.5 | 2 | 2 | 2 | 3 | 3 | 3 | Same | 2 | ReLU |
| 17 | 240 | 7 | Adamax | 0 | 0 | 0 | 0.5 | 2 | 2 | 2 | 3 | 3 | 3 | Same | 2 | ReLU |
| 18 | 240 | 2.17157 | Adamax | 0 | 0 | 0 | 0.3 | 2 | 2 | 2 | 3 | 3 | 3 | Same | 2 | ReLU |
| 19 | 240 | 7.82843 | Adamax | 0 | 0 | 0 | 0.3 | 2 | 2 | 2 | 3 | 3 | 3 | Same | 2 | ReLU |
| 20 | 240 | 5 | Adamax | 0 | 0 | 0 | 0.01716 | 2 | 2 | 2 | 3 | 3 | 3 | Same | 2 | ReLU |
| 21 | 240 | 5 | Adamax | 0 | 0 | 0 | 0.58284 | 2 | 2 | 2 | 3 | 3 | 3 | Same | 2 | ReLU |
| 22 | 240 | 5 | Adamax | 0 | 0 | 0 | 0.3 | 2 | 2 | 2 | 3 | 3 | 3 | Same | 2 | ReLU |
| 23 | 240 | 5 | Adamax | 0 | 0 | 0 | 0.3 | 2 | 2 | 2 | 3 | 3 | 3 | Same | 2 | ReLU |
| 24 | 240 | 5 | Adamax | 0 | 0 | 0 | 0.3 | 2 | 2 | 2 | 3 | 3 | 3 | Same | 2 | ReLU |
| 25 | 240 | 5 | Adamax | 0 | 0 | 0 | 0.3 | 2 | 2 | 2 | 3 | 3 | 3 | Same | 2 | ReLU |
| 26 | 240 | 5 | Adamax | 0 | 0 | 0 | 0.3 | 2 | 2 | 2 | 3 | 3 | 3 | Same | 2 | ReLU |

Table 12 – CCD Experimental Matrix

### 5.3.3.    CCD Experimental Results

Each treatment combination in the matrix was iterated through 100 training epochs, where for each epoch the CNN model is subjected to one forward pass and one backward pass for all training images. At the end of each epoch, the CNN model attempted to classify the validation image samples (not seen during the training iterations). Once all 100 epochs were completed, the trained CNN model weights with the highest validation accuracy (see Footnote 14) from the 100 epochs was shown the test image samples; the test image samples were not part of either the training or validation sets used during the epoch training iterations.

The responses for the 100 treatments combinations are given in Table 13. The training and validation accuracies represent the average values of the last 5 epochs (95 to 100) and test accuracy represented the overall test accuracy achieved on the test data set.  All of the treatment combinations were able to learn the training data.

| TC | EPOCH | First Experiment | | | | | | | Replicated Experiment | | | | | | |
|----|-------|------|---------|---------|---------|-------------|-------------|-----------|------|---------|---------|---------|-------------|-------------|-----------|
| | | Time | TRN ACC | VAL ACC | TST ACC | Specificity | Sensitivity | Precision | Time | TRN ACC | VAL ACC | TST ACC | Specificity | Sensitivity | Precision |
| 1 | 100 | 22.917 | 0.80632 | 0.84205 | 0.83375 | 0.6975 | 0.9700 | 0.7623 | 22.850 | 0.80908 | 0.81333 | 0.8538 | 0.7725 | 0.9350 | 0.8043 |
| 2 | 100 | 23.050 | 0.82084 | 0.86280 | 0.89250 | 0.8675 | 0.9175 | 0.8738 | 22.850 | 0.81460 | 0.82097 | 0.8800 | 0.8625 | 0.8975 | 0.8672 |
| 3 | 100 | 22.817 | 0.79586 | 0.84155 | 0.86375 | 0.8425 | 0.8850 | 0.8489 | 22.833 | 0.79398 | 0.80569 | 0.8438 | 0.8775 | 0.8100 | 0.8686 |
| 4 | 100 | 22.767 | 0.80908 | 0.83540 | 0.82750 | 0.6850 | 0.9700 | 0.7549 | 22.850 | 0.79251 | 0.79264 | 0.8563 | 0.7475 | 0.9650 | 0.7926 |
| 5 | 100 | 22.750 | 0.79837 | 0.81983 | 0.85125 | 0.8500 | 0.8525 | 0.8504 | 23.783 | 0.80209 | 0.80306 | 0.8525 | 0.7775 | 0.9275 | 0.8065 |
| 6 | 100 | 22.767 | 0.79883 | 0.84356 | 0.86125 | 0.8150 | 0.9075 | 0.8307 | 22.883 | 0.80908 | 0.80653 | 0.8625 | 0.7700 | 0.9550 | 0.8059 |
| 7 | 100 | 22.750 | 0.81561 | 0.84877 | 0.82875 | 0.9575 | 0.7000 | 0.9428 | 22.717 | 0.82473 | 0.84306 | 0.8838 | 0.8200 | 0.9475 | 0.8404 |
| 8 | 100 | 22.733 | 0.74251 | 0.78310 | 0.77000 | 0.8975 | 0.6425 | 0.8624 | 22.667 | 0.78988 | 0.80833 | 0.8375 | 0.7225 | 0.9525 | 0.7744 |
| 9 | 100 | 22.767 | 0.80887 | 0.84314 | 0.85750 | 0.7550 | 0.9600 | 0.7967 | 22.667 | 0.82707 | 0.82986 | 0.8738 | 0.8150 | 0.9325 | 0.8345 |
| 10 | 100 | 22.850 | 0.82100 | 0.84285 | 0.83000 | 0.6825 | 0.9775 | 0.7548 | 22.683 | 0.79121 | 0.78944 | 0.8363 | 0.7700 | 0.9025 | 0.7969 |
| 11 | 100 | 22.817 | 0.80711 | 0.80889 | 0.85750 | 0.7575 | 0.9575 | 0.7979 | 22.633 | 0.79674 | 0.80792 | 0.8513 | 0.7700 | 0.9325 | 0.8022 |
| 12 | 100 | 22.750 | 0.81276 | 0.80139 | 0.87500 | 0.7950 | 0.9550 | 0.8233 | 22.717 | 0.80695 | 0.81181 | 0.8800 | 0.9375 | 0.8225 | 0.9294 |
| 13 | 100 | 22.850 | 0.80481 | 0.82181 | 0.83500 | 0.8375 | 0.8325 | 0.8367 | 22.683 | 0.79590 | 0.81153 | 0.8538 | 0.7400 | 0.9675 | 0.7882 |
| 14 | 100 | 22.983 | 0.80251 | 0.79097 | 0.82750 | 0.6675 | 0.9875 | 0.7481 | 22.750 | 0.76368 | 0.76222 | 0.8213 | 0.6825 | 0.9600 | 0.7515 |
| 15 | 100 | 22.900 | 0.80444 | 0.81667 | 0.85500 | 0.7700 | 0.9400 | 0.8034 | 22.683 | 0.80933 | 0.82556 | 0.8500 | 0.7800 | 0.9200 | 0.8070 |

## Automated Copper Alloy Grain Size Evaluation Using a Deep-learning CNN

| TC | EPOCH | First Experiment | | | | | | | Replicated Experiment | | | | | | |
|----|-------|------|---------|---------|---------|-------------|-------------|-----------|------|---------|---------|---------|-------------|-------------|-----------|
|    |       | Time | TRN ACC | VAL ACC | TST ACC | Specificity | Sensitivity | Precision | Time | TRN ACC | VAL ACC | TST ACC | Specificity | Sensitivity | Precision |
| 16 | 100 | 22.850 | 0.80699 | 0.82389 | 0.85875 | 0.8125 | 0.9050 | 0.8284 | 22.683 | 0.80699 | 0.82389 | 0.8588 | 0.8125 | 0.9050 | 0.8284 |
| 17 | 100 | 23.033 | 0.80536 | 0.80292 | 0.84250 | 0.8925 | 0.7925 | 0.8806 | 22.700 | 0.79431 | 0.80069 | 0.8575 | 0.7825 | 0.9325 | 0.8109 |
| 18 | 100 | 22.933 | 0.80000 | 0.82319 | 0.82375 | 0.6825 | 0.9650 | 0.7524 | 22.717 | 0.79226 | 0.81208 | 0.8413 | 0.7450 | 0.9375 | 0.7862 |
| 19 | 100 | 22.800 | 0.81364 | 0.82875 | 0.87500 | 0.8250 | 0.9250 | 0.8409 | 22.683 | 0.79829 | 0.80625 | 0.8625 | 0.8175 | 0.9075 | 0.8326 |
| 20 | 100 | 22.867 | 0.80845 | 0.81736 | 0.86875 | 0.7950 | 0.9425 | 0.8214 | 22.667 | 0.81331 | 0.81556 | 0.8713 | 0.8275 | 0.9150 | 0.8414 |
| 21 | 100 | 22.967 | 0.78372 | 0.78611 | 0.82750 | 0.6950 | 0.9600 | 0.7589 | 22.733 | 0.79326 | 0.80306 | 0.8338 | 0.7200 | 0.9475 | 0.7719 |
| 22 | 100 | 22.767 | 0.79385 | 0.81125 | 0.85500 | 0.8100 | 0.9000 | 0.8257 | 22.767 | 0.79385 | 0.81125 | 0.8550 | 0.8100 | 0.9000 | 0.8257 |
| 23 | 100 | 22.850 | 0.79983 | 0.80792 | 0.85625 | 0.7800 | 0.9325 | 0.8091 | 22.667 | 0.80180 | 0.78306 | 0.8450 | 0.8450 | 0.8450 | 0.8450 |
| 24 | 100 | 22.830 | 0.80678 | 0.81708 | 0.79750 | 0.6100 | 0.9850 | 0.7164 | 22.767 | 0.78962 | 0.79972 | 0.8525 | 0.7500 | 0.9550 | 0.7925 |
| 25 | 100 | 23.550 | 0.80678 | 0.81861 | 0.84875 | 0.8175 | 0.8800 | 0.8282 | 22.700 | 0.79347 | 0.82042 | 0.8400 | 0.8400 | 0.8400 | 0.8400 |
| 26 | 100 | 23.000 | 0.80749 | 0.82778 | 0.85625 | 0.7625 | 0.9500 | 0.8000 | 22.783 | 0.80481 | 0.82903 | 0.8163 | 0.6700 | 0.9625 | 0.7447 |

Table 13 – CCD Response Matrix

The experiment required approximately 10 hours to conduct all 26 treatments and 100 epochs for each experiment; the total experimental run-time in minutes is equal to the sum of the time column responses.

ANOVAs for the statistical models of the CCD experiment's Group-A and B output responses are given in Table 15, Table 16 and Table 17. As before, statistically-significant model factors with $\alpha < 0.05$ are shaded blue, while model factors with $\alpha < 0.10$ are shaded orange. The $\alpha$-risk indicates if we can reject the null-hypothesis with either a 95% (blue) or 90% (orange) confidence. The 90% confidence factors are highlighted for reference purposes. The VIF values indicate that no multicollinearity was present.

The experimental error for the responses (training accuracy, validation accuracy, test accuracy, specificity, sensitivity, and precision) ranged from 61.8% to 78.2% and the pure error component (reproducibility or results) accounted for approximately 50% of all the variation observed in the output response vector. This demonstrates the stochastic nature of CNN training and shows why a large data sample of outcomes is necessary to properly estimate a change in CNN performance between one configuration and another. As shown in Table 14, the sample standard deviation (S) for accuracy percentages of the experimental (lack-of-fit) residuals ranges from 1.2% and 11.7% (taken from the Model Summary for each response ANOVA). With 52 total treatments, the standard error of the mean accuracy percentages ranged from 0.2% and 1.6%. The three CCD factors injected less variation into the experimental output when compared to the DSD, and the DSD had a majority of treatments that did not train, producing a much wider variation in the experimental results than those for the CCD. This is why the R-squared values for the statistical models for the CCD experiment are lower than those for the DSD experiment. Also note that the error variation observed for the Group-B responses (specificity, sensitivity, and precision) was much higher than that of the Group-A responses (training, validation, and test accuracies), indicating that confusion matrix effects are more difficult to correlate with experimental factors than the bulk accuracy measurements.

| Response | SE (mean) | S (sample SD) | n (samples) |
|----------|-----------|---------------|-------------|
| Training Accuracy | 0.167% | 1.202% | 52 |
| Validation Accuracy | 0.245% | 1.765% | 52 |
| Test Accuracy | 0.271% | 1.953% | 52 |
| Specificity | 0.943% | 6.802% | 52 |
| Sensitivity | 1.616% | 11.651% | 52 |
| Precision | 0.566% | 4.085% | 52 |

Table 14 – Experimental Standard Error of the Means

For training accuracy, the *Constraint* main effect and the *Constraint×Optimizer* two-factor interaction effect were statistically significant with high F-values and corresponding P-values < 0.05, while the *Constraint×DropD1* interaction effect P-value was < 0.10. Together these three effects contributed to 32.8% of the observed variation in the experimental output.

## Automated Copper Alloy Grain Size Evaluation Using a Deep-learning CNN

For validation accuracy, the *Constraint×Optimizer* and *Constraint×DropD1* two-factor interaction effects were statistically significant with high F-values and corresponding P-values < 0.05, while the *Constraint* and *DropD1* main effect P-values were < 0.10. Together, these four effects contributed to 33.9% of the observed variation in the experimental output.

**Analysis of Variance (ANOVA)**

| Source | DF | Training Accuracy | | | | | | Validation Accuracy | | | | | |
|---|---|---|---|---|---|---|---|---|---|---|---|---|---|
| | | Adj. SS | Contrib. | Adj. MS | F-Val | P-Val | VIF | Adj. SS | Contrib. | Adj. MS | F-Val | P-Val | VIF |
| Model | 8 | 0.00384 | 38.19% | 0.00048 | 3.32 | 0.005 | | 0.007010 | 34.35% | 0.000876 | 2.81 | 0.013 | |
| Linear | 3 | 0.00195 | 19.38% | 0.00065 | 4.49 | 0.008 | | 0.002181 | 10.69% | 0.000727 | 2.33 | 0.087 | |
| optimizer | 1 | 0.00023 | 2.33% | 0.00023 | 1.62 | 0.210 | 1.00 | 0.000102 | 0.50% | 0.000102 | 0.33 | 0.569 | 1.00 |
| constrain | 1 | 0.00158 | 15.70% | 0.00158 | 10.93 | 0.002 | 1.00 | 0.000978 | 4.79% | 0.000978 | 3.14 | 0.083 | 1.00 |
| dropD1 | 1 | 0.00014 | 1.35% | 0.00014 | 0.94 | 0.338 | 1.00 | 0.001100 | 5.39% | 0.001100 | 3.53 | 0.067 | 1.00 |
| Square | 2 | 0.00018 | 1.75% | 0.00009 | 0.61 | 0.549 | | 0.000154 | 0.75% | 0.000077 | 0.25 | 0.782 | |
| constrain*constrain | 1 | 0.00000 | 0.00% | 0.00000 | 0.00 | 0.968 | 1.02 | 0.000151 | 0.74% | 0.000151 | 0.48 | 0.491 | 1.02 |
| dropD1*dropD1 | 1 | 0.00017 | 1.70% | 0.00017 | 1.18 | 0.283 | 1.02 | 0.000000 | 0.00% | 0.000000 | 0.00 | 0.990 | 1.02 |
| 2-Way Interaction | 3 | 0.00172 | 17.06% | 0.00057 | 3.96 | 0.014 | | 0.004675 | 22.91% | 0.001558 | 5.00 | 0.005 | |
| constrain*dropD1 | 1 | 0.00047 | 4.71% | 0.00047 | 3.28 | 0.077 | 1.00 | 0.002907 | 14.25% | 0.002907 | 9.33 | 0.004 | 1.00 |
| constrain*optimizer | 1 | 0.00000 | 0.00% | 0.00000 | 0.00 | 0.979 | 1.00 | 0.000045 | 0.22% | 0.000045 | 0.15 | 0.705 | 1.00 |
| dropD1*optimizer | 1 | 0.00124 | 12.35% | 0.00124 | 8.59 | 0.005 | 1.00 | 0.001724 | 8.45% | 0.001724 | 5.53 | 0.023 | 1.00 |
| Error | 43 | 0.00622 | 61.80% | 0.00015 | | | | 0.013395 | 65.64% | 0.000312 | | | |
| Lack-of-Fit | 9 | 0.00214 | 21.26% | 0.00024 | 1.98 | 0.073 | | 0.003438 | 16.85% | 0.000382 | 1.30 | 0.271 | |
| Pure Error | 34 | 0.00408 | 40.55% | 0.00012 | | | | 0.012488 | 49.43% | 0.000367 | | | |
| Total | 51 | 0.01006 | | | | | | 0.025264 | | | | | |

| Model Summary | S | R-sq | R-sq(adj) | R-sq(pred) | | S | R-sq | R-sq(adj) | R-sq(pred) |
|---|---|---|---|---|---|---|---|---|---|
| | 0.012023 | 38.19% | 26.69% | 2.62% | | 0.01765 | 34.35% | 22.14% | 0.00% |

Table 15 – CCD Analysis of Variance (ANOVA) for Training and Validation Accuracies

For test accuracy, the *Constraint* and *Optimizer* main effects were statistically significant with high F-values and corresponding P-values < 0.05, while the *Constraint×DropD1* two-factor interaction effect P-value was < 0.10. Together these three effects contributed to 26.1% of the observed variation in the experimental output. Training accuracy had the highest R-squared (adjusted) value at 26.7%, followed by test accuracy at 23.0% and validation accuracy at 23.0%.

For specificity (true negative rate) only the *Constraint×DropD1* two-factor interaction effect was statistically significant with a high F-value and a corresponding P-value < 0.05. This one effect contributed to 14.3% of the observed variation in the experimental output. R-squared (adjusted) for specificity was only 13.6%.

**Analysis of Variance (ANOVA)**

| Source | DF | Test Accuracy | | | | | | Specificity (TNR) | | | | | |
|---|---|---|---|---|---|---|---|---|---|---|---|---|---|
| | | Adj. SS | Contrib. | Adj. MS | F-Val | P-Val | VIF | Adj. SS | Contrib. | Adj. MS | F-Val | P-Val | VIF |
| Model | 8 | 0.008859 | 35.07% | 0.001107 | 2.90 | 0.011 | | 0.074179 | 27.16% | 0.009272 | 2.00 | 0.069 | |
| Linear | 3 | 0.006174 | 24.44% | 0.002058 | 5.39 | 0.003 | | 0.014706 | 5.38% | 0.004902 | 1.06 | 0.376 | |
| optimizer | 1 | 0.002399 | 9.50% | 0.002399 | 6.29 | 0.016 | 1.00 | 0.004370 | 1.60% | 0.004370 | 0.94 | 0.337 | 1.00 |
| constrain | 1 | 0.003025 | 11.97% | 0.003025 | 7.93 | 0.007 | 1.00 | 0.000158 | 0.06% | 0.000158 | 0.03 | 0.854 | 1.00 |
| dropD1 | 1 | 0.000750 | 2.97% | 0.00075 | 1.97 | 0.168 | 1.00 | 0.010178 | 3.73% | 0.010178 | 2.20 | 0.145 | 1.00 |
| Square | 2 | 0.000749 | 2.96% | 0.000374 | 0.98 | 0.383 | | 0.004160 | 1.52% | 0.002080 | 0.45 | 0.641 | |
| constrain*constrain | 1 | 0.000243 | 0.96% | 0.000243 | 0.64 | 0.430 | 1.02 | 0.000195 | 0.07% | 0.000195 | 0.04 | 0.838 | 1.02 |
| dropD1*dropD1 | 1 | 0.000411 | 1.63% | 0.000411 | 1.08 | 0.305 | 1.02 | 0.004128 | 1.51% | 0.004128 | 0.89 | 0.350 | 1.02 |
| 2-Way Interaction | 3 | 0.001936 | 7.66% | 0.000645 | 1.69 | 0.183 | | 0.055313 | 20.25% | 0.018438 | 3.99 | 0.014 | |
| constrain*dropD1 | 1 | 0.001160 | 4.59% | 0.00116 | 3.04 | 0.088 | 1.00 | 0.039006 | 14.28% | 0.039006 | 8.43 | 0.006 | 1.00 |
| constrain*optimizer | 1 | 0.000300 | 1.19% | 0.0003 | 0.79 | 0.380 | 1.00 | 0.009424 | 3.45% | 0.009424 | 2.04 | 0.161 | 1.00 |
| dropD1*optimizer | 1 | 0.000476 | 1.88% | 0.000476 | 1.25 | 0.270 | 1.00 | 0.006882 | 2.52% | 0.006882 | 1.49 | 0.229 | 1.00 |
| Error | 43 | 0.016405 | 64.93% | 0.000382 | | | | 0.198945 | 72.84% | 0.004627 | | | |
| Lack-of-Fit | 9 | 0.003917 | 15.50% | 0.000435 | 1.18 | 0.336 | | 0.058060 | 21.26% | 0.006451 | 1.56 | 0.168 | |
| Pure Error | 34 | 0.012488 | 49.43% | 0.000367 | | | | 0.140884 | 51.58% | 0.004144 | | | |
| Total | 51 | 0.025264 | | | | | | 0.273123 | | | | | |

5/19/2020

## Automated Copper Alloy Grain Size Evaluation Using a Deep-learning CNN

**Analysis of Variance (ANOVA)**

| Model Summary | S | R-sq | R-sq(adj) | R-sq(pred) | | S | R-sq | R-sq(adj) | R-sq(pred) |
|---|---|---|---|---|---|---|---|---|---|
| | 0.01953 | 35.06% | 22.98% | 2.71% | | 0.068019 | 27.16% | 13.61% | 0.00% |

Table 16 – CCD Analysis of Variance (ANOVA) for Test Accuracy and Specificity

For sensitivity (true positive rate), again, only the *Constraint×DropD1* two-factor interaction effect was statistically significant with a high F-value and a corresponding P-value < 0.05. This one effect contributed to 9.2% of the observed variation in the experimental output. R-squared (adjusted) for sensitivity was only 7.3%.

For precision (positive predictive rate), the *Constraint×DropD1* two-factor interaction effect was statistically significant with a high F-value and a corresponding P-value < 0.05, while the *DropD1* main effect P-value was < 0.10. These two effects contributed to 16.8% of the observed variation in the experimental output. R-squared (adjusted) for sensitivity was only 16.1%.

**Analysis of Variance (ANOVA)**

| | | Sensitivity (TPR) | | | | | Precision | | | | |
|---|---|---|---|---|---|---|---|---|---|---|---|
| Source | DF | Adj. SS | Contrib. | Adj. MS | F-Val | P-Val | VIF | Adj. SS | Contrib. | Adj. MS | F-Val | P-Val | VIF |
| Model | 8 | 0.16322 | 21.85% | 0.02040 | 1.50 | 0.185 | | 0.02962 | 29.22% | 0.00370 | 2.22 | 0.045 | |
| Linear | 3 | 0.05333 | 7.14% | 0.01778 | 1.31 | 0.284 | | 0.01019 | 10.06% | 0.00340 | 2.04 | 0.123 | |
| optimizer | 1 | 0.01587 | 2.13% | 0.01587 | 1.17 | 0.286 | 1.00 | 0.00304 | 3.00% | 0.00304 | 1.82 | 0.184 | 1.00 |
| constrain | 1 | 0.03669 | 4.91% | 0.03669 | 2.70 | 0.107 | 1.00 | 0.00122 | 1.21% | 0.00122 | 0.73 | 0.397 | 1.00 |
| dropD1 | 1 | 0.00077 | 0.10% | 0.00077 | 0.06 | 0.813 | 1.00 | 0.00593 | 5.85% | 0.00593 | 3.55 | 0.066 | 1.00 |
| Square | 2 | 0.02216 | 2.97% | 0.01108 | 0.82 | 0.449 | | 0.00148 | 1.46% | 0.00074 | 0.44 | 0.646 | |
| constrain*constrain | 1 | 0.00071 | 0.10% | 0.00071 | 0.05 | 0.820 | 1.02 | 0.00001 | 0.01% | 0.00001 | 0.00 | 0.946 | 1.02 |
| dropD1*dropD1 | 1 | 0.02211 | 2.96% | 0.02211 | 1.63 | 0.209 | 1.02 | 0.00147 | 1.45% | 0.00147 | 0.88 | 0.353 | 1.02 |
| 2-Way Interaction | 3 | 0.08773 | 11.75% | 0.02924 | 2.15 | 0.107 | | 0.01796 | 17.71% | 0.00599 | 3.59 | 0.021 | |
| constrain*dropD1 | 1 | 0.06891 | 9.23% | 0.06891 | 5.08 | 0.029 | 1.00 | 0.01106 | 10.91% | 0.01106 | 6.63 | 0.014 | 1.00 |
| constrain*optimizer | 1 | 0.00101 | 0.13% | 0.00101 | 0.07 | 0.787 | 1.00 | 0.00407 | 4.01% | 0.00407 | 2.44 | 0.126 | 1.00 |
| dropD1*optimizer | 1 | 0.01782 | 2.39% | 0.01782 | 1.31 | 0.258 | 1.00 | 0.00283 | 2.79% | 0.00283 | 1.70 | 0.200 | 1.00 |
| Error | 43 | 0.58369 | 78.15% | 0.01357 | | | 1.00 | 0.07175 | 70.78% | 0.00167 | | | 1.00 |
| Lack-of-Fit | 9 | 0.17877 | 23.93% | 0.01986 | 1.67 | 0.136 | | 0.02091 | 20.63% | 0.00232 | 1.55 | 0.169 | |
| Pure Error | 34 | 0.40493 | 54.21% | 0.01191 | | | | 0.05084 | 50.15% | 0.00150 | | | |
| Total | 51 | 0.74691 | | | | | | 0.10137 | | | | | |

| Model Summary | S | R-sq | R-sq(adj) | R-sq(pred) | | S | R-sq | R-sq(adj) | R-sq(pred) |
|---|---|---|---|---|---|---|---|---|---|
| | 0.116509 | 21.85% | 7.31% | 0.00% | | 0.040848 | 29.22% | 16.06% | 0.00% |

Table 17 – CCD Analysis of Variance (ANOVA) for Sensitivity and Precision

The factorial plots for the six CCD experimental responses provide direction for where to set the factor levels to achieve the best CNN performance.

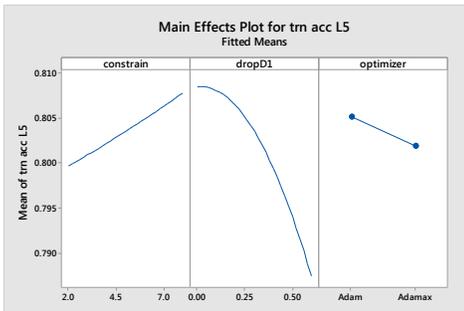

Figure 25 – CCD Factorial Plots: Training Accuracy

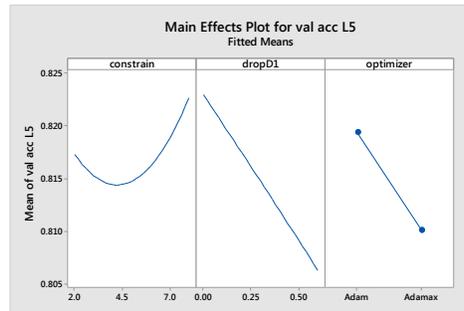

Figure 26 – CCD Factorial Plots: Validation Accuracy

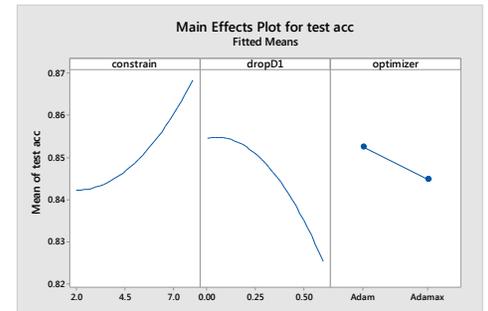

Figure 27 – CCD Factorial Plots: Test Accuracy

All three of the Group-A accuracy CCD responses exhibited a similar pattern for the three main-effect plots as shown in Figure 25, Figure 26 and Figure 27. The highest accuracies were yielded by the highest *Kernel Constraint*

                                                     5/19/2020

**Automated Copper Alloy Grain Size Evaluation Using a Deep-learning CNN**

level and the lowest drop values for the CNN's dense layer-1 training settings. The Adam *Optimizer* yielded higher accuracies than the Adamax *Optimizer*.

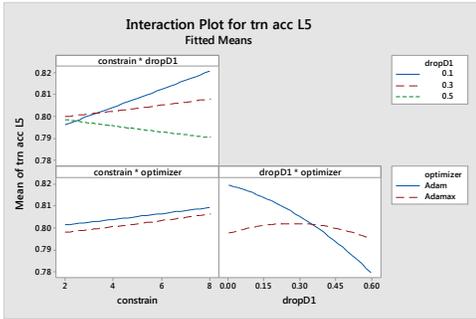 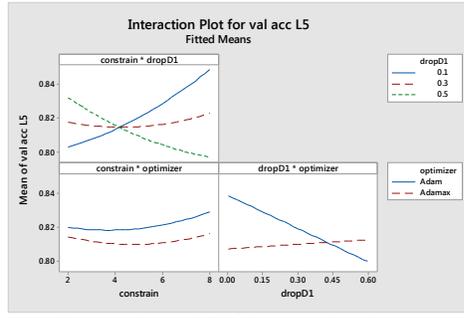 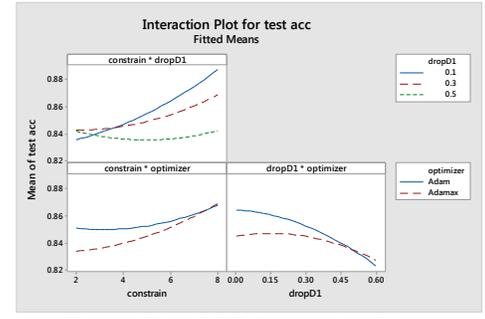

Figure 28 – CCD Interaction Plots: Training Accuracy  Figure 29 – CCD Interaction Plots: Validation Accuracy  Figure 30 – CCD Interaction Plots: Test Accuracy

All three of the Group-A accuracy CCD responses exhibited similar patterns for the three two-factor interaction effect plots as shown in Figure 28, Figure 29 and Figure 30. The surprising interaction was the *Constraint×DropD1* that again showed that higher accuracies were yielded by the largest kernel constraint level and the lowest drop values for the CNN's dense layer-1. The *DropD1×Optimizer* interaction showed that the Adam *Optimizer* yielded higher accuracies when the CNN's dense layer-1 *DropD1* values were lowest, while accuracies were not as affected by the dense layer-1 *DropD1* values when the Adamax *Optimizer* was used. Interactions were not studied during the DSD experiment, which may partially explain why the relationship between the Adam and Adamax *Optimizers* was different for the screening phase.

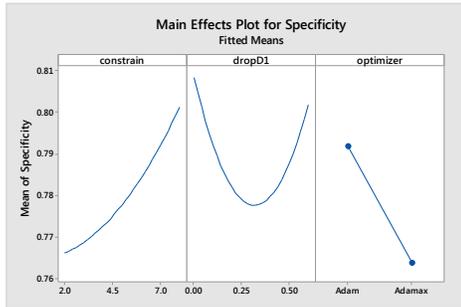 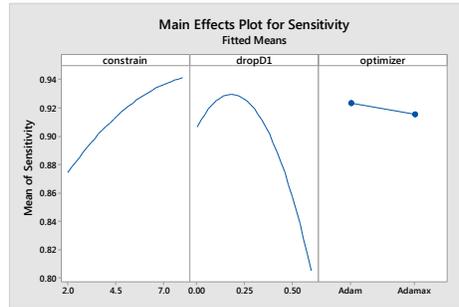 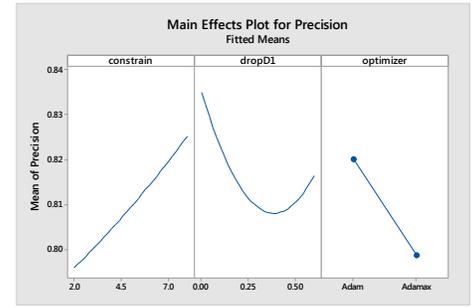

Figure 31 – CCD Factorial Plots: Specificity  Figure 32 – CCD Factorial Plots: Sensitivity  Figure 33 – CCD Factorial Plots: Precision

The three Group-B confusion matrix main effect responses are shown in Figure 31, Figure 32 and Figure 33, and for precision, only the linear component of the CNN's dense layer-1 drop value (Figure 33 center panel) was significant at 90% confidence with a P-value < 0.10.

The three Group-B confusion matrix CCD two-factor interaction responses are shown in Figure 34, Figure 35 and Figure 36, and only the *Constraint×Optimizer* interactions were significant with P-values < 0.05 for all three responses. The sensitivity interaction (Figure 35 upper right panel) appears to run counter to the same interaction for the specificity and precision responses, where the relationship of the 0.1 and 0.5 *DropD1* curves are reversed, suggesting that if the CNN's dense layer-1 *DropD1* value is lowered to 0.1 or less with the kernel constraint set high, the sensitivity would be lowered. This may not be detrimental because the CNN appears to be biased towards the correct classification of 'good grains', with the average sensitivity approaching 1.0, which tends to elevate the sensitivity (true positive rate) and lower both the specificity (true negative rate) and precision (positive predictive rate). Elevating the specificity at the expense of sensitivity would encourage the reduction of true positive rate bias in the CNN.

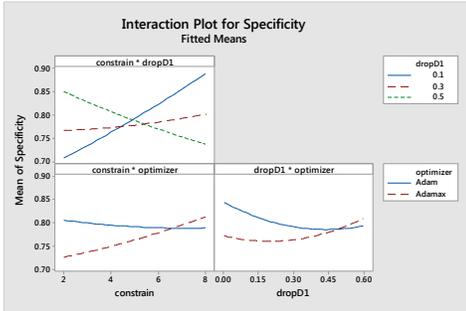

Figure 34 – CCD Interaction Plots: Specificity

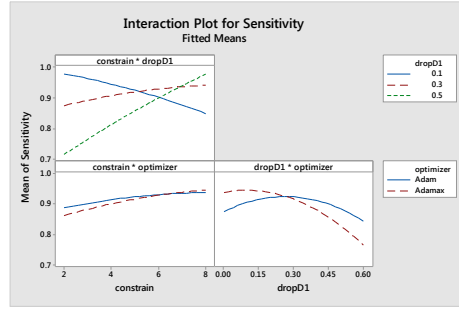

Figure 35 – CCD Interaction Plots: Sensitivity

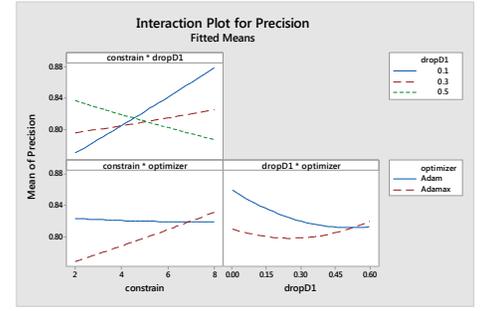

Figure 36 – CCD Interaction Plots: Precision

### 5.3.4. CCD Verification Check

The CNN CCD-configuration for the highest training accuracy, validation accuracy, test accuracy and specificity was run through 250 epochs of training on the existing data set. The *Constraint×DropD1* two-factor interaction was used to set the dense layer-1 *Kernel Constraint* to 7.5 and the dense layer-1 drop level to 0, which departed from settings recommended by best practice (*DropD1* between 20% and 50% and *Kernel Constraint* = 3 to 4) for CNN training.

| batch | constrain | optimizer | dropC1 | dropC2 | dropC3 | dropD1 | maxpC1 |
|-------|-----------|-----------|--------|--------|--------|--------|--------|
| 240 | 7.5 | Adam | 0 | 0 | 0 | 0 | 2 |
| **maxpC2** | **maxpC3** | **filtC1** | **filtC2** | **filtC3** | **padding** | **strideC1** | **activation** |
| 2 | 2 | 3 | 3 | 3 | same | 1 | ReLU |

Table 18 – CNN configuration for CCD Verification Check

The CCD-configured CNN successfully trained for all 5 test runs with 250 epochs of training for each run, with no evidence of overfitting. The average test accuracy improved to 90.5% from the 88.3% test accuracy achieved for the DSD-configured CNN. The average specificity achieved was 86.4%, which was lower than the 94.7% average sensitivity, but this was an improvement over the DSD-configured CNN that had an average specificity of 81.2% and an average sensitivity of 95.5%; the CCD-configured CNN, while improved over the DSD-configuration, still favors classification of good over bad grain structure (the average false positive rate also improved to 13.6% from 18.8%). With the exception of Run 2, the CCD-configured CNN test results were consistent. The CCD verification runs with confusion matrix metrics are shown in Table 19.

| Run | Training Plots | Confusion Matrix Metrics | | |
|-----|----------------|--------------------------|---|---|
| | | **Predicted Class Labels** | | |
| | | Class | Bad Grains | Good Grains | True Rates |
| | | **Actual Class Labels** | | | |
| 1 |  | Bad Grains | TN = 359 | FP = 41 | Specificity = 89.75% |
| | | Good Grains | FN = 28 | TP = 372 | Sensitivity = 93.00% |
| | | | False Positive Rate = 10.25% | Precision = 90.07% | Accuracy = 91.38% |
| | Train. Acc. (L5): 89.77% Val. Acc. (L5): 91.57% | | | | Time: 57.20 min. |

**Automated Copper Alloy Grain Size Evaluation Using a Deep-learning CNN**

| Run | Training Plots | Confusion Matrix Metrics |
|-----|----------------|--------------------------|

**Run 2**

| | | **Predicted Class Labels** | | |
|---|---|---|---|---|
| | **Class** | Bad Grains | Good Grains | True Rates |
| **Actual Class Labels** | | | | |
| | Bad Grains | TN = 307 | FP = 93 | Specificity = 76.75% |
| | Good Grains | FN = 9 | TP = 391 | Sensitivity = 97.75% |
| | | False Positive Rate = 23.25% | Precision = 80.79% | Accuracy = 87.25% |
| Train. Acc. (L5): 89.95% | | | | |
| Val. Acc. (L5): 91.42% | | | | Time: 57.20 min. |

**Run 3**

| | | **Predicted Class Labels** | | |
|---|---|---|---|---|
| | **Class** | Bad Grains | Good Grains | True Rates |
| **Actual Class Labels** | | | | |
| | Bad Grains | TN = 357 | FP = 43 | Specificity = 89.25% |
| | Good Grains | FN = 26 | TP = 374 | Sensitivity = 93.50% |
| | | False Positive Rate = 10.75% | Precision = 89.69% | Accuracy = 91.38% |
| Train. Acc. (L5): 89.98% | | | | |
| Val. Acc. (L5): 90.43% | | | | Time: 56.80 min. |

**Run 4**

| | | **Predicted Class Labels** | | |
|---|---|---|---|---|
| | **Class** | Bad Grains | Good Grains | True Rates |
| **Actual Class Labels** | | | | |
| | Bad Grains | TN = 348 | FP = 52 | Specificity = 87.00% |
| | Good Grains | FN = 19 | TP = 381 | Sensitivity = 95.25% |
| | | False Positive Rate = 13.00% | Precision = 87.99% | Accuracy = 91.13% |
| Train. Acc. (L5): 91.28% | | | | |
| Val. Acc. (L5): 92.47% | | | | Time: 56.50 min. |

**Run 5**

| | | **Predicted Class Labels** | | |
|---|---|---|---|---|
| | **Class** | Bad Grains | Good Grains | True Rates |
| **Actual Class Labels** | | | | |
| | Bad Grains | TN = 357 | FP = 43 | Specificity = 89.25% |
| | Good Grains | FN = 25 | TP = 375 | Sensitivity = 93.75% |
| | | False Positive Rate = 10.75% | Precision = 89.71% | Accuracy = 91.50% |
| Train. Acc. (L5): 90.30% | | | | |
| Val. Acc. (L5): 90.78% | | | | Time: 56.38 min. |

Table 19 – CCD Verification Tests Results

## 5.4. Data Set Verification

### 5.4.1. K-Fold Cross-Validation

The 7420 sample dataset was randomly partitioned 10 times into separate datasets of 5020 training samples, 1600 training validation samples, and 800 test verification samples, creating a 10-fold cross-validation dataset. The optimized CNN-configuration from the CCD experimental output was then trained using randomly initialized weights to start each *Run*, with 5 training, validation and test *Runs* performed on each of the *k-Fold* datasets.

The 10-fold cross-validation results appear in the GLM (General Linear Model) ANOVA of Table 20, Figure 37, Figure 38 and Figure 39, as well as Table 21. Although much more variation came from *k-Fold*, only the orange-highlighted *k-Fold* factor was statistically significant with a P value < 0.10. Figure 37 provides a graphical representation of the ANOVA results in Table 20, showing the variation between *k-Fold* 1 and 2 was greater than the variation between test *Runs* 1 through 5 within each *k-Fold* level.

## Automated Copper Alloy Grain Size Evaluation Using a Deep-learning CNN

The ability of the CCD-configured CNN to act as a generalized Cu-alloy grains classifier is demonstrated in Figure 39, where the average accuracy was highest for the test datasets. The variation also increases at each stage, with the highest variance observed in test accuracy results; several lower-accuracy outliers are visible in the test results.

**Analysis of Variance (General Linear Model)**

**Test Accuracy**

| Source | DF | Adj. SS | Contribution | Adj. MS | F-Value | P-Value | VIF |
|--------|-----|---------|--------------|----------|---------|---------|------|
| k-Fold | 9 | 0.005673 | 30.01% | 0.000630 | 1.88 | 0.087 | 1.80 |
| Run | 4 | 0.001147 | 6.07% | 0.000287 | 0.85 | 0.501 | 1.60 |
| Error | 36 | 0.012085 | 63.92% | 0.000336 | | | |
| Total | 49 | 0.018905 | 100.00% | | | | |

**Model Summary**

| S | R-sq | R-sq(adj) | R-sq(pred) |
|---|------|-----------|------------|
| 0.0183219 | 36.07% | 12.99% | 0.00% |

Table 20 – 10-Fold Cross-Validation Test Accuracy ANOVA

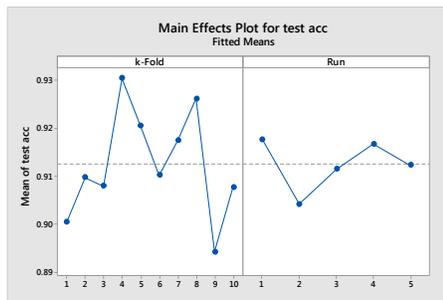

Figure 37 – 10-Fold Cross-Validation Mean Test Accuracies Partitioned between Fold and Run

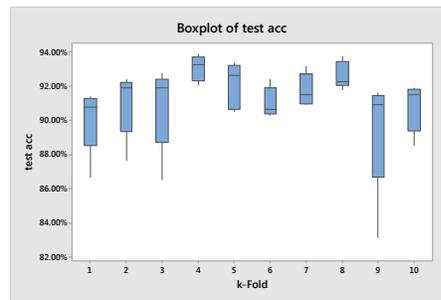

Figure 38 – 10-Fold Cross-Validation Boxplot of Test Accuracies for each Dataset Fold

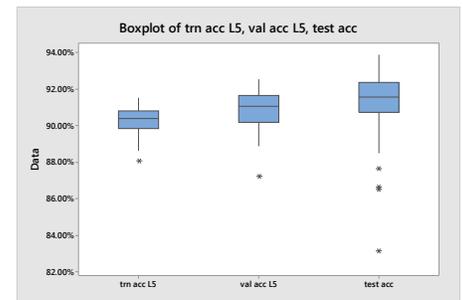

Figure 39 – 10-Fold Cross-Validation Boxplots of Accuracies for Training, Validation and Test Data

| | Run Time (minutes) | Training Accuracy | Validation Accuracy | Test Accuracy | Specificity | Sensitivity | FPR | Precision |
|---|---|---|---|---|---|---|---|---|
| X-bar | 56.903 | 90.296% | 90.854% | 91.253% | 89.710% | 92.795% | 10.290% | 90.408% |
| S | 0.264 | 0.727% | 1.069% | 1.964% | 5.790% | 4.783% | 5.790% | 4.372% |

Table 21 – 10-Fold Cross-Validation Test Accuracy Summary Statistics

The CCD-optimized CNN achieved 91.3% ± 2.0% test accuracy on the 10-fold cross-validation datasets as summarized in Table 21, where S is the sample standard deviation. The variation between the 10 cross-validation *k-Fold* datasets was much greater than the variation within the 5 training *Runs* performed on each cross-validation dataset. This can also be seen by comparing the two best *k-Fold* datasets 4 and 8 with the worst dataset 9; test accuracies ± S were 93.1 ±0.7%, 92.3 ±0.8% and 89.4 ±3.6% respectively for datasets 4, 8 and 9.

The 10-fold cross-validation represents an estimate of what would possible to achieve with this Cu-alloy dataset using the CCD-optimized CNN, where a training, validation and test dataset could be created that could either be more like datasets 4 and 8 that achieve classification accuracies around 93 ±1%, or more like dataset 9, with classification accuracies of only 89 ±4%.

The k-fold cross-validation, with 10 folds and 5 runs-per-fold, required 47 hours and 25 minutes to complete.

## 5.5.   Regularization Experiments

A preliminary DSD screening experiment was performed using both the *L1* and *L2* regularization terms and these were initially applied to all 4 layers (convolutional layers 1, 2 and 3, and dense layer-1) of the CCD-optimized CNN. The *L1* and *L2* terms were varied logarithmically between 1.E-3, 1.E-5 and 1.E-7 using a DSD experimental matrix with 17 treatment combinations, and the experiment was replicated once requiring 34 total runs; however, the DSD was unable to learn the training set with this regularization applied. Instead of an exhaustive search of the experimental input space[27], the decision was made to simply eliminate the *L1* regularization terms from all layers, and this allowed the CNN to learn the dataset. The *L2* weight regularization experimental matrix appears in Table 22.

### 5.5.1.       Regularization Matrix

The L2 regularization values were logarithmically varied as shown below, while the other factors remained at the CCD-optimization settings shown in Table 18.

| TC | C1_L2 | C2_L2 | C3_L2 | D1_L2 |
|----|-------|-------|-------|-------|
| 1  | 1.E-03 | 1.E-03 | 1.E-03 | 1.E-03 |
| 2  | 1.E-07 | 1.E-07 | 1.E-07 | 1.E-07 |
| 3  | 1.E-05 | 1.E-03 | 1.E-07 | 1.E-07 |
| 4  | 1.E-05 | 1.E-07 | 1.E-03 | 1.E-03 |
| 5  | 1.E-03 | 1.E-05 | 1.E-03 | 1.E-07 |
| 6  | 1.E-07 | 1.E-05 | 1.E-07 | 1.E-03 |
| 7  | 1.E-07 | 1.E-03 | 1.E-05 | 1.E-03 |
| 8  | 1.E-03 | 1.E-07 | 1.E-05 | 1.E-07 |
| 9  | 1.E-07 | 1.E-07 | 1.E-03 | 1.E-05 |
| 10 | 1.E-03 | 1.E-03 | 1.E-07 | 1.E-05 |
| 11 | 1.E-03 | 1.E-07 | 1.E-07 | 1.E-03 |
| 12 | 1.E-07 | 1.E-03 | 1.E-03 | 1.E-07 |
| 13 | 1.E-05 | 1.E-05 | 1.E-05 | 1.E-05 |

Table 22 – DSD Regularization Experimental Matrix

The experiment was replicated 3 additional times with a total of 52 runs performed. Each of the runs was subjected to 100 epochs of training, validation and test verification, and the entire experiment required 19 hours and 48 minutes to complete.  The sample standard deviations for test accuracy and specificity (TNR – true negative rate) for each treatment combination across the 4 replicated experiments were used as the primary experimental responses, while the test accuracy and specificity percentages served as secondary responses. The experimental objective was to prove the hypothesis that the *L2* weight regularization terms had an effect on the variation observed between identical test cases. Test accuracy provided a measure of the CNN's ability to generalize what it has learned during training. Specificity provided an indication for misclassification of the bad grain structures, because the CNN was biased towards the correct classification of good grain structure over bad grain structure.

### 5.5.2.       Regularization Experimental Results

The ANOVA of the replication standard deviation for the test accuracy and specificity responses are shown in Table 23. The test accuracy standard deviation response had no significant effects, and for the specificity standard deviation response, only the *L2* regularization for dense layer-1 (*D1_L2*) factor had a statistically significant effect at 90% confidence with a P-value < 0.10 (highlighted orange). The main effect responses are plotted in Figure 40 and Figure 41, using the same Y-scale range of 3%, with the *L2* exponent values on the X-axes. The trends for the

---

[27] A preliminary exploration of the experimental input space, prior to designing a new DoE could be achieved using a low-discrepancy quasi-random sampling Sobol Sequence, or Latin Hypercube sampling, which would provide more uniformity than the Sobol Sequence. This approach was rejected because of the required resources when compared to the performance already achieved by the CCD-optimized CNN as documented in Section 5.4.1.

### Automated Copper Alloy Grain Size Evaluation Using a Deep-learning CNN

four *L2* regularization factors were identical between the two response factors, where higher value of *L2* regularization for convolutional layer-1 (*C1_L2*) reduced replication variation, while lower values of *L2* regularization for the other 3 CNN layers reduced replication variation. Overall, the effect on run-to-run variation from *L2* regularization was weak.

**Analysis of Variance (ANOVA) – Replication Sample Standard Deviation**

| | | Test Accuracy | | | | | | Specificity (TNR) | | | | | |
|---|---|---|---|---|---|---|---|---|---|---|---|---|---|
| Source | DF | Adj. SS | Contrib. | Adj. MS | F-Val | P-Val | VIF | Adj. SS | Contrib. | Adj. MS | F-Val | P-Val | VIF |
| Model | 4 | 0.000076 | 8.50% | 0.000019 | 0.19 | 0.939 | | 0.004216 | 53.52% | 0.001054 | 2.30 | 0.147 | |
| Linear | 4 | 0.000076 | 8.50% | 0.000019 | 0.19 | 0.939 | | 0.004216 | 53.52% | 0.001054 | 2.30 | 0.147 | |
| C1_L2 | 1 | 0.000007 | 0.78% | 0.000007 | 0.06 | 0.806 | 1.00 | 0.001115 | 14.15% | 0.001115 | 2.44 | 0.157 | 1.00 |
| C2_L2 | 1 | 0.000001 | 0.11% | 0.000001 | 0.01 | 0.920 | 1.00 | 0.000293 | 3.72% | 0.000293 | 0.64 | 0.447 | 1.00 |
| C3_L2 | 1 | 0.000001 | 0.11% | 0.000001 | 0.01 | 0.917 | 1.00 | 0.001065 | 13.52% | 0.001065 | 2.33 | 0.166 | 1.00 |
| D1_L2 | 1 | 0.000067 | 7.49% | 0.000067 | 0.66 | 0.441 | 1.00 | 0.001744 | 22.14% | 0.001744 | 3.81 | 0.087 | 1.00 |
| Error | 8 | 0.000818 | 91.50% | 0.000102 | | | | 0.003662 | 46.48% | 0.000458 | | | |
| Total | 12 | 0.000894 | 100.00% | | | | | 0.007878 | 100.00% | | | | |

| Model Summary | | S | R-sq | R-sq(adj) | R-sq(pred) | | | S | R-sq | R-sq(adj) | R-sq(pred) |
|---|---|---|---|---|---|---|---|---|---|---|---|
| | | 0.000894 | 8.49% | 0.00% | 0.00% | | | 0.0213957 | 53.52% | 30.27% | 0.00% |

Table 23 –Analysis of Variance (ANOVA), Regularization L2 for
Standard Deviation in Replicates of Test Accuracy and Specificity

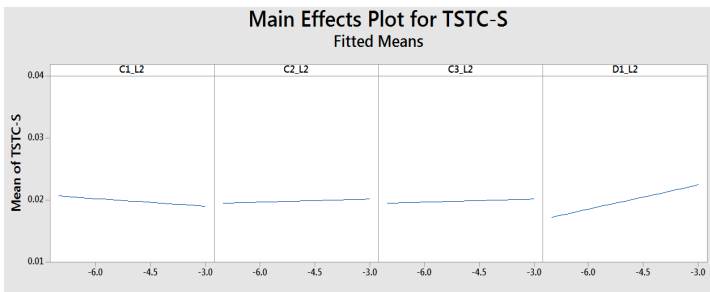

Figure 40 – L2 Regularization for Replicate Sample Standard
Deviation for Test Accuracy

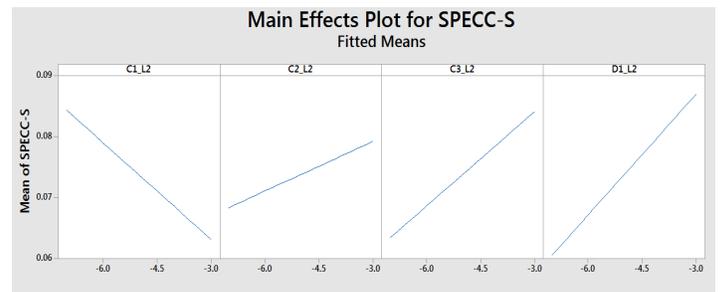

Figure 41 – L2 Regularization for Replicate Sample Standard
Deviation for Specificity

The effects of *L2* regularization on test accuracy and specificity were not significant. This is shown in Table 24 summary of ANOVA, where no main effect had a P-value below 0.26 for test accuracy or below 0.15 for specificity. Pure error from replication contributed 88.5% and 71.5% to the observed variation for test accuracy and specificity respectively. The main-effect responses are plotted in Figure 40, using the same Y-scale range of 4%, with the *L2* exponent values on the X-axes. The trends for the four *L2* regularization factors were the same between the 2 response factors, where lower values of *L2* regularization for the second and third convolutional layers tended to increase test accuracy and specificity, while *L2* regularization to the first convolutional layer had no effect. Overall, the effects from the *L2* regularization on test accuracy and specificity were not significant.

**Analysis of Variance (ANOVA)**

| | | Test Accuracy | | | | | | Specificity (TNR) | | | | | |
|---|---|---|---|---|---|---|---|---|---|---|---|---|---|
| Source | DF | Adj. SS | Contrib. | Adj. MS | F-Val | P-Val | VIF | Adj. SS | Contrib. | Adj. MS | F-Val | P-Val | VIF |
| Model | 4 | 0.000981 | 4.83% | 0.000245 | 0.60 | 0.667 | | 0.032775 | 9.95% | 0.008194 | 1.30 | 0.284 | |
| Linear | 4 | 0.000981 | 4.83% | 0.000245 | 0.60 | 0.667 | | 0.032775 | 9.95% | 0.008194 | 1.30 | 0.284 | |
| C1_L2 | 1 | 0.000000 | 0.00% | 0.000000 | 0.00 | 1.000 | 1.00 | 0.000000 | 0.00% | 0.000000 | 0.00 | 1.000 | 1.00 |
| C2_L2 | 1 | 0.000545 | 2.68% | 0.000545 | 1.32 | 0.256 | 1.00 | 0.013783 | 4.18% | 0.013783 | 2.18 | 0.146 | 1.00 |
| C3_L2 | 1 | 0.000257 | 1.27% | 0.000257 | 0.62 | 0.434 | 1.00 | 0.011223 | 3.41% | 0.011223 | 1.78 | 0.189 | 1.00 |
| D1_L2 | 1 | 0.000179 | 0.88% | 0.000179 | 0.44 | 0.512 | 1.00 | 0.007770 | 2.36% | 0.007770 | 1.23 | 0.273 | 1.00 |
| Error | 47 | 0.019321 | 95.17% | 0.000411 | | | | 0.296641 | 90.05% | 0.006312 | | | |
| Lack-of-Fit | 8 | 0.001362 | 6.71% | 0.000170 | 0.37 | 0.930 | | 0.061126 | 18.56% | 0.007641 | 1.27 | 0.289 | |
| Pure Error | 39 | 0.017959 | 88.46% | 0.000460 | | | | 0.235516 | 71.49% | 0.006039 | | | |
| Total | 51 | 0.020302 | 100.00% | | | | | 0.329417 | 100.00% | | | | |

| Analysis of Variance (ANOVA) | | | | | | | | |
|---|---|---|---|---|---|---|---|---|
| **Model Summary** | S | R-sq | R-sq(adj) | R-sq(pred) | | S | R-sq | R-sq(adj) | R-sq(pred) |
| | 0.0202753 | 4.83% | 0.00% | 0.00% | | 0.0794451 | 9.95% | 2.29% | 0.00% |

<div align="center">Table 24 –Analysis of Variance (ANOVA), L2 Regularization for Test Accuracy and Specificity</div>

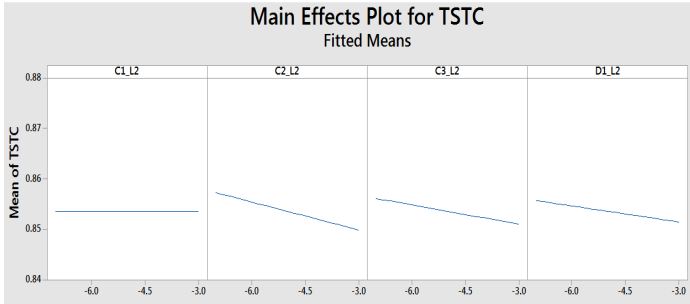

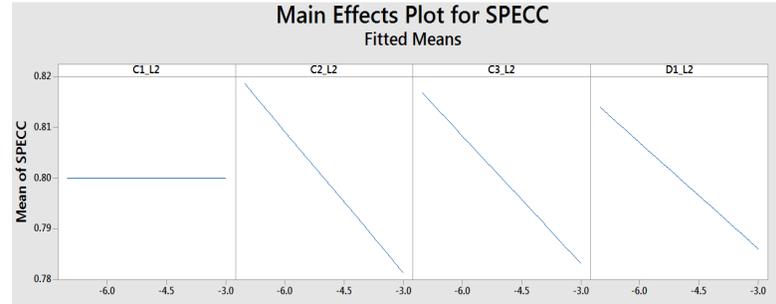

<div align="center">Figure 42 – L2 Regularization for Test Accuracy      Figure 43 – L2 Regularization for Specificity</div>

### 5.5.3. Regularization Verification Check

The CNN CCD-configuration defined in Table 18 was used with the L2 regularization settings as shown in Table 25.

| C1_L2 | C2_L2 | C3_L3 | D1_L2 |
|---|---|---|---|
| 1.E-3 | 1.E-7 | 1.E-7 | 1.E-7 |

<div align="center">Table 25 – CNN L2 Settings for Regularization Verification Check</div>

Five successful test runs were conducted on the dataset, with 250 epochs of training for each run and no indication of overfitting.

| | Run Time (minutes) | Training Accuracy | Validation Accuracy | Test Accuracy | Specificity | Sensitivity | FPR | Precision |
|---|---|---|---|---|---|---|---|---|
| X-bar | 56.753 | 81.367% | 87.784% | 90.500% | 86.600% | 94.400% | 13.400% | 87.754% |
| S | 0.0974 | 1.155% | 1.200% | 1.007% | 4.284% | 2.866% | 4.284% | 3.210% |

<div align="center">Table 26 –Regularization Verification Check Results</div>

Any evidence of improvement could not be determined from this data because the test results fell well within the results for the 10-fold cross-validation of Section 5.4.1; therefore, the 10-fold cross-validation was repeated with the *L2* weight regularization settings.

The 10-fold cross-validation results appear in the GLM ANOVA of Table 27, Figure 44, Figure 45 and Figure 46, as well as Table 29. More variation came from the *Run* factor, and only the orange-highlighted *Run* factor was statistically significant with a P value < 0.10 at 90% confidence. Figure 44 provides a graphical representation of the ANOVA results in Table 27, showing the variation between *k-Fold* and the variation between tests *Run* within the *k-Fold*.

With the exception of Fold 6, the 10-fold cross-validation run with optimized *L2* weight regularization appears to be relatively consistent from fold to fold. The change in patterns between Figure 38 and Figure 45 indicates that the variation between *k-Fold* 1 and 2 is primarily due the stochastic nature of the CNN during training and not because of a difference between *k-Fold* 1 and 2 (otherwise the patterns would have been similar).

# Automated Copper Alloy Grain Size Evaluation Using a Deep-learning CNN

**Analysis of Variance (General Linear Model)**

**Test Accuracy**

| Source | DF | Adj. SS | Contribution | Adj. MS | F-Value | P-Value | VIF |
|--------|----|---------|--------------|---------|---------|---------|-----|
| k-Fold | 9 | 0.004731 | 25.55% | 0.000526 | 1.71 | 0.124 | 1.80 |
| Run | 4 | 0.002687 | 14.51% | 0.000672 | 2.18 | 0.091 | 1.60 |
| Error | 36 | 0.011097 | 59.93% | 0.000308 | | | |
| Total | 49 | 0.018516 | 100.00% | | | | |

**Model Summary**

| S | R-sq | R-sq(adj) | R-sq(pred) |
|---|------|-----------|------------|
| 0.0175572 | 40.07% | 18.42% | 0.00% |

Table 27 – L2 Regularization 10-Fold Cross-Validation Test Accuracy ANOVA

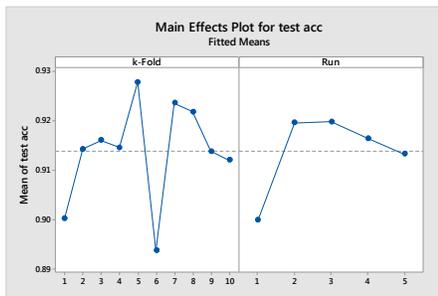

Figure 44 – L2 10-Fold Cross-Validation Mean Test Accuracies Partitioned between Fold and Run

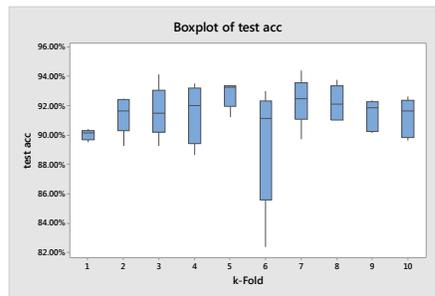

Figure 45 – L2 10-Fold Cross-Validation Boxplot of Test Accuracies for each Dataset Fold

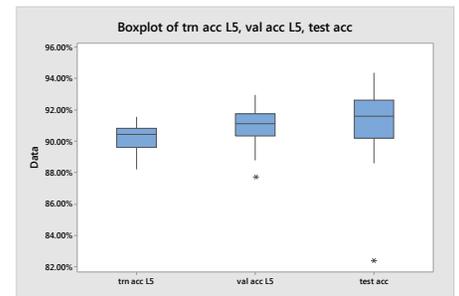

Figure 46 – L2 10-Fold Cross-Validation Boxplots of Accuracies for Training, Validation and Test Data

| | Run Time (minutes) | Training Accuracy | Validation Accuracy | Test Accuracy | Specificity | Sensitivity | FPR | Precision |
|---|---|---|---|---|---|---|---|---|
| X-bar | 57.067 | 90.259% | 91.020% | 91.375% | 88.865% | 93.885% | 11.135% | 89.612 |
| S | 0.154 | 0.752% | 1.028% | 1.944% | 4.183% | 4.863% | 4.183% | 3.258 |

Table 28 – L2 10-Fold Cross-Validation Test Accuracy Summary Statistics

The pre and post *L2* weight regularization 10-fold cross-validation results for test accuracy means are compared in the Table 29 ANOVA and Figure 47. The factor *Cross-Validation* reflects running the 10-fold cross-validation before (1) and after (2) the *L2* weight regularization. The Table 27 ANOVA indicates that the pre and post *L2* regularization cross-validations are statistically identical when compared to the variation between folds (*k-Fold*) and within folds (*Run*); therefore, the null hypothesis was accepted. This is shown graphically in the scatterplot of Figure 48.

**Analysis of Variance (General Linear Model)**

**Test Accuracy**

| Source | DF | Adj. SS | Contribution | Adj. MS | F-Value | P-Value | VIF |
|--------|----|---------|--------------|---------|---------|---------|-----|
| Cross-Validation | 1 | 0.000038 | 0.10% | 0.000038 | 0.11 | 0.741 | 1.00 |
| k-Fold | 9 | 0.007642 | 20.40% | 0.000849 | 2.49 | 0.014 | 1.80 |
| Run | 4 | 0.000766 | 2.04% | 0.000192 | 0.56 | 0.691 | 1.60 |
| Error | 85 | 0.029011 | 77.45% | 0.000341 | | | |
| Total | 99 | 0.037458 | | | | | |

**Model Summary**

| S | R-sq | R-sq(adj) | R-sq(pred) |
|---|------|-----------|------------|
| 0.0184746 | 22.55% | 9.79% | 0.00% |

Table 29 – Pre and Post L2 Regularization 10-Fold Cross-Validation Test Accuracy ANOVA

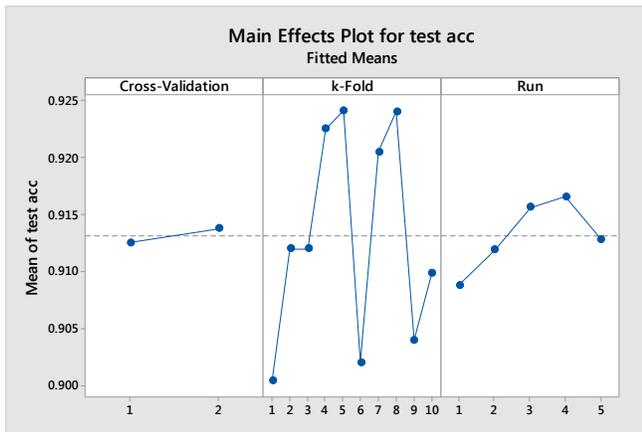

Figure 47 –Mean Test Accuracies Partitioned between Cross-Validation, k-Fold and Run

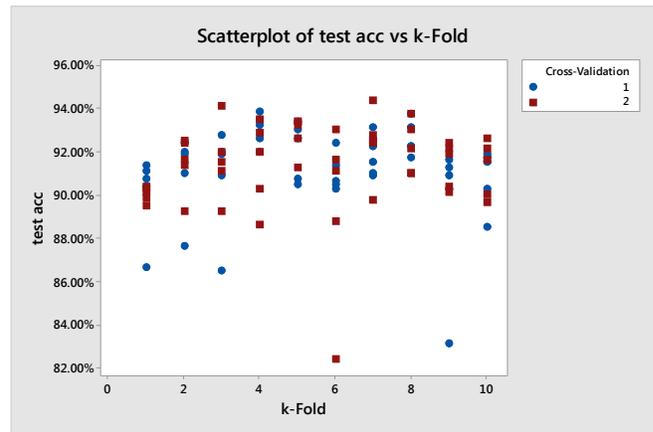

Figure 48 –Mean Test Accuracies vs k-Fold

## 5.6. Expanded Dataset K-Fold Cross-Validation

The initial dataset was expanded as described in Section 3.7.2, and the 10-fold cross-validation of Section 5.4.1 was repeated on the expanded dataset discussed in Section 3.7.2 using the CCD-optimized CNN. The *L2* weight regularization settings described in Section 5.5 were also applied to the CNN.

The CNN was successfully retrained on the expanded dataset, using the best weight (see Footnote 14) from the fourth fold of the *L2* 10-fold cross-validation. We reduced training epochs to 200 from 250 based on a trial run that indicated the possibility of overfitting[29] beyond 200 epochs.

The CNN successfully completed all 50 runs and achieved a test accuracy of 91.1 ±0.7% for the expanded 10-fold dataset, and required 71 hours and 10 minutes to complete. The CNN was now slightly more biased towards the classification of bad grains, with specificity exceeding sensitivity at 92.6 ±2.2% and 89.5 ±2.3%, respectively. Note that initially, we attempted to use fine tuning [27] on the pre-trained CNN to conduct the cross-validation, where only the weights for last convolutional feature extraction layer and the 2 dense classification layers were allowed to update during training, but the test accuracy achieved was lower and variation was greater at 86.7 ±1.3%. For additional details, refer to Appendix B: k-Fold Fine Tuning.

The 10-fold cross-validation results appear in the GLM ANOVA of Table 30, Figure 49, Figure 50 and Figure 51, as well as Table 31. More variation came from the *k-Fold* factor, and only the orange-highlighted *k-Fold* factor was statistically significant with a P value < 0.05 at 95% confidence. Figure 49 provides a graphical representation of the ANOVA results in Table 30, showing the variation between *k-Fold* and the variation between test runs within *k-Fold*. To illustrate the reduction in variation achieved with the expanded dataset cross-validation, the vertical axes scales for the previous cross-validations of Sections 5.4.1 and 5.5.3 were retained for Figure 49, Figure 50 and Figure 51. The variation within the expanded dataset 10-fold cross-validation was reduced by a factor of 2 to 3 from that of the previous 10-fold cross-validations primarily because the CNN had been pre-trained on the initial dataset thereby reducing the stochastic component from training.

---

[29] Validation accuracy dropping below training accuracy

**Automated Copper Alloy Grain Size Evaluation Using a Deep-learning CNN**

**Analysis of Variance (General Linear Model)**

**Test Accuracy**

| Source | DF | Adj. SS | Contribution | Adj. MS | F-Value | P-Value | VIF |
|--------|----|---------|--------------|---------|---------|---------|-----|
| k-Fold | 9 | 0.000937 | 38.46% | 0.000104 | 2.67 | 0.017 | 1.80 |
| Run | 4 | 0.000097 | 3.98% | 0.000024 | 0.62 | 0.650 | 1.60 |
| Error | 36 | 0.001403 | 57.59% | 0.000039 | | | |
| Total | 49 | 0.002436 | 100.00% | | | | |

**Model Summary**

| S | R-sq | R-sq(adj) | R-sq(pred) |
|---|------|-----------|------------|
| 0.0062426 | 42.42% | 21.63% | 0.00% |

Table 30 – Expanded Dataset 10-Fold Cross-Validation Test Accuracy ANOVA

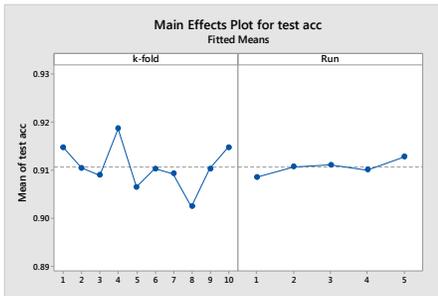

Figure 49 – Expanded Dataset 10-Fold Cross-Validation Mean Test Accuracies Partitioned between Fold and Run

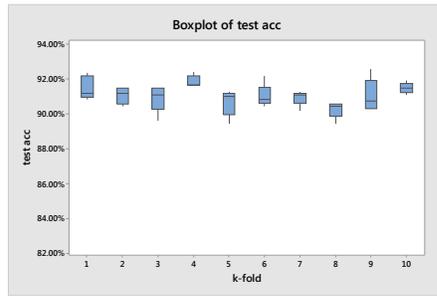

Figure 50 – Expanded Dataset 10-Fold Cross-Validation Boxplot of Test Accuracies for each Dataset Fold

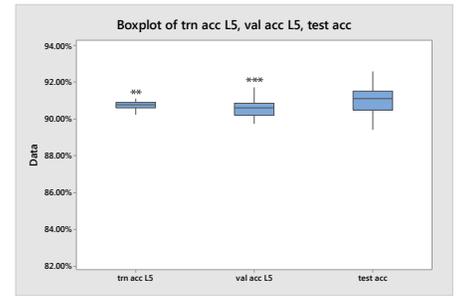

Figure 51 – Expanded Dataset 10-Fold Cross-Validation Boxplots of Accuracies for Training, Validation and Test Data

| | Run Time (minutes) | Training Accuracy | Validation Accuracy | Test Accuracy | Specificity | Sensitivity | FPR | Precision |
|---|---|---|---|---|---|---|---|---|
| X-bar | 85.408 | 90.774% | 90.689% | 91.068% | 92.632% | 89.511% | 7.368% | 92.501% |
| S | 0.201 | 0.224% | 0.500% | 0.705% | 2.187% | 2.260% | 2.187% | 1.871% |

Table 31 – Expanded Dataset 10-Fold Cross-Validation Test Accuracy Summary Statistics

# 6. Results Summary

A simple binary classification CNN was trained to recognize good and bad grain structure in Cu-alloy using a series of statistically designed experiments to screen and then optimize both the CNN configuration and training hyperparameters simultaneously during the training and test phase. The results from each experiment was verified with a 5-run check using the experimentally-determined best factor settings. The results of the optimization were then verified using a series of 10-fold cross-validation tests. The summary results appear in tables and figures below.

| | Runs | Run Time (minutes) | Training Accuracy | Validation Accuracy | Test Accuracy | Specificity | Sensitivity | FPR | Precision |
|---|---|---|---|---|---|---|---|---|---|
| A | 5 | 56.51 | 88.94% | 91.04% | 88.33% | 81.20% | 95.45% | 18.80% | 84.05% |
| B | 5 | 56.82 | 90.26% | 91.33% | 90.53% | 86.40% | 94.65% | 13.60% | 87.65% |
| C | 50 | 56.90 | 90.30% | 90.85% | 91.25% | 89.71% | 92.80% | 10.29% | 90.41% |
| D | 5 | 56.75 | 81.37% | 87.78% | 90.50% | 86.60% | 94.40% | 13.40% | 87.75% |
| E | 50 | 57.07 | 90.26% | 91.02% | 91.38% | 88.87% | 93.89% | 11.14% | 89.61% |
| F | 50 | 85.41 | 90.77% | 90.69% | 91.07% | 92.63% | 89.51% | 7.37% | 92.50% |

| | |
|---|---|
| A | DSD Screening Experiment |
| B | CCD Optimization Experiment |
| C | 10-Fold Cross-Validation Using CCD-Optimized Configuration on Initial Dataset |
| D | Weight Regularization Experiment |
| E | 10-Fold Cross-Validation Using CCD-Optimized Configuration with L2 Regularization on Initial Dataset |
| F | 10-Fold Cross-Validation Using the Pre-Trained CCD-Optimized Configuration with L2 Regularization on Expanded Dataset |

Table 32 – Mean (X-bar) Accuracy Summary Statistics

### Automated Copper Alloy Grain Size Evaluation Using a Deep-learning CNN

| | Runs | Run Time (minutes) | Training Accuracy | Validation Accuracy | Test Accuracy | Specificity | Sensitivity | FPR | Precision |
|---|---|---|---|---|---|---|---|---|---|
| A | 5 | 0.046 | 0.74% | 0.86% | 2.48% | 8.37% | 3.97% | 8.37% | 5.99% |
| B | 5 | 0.382 | 0.61% | 0.79% | 1.84% | 5.50% | 1.93% | 5.50% | 3.92% |
| C | 50 | 0.264 | 0.73% | 1.07% | 1.96% | 5.79% | 4.78% | 5.79% | 4.37% |
| D | 5 | 0.097 | 1.16% | 1.20% | 1.01% | 4.28% | 2.87% | 4.28% | 3.21% |
| E | 50 | 0.154 | 0.75% | 1.03% | 1.94% | 4.18% | 4.86% | 4.18% | 3.26% |
| F | 50 | 0.201 | 0.22% | 0.50% | 0.71% | 2.19% | 2.26% | 2.19% | 1.87% |

A   DSD Screening Experiment
B   CCD Optimization Experiment
C   10-Fold Cross-Validation Using CCD-Optimized Configuration on Initial Dataset
D   Weight Regularization Experiment
E   10-Fold Cross-Validation Using CCD-Optimized Configuration with L2 Regularization on Initial Dataset
F   10-Fold Cross-Validation Using the Pre-Trained CCD-Optimized Configuration with L2 Regularization on Expanded Dataset

Table 33 – Sample Standard Deviation (S) Accuracy Summary Statistics

| | Runs | Run Time (minutes) | Training Accuracy | Validation Accuracy | Test Accuracy | Specificity | Sensitivity | FPR | Precision |
|---|---|---|---|---|---|---|---|---|---|
| A | 5 | 0.0092 | 0.148% | 0.171% | 0.495% | 1.674% | 0.795% | 1.674% | 1.197% |
| B | 5 | 0.0764 | 0.121% | 0.158% | 0.367% | 1.100% | 0.385% | 1.100% | 0.784% |
| C | 50 | 0.0053 | 0.015% | 0.021% | 0.039% | 0.116% | 0.096% | 0.116% | 0.087% |
| D | 5 | 0.0194 | 0.231% | 0.240% | 0.201% | 0.857% | 0.573% | 0.857% | 0.642% |
| E | 50 | 0.0031 | 0.015% | 0.021% | 0.039% | 0.084% | 0.097% | 0.084% | 0.065% |
| F | 50 | 0.0040 | 0.004% | 0.010% | 0.014% | 0.044% | 0.045% | 0.044% | 0.037% |

A   DSD Screening Experiment
B   CCD Optimization Experiment
C   10-Fold Cross-Validation Using CCD-Optimized Configuration on Initial Dataset
D   Weight Regularization Experiment
E   10-Fold Cross-Validation Using CCD-Optimized Configuration with L2 Regularization on Initial Dataset
F   10Fold Cross-Validation Using the Pre-Trained CCD-Optimized Configuration with L2 Regularization on Expanded Dataset

Table 34 – Standard Error of the Mean (SE) Accuracy Summary Statistics

Although the reduction in stochastic variation could only be statistically proven for the specificity response of the *L2* regularization experiment in Section 5.5, Figure 53 clearly shows that the weight regularization (yellow) standard errors (SE) for the mean accuracies are reduced from those of the DSD (light blue) and CCD (orange) experiments; however, the reduction in variation is not as evident for the pre and post regularization cross-validations (grey and dark blue). This last point is shown graphically in Figure 54.

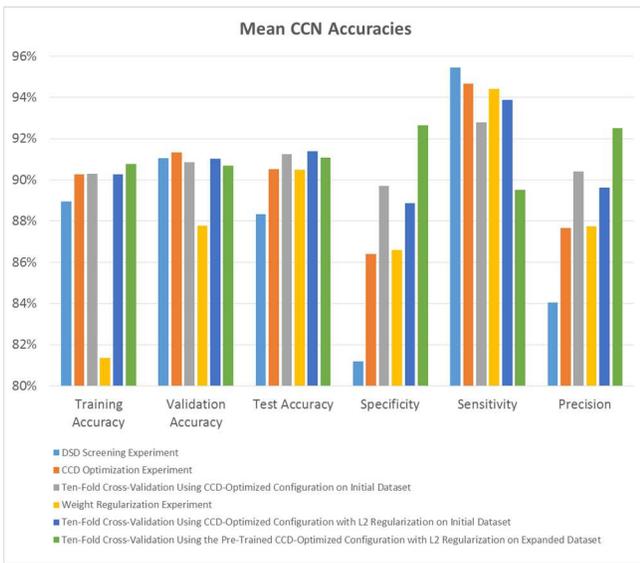

Figure 52 – Table 32 Mean CNN Accuracies

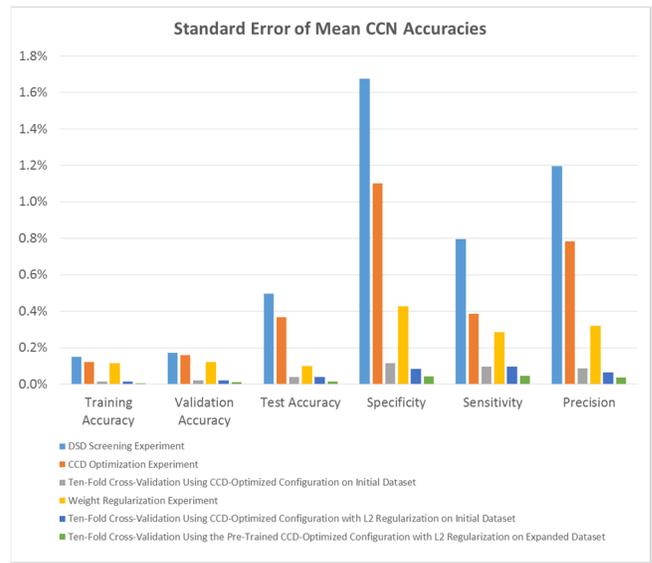

Figure 53 – Table 34 SE of Mean CNN Accuracies

**Automated Copper Alloy Grain Size Evaluation Using a Deep-learning CNN**

The test accuracy histograms from both the pre-regularization (C) and post-regularization (E) 10-fold cross-validations are superimposed in Figure 54, and show the same general shape in both the distribution and outliers.

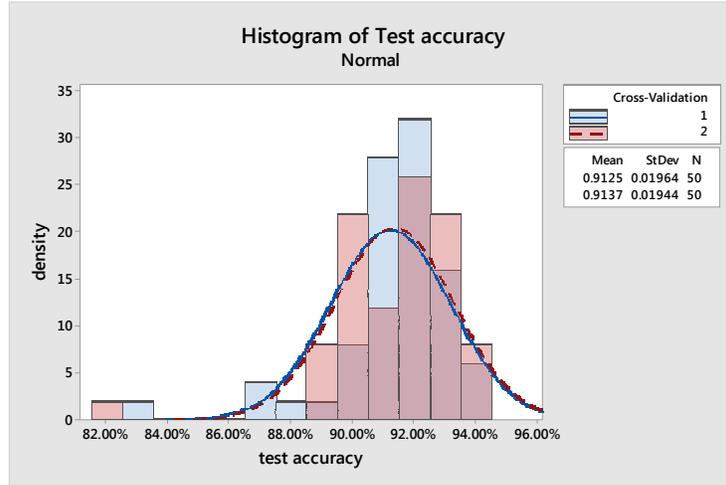

Figure 54 – Histograms for Pre- and Post-L2 Regularization 10-Fold Cross-Validations (C & E) on Initial Dataset

The horizontal axis of the histogram for the Figure 55 expanded dataset 10-fold cross-validation (F) is identically scaled to that of the initial dataset 10-fold cross-validation (C & E) in Figure 54.

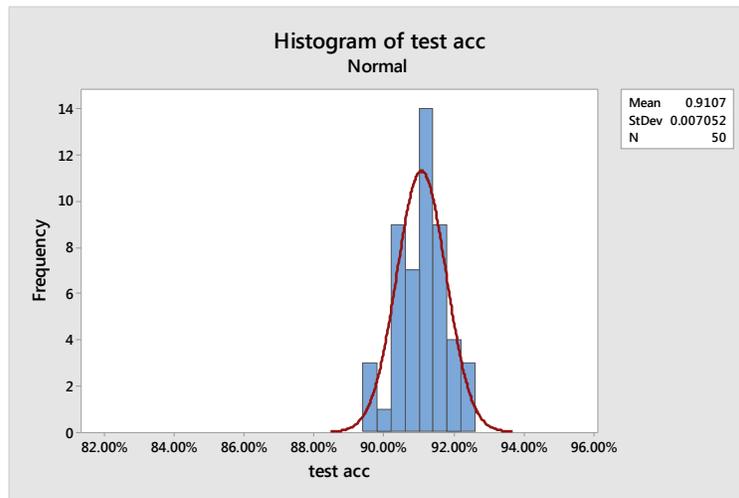

Figure 55 – Histogram for 10-Fold Cross-Validation (F) on Expanded Dataset

# 7. Conclusions

The effort to train a deep learning algorithm to recognize good and bad grain structures in Cu-alloy was successful.

- The CNN achieved a final accuracy of 91.1 ±0.7% test accuracy on the expanded dataset, which consisted of 12300 total sub-images images, with 8700 images for training, 2400 images for training validation and 1200 images for test verification.
- The efforts to balance the classification were somewhat successful, with the CNN slightly biased towards the recognition of bad grains over good grains, with a true negative rate specificity of 92.7 ±2.2% and a true positive rate sensitivity of 89.5 ±2.3%.
- Ensemble classification of the sub-images was 100%, because the presence of excessive bad grains within the test coupon image was color-coded by red tinting to indicate a bad sub-image, with more tinting applied with increasing probability of bad grain structure; this allows a technician to quickly recognize the overall state of the test coupon.

- The sub-image ensemble classification also provided a degree of explainability (i.e. XAI), since individual sub-images can be examined in detail to understand why the CNN produced the classification.

- The use of Design of Experiments (DoE) for simultaneously optimizing both the CNN configuration and training hyperparameters during CNN training and testing was successfully demonstrated, and this allowed the use of statistically-based metrics to determine if improvements in CNN performance actually occurred. This was especially critical because of the stochastic nature exhibited by the CNN when training. This is very important for industrial applications.

- The DoE method revealed that this particular CNN, with the Cu-alloy dataset, should be trained with settings that did not conform to best practices (e.g. large kernel constraints and no random dropouts).

- Although the experimentally determined *L2* CNN configuration was retained for 10-fold cross-validation on the expanded dataset, weight regularization was not entirely successful because no statistically significant lift in performance or reduction in run-to-run variation could be detected with the *L2* regularization settings applied.

- The configuration of the CNN and the structure of the Cu-alloy dataset may have contributed to the conclusions outlined in the previous two bullets (non-conformance to best practices for kernel constraints and random dropouts, as well as ineffective weight regularization). The simple CNN with a limited number of trainable parameters ($\sim$ 291K) naturally resists overfitting. This conclusion needs to be further explored.

- Because the DoE allows interactions between factors to be detected, as well as the quantification of experimental error (both lack-of-fit and pure error from reproducibility) in comparison to the factor main and interaction effects, the statistical DoE approach is much more efficient form of experimentation than either one-factor-at-a-time (1FAT) or 2 x 2 input grid-search approaches.

- Although not documented previously in this report, we found that in order to gain the highest resolution for the effects of experimental factors, the optimal number of training epochs when conducting DoEs was approximately 40% to 50% of the maximum number of epochs before the CNN began memorizing the training dataset images (i.e. overfitting the data); hence, this is why we used 100 epochs for the CCD experiments when 250 epochs were needed for both final verification checks and the 10-fold cross-validations. This conclusion should be further investigated.

- Automation of the CNN training and test validation using DoE experimental matrices facilitated a rapid development and optimization of the CNN hyperparameter configurations.

# 8. Additional Information

## 8.1. Funding, Export Compliance and IP

This project was implemented under Moog IR&D and was vetted by Moog Trade Compliance to ensure that no ITAR and/or EAR licensable data is contained in this document, and by Moog Intellectual Property to remove any Moog proprietary or confidential information or data.

**Automated Copper Alloy Grain Size Evaluation Using a Deep-learning CNN**

## 8.2. Glossary of Terms

| Term | Description |
| --- | --- |
| Adadelta | SGD adaptive moment estimation that mitigates aggressive reductions in learning rate |
| Adam | SGD with adaptive moment estimation |
| Adamax | Infinity-constrained SGD adaptive moment estimation |
| AI | Artificial Intelligence – a catch-all acronym for model-building learning algorithms that are currently very narrow in scope |
| AM | Additive Manufacturing |
| API | Application Programming Interface |
| CCC | Central-Composite Circumscribed – a type of CCD where the star-points are expanded outward |
| CCD | Central-Composite Design – a type of RSM DoE |
| CCF | Central-Composite Face – a type of CCD where the star-points are on the factor interaction planes |
| CCI | Central-Composite Inscribed – a type of CCD where the star-points are retracted inward |
| CNN | Convolutional Neural Network – a type of deep-learning AI that excels at recognizing patterns that exist within the spatial domain |
| Cu | Copper (the metal) |
| Deep-Learning | An AI with many hidden layers where feature maps are created automatically during training |
| DoE | Statistically-based Design of Experiments |
| DSD | Definitive Screening Design – a type of DoE |
| EAR | Export Administration Regulations |
| EVOP | Evolutionary Operations (an experimental method that follows the path of steepest gradient descent along a hypersurface of dimension n-1, where n = the number of experimental factor variables ) |
| Feature-Maps | The deep-learning-equivalent to a factor or variable in classic model building |
| FCC | Face-Centered Cubic – a form of CCD with star-points located on the two-factor interaction planes |
| FPR | False Positive Rate |
| GLM | General Linear Model: a statistical linear model for a continuous response variable using multivariate regression given continuous and/or categorical predictors |
| GPU | Graphics Processing Unit |
| iid | independent and identically distributed (a collection of random variables) |
| IR&D | Internal Research and Development |
| IP | Intellectual Property |
| ITAR | International Traffic in Arms Regulations |
| JPEG | Joint Photographic Experts Group – a standardized compressed graphic format (the image is reduced in file-size but undergoes information-loss when converted from TIFF to JPEG format) |
| L1 | An output weight regularization technique that uses lasso regression |
| L2 | An output weight regularization technique that uses ridge regression |
| Machine-learning | A type of AI where feature-maps are created for the training set through initial human intervention |
| MNIST | (Modified National Institute of Standards and Technology database) is a database of handwritten digits used for training and performance-benchmarking various image processing systems and is available in the public domain. |
| Nadam | Nesterov-accelerated SGD adaptive moment estimation |

**Automated Copper Alloy Grain Size Evaluation Using a Deep-learning CNN**

| Term | Description |
|------|-------------|
| NIST | National Institute of Standards and Technology, U.S. Department of Commerce |
| ReLU | Rectified Linear Unit – a neuron output activation function |
| RGB | Red Green Blue color model |
| RSM | Response Surface Method – a type of DoE |
| SELU | Scaled Exponential Linear Unit – a neuron output activation function |
| SGD | Stochastic Gradient Descent – an optimization function and method |
| SME | Subject Matter Expert |
| TIF (or TIFF) | Tagged Image File Format – a computer file format for storing raster graphics images |
| Treatment Combination | A test case for each combination of factors variable levels, which represents a single line in the experimental matrix (a holdover term from the agricultural origins of DoE where 'treatments' referred to soil and fertilizer configurations). |
| VIF | Variance Inflation Factor - a measure of multicollinearity between model factors |
| XAI | Explainable AI |

Table 35 – Glossary of Terms

## 8.3. Technical Information

- Robotic microscope: Nikon AZ100 (*NIS-Elements software*) with Prior Scientific motorized Shuttle Stage
- Sample preparation: Mager Scientific Saphir X-Change Fully Automatic Grinder/Polisher
- Hardware: Nvidia Titan Xp GPU
- Software: *Keras 2.0.6, Tensorflow 1.4.0, Scikit-learn 0.19.1, Numpy 1.13.3, Pandas 0.20.3, Matplotlib 2.0.2, Scikit-image 0.13.0, Python 3.6, Minitab 18.1, Image Slicer 0.1.0.*

## 8.4. Contact information

Dr. Paul Guerrier
CEng FIMechE | Engineering Manager
Moog Space and Defense Group
500 Jamison Rd. – Plant 20
East Aurora, New York 14052-0018
United States of America
716-652-2000

# References

[1]   T. M. Mitchell, The Discipline of Machine Learning, Pittsburgh: Carnegie Mellon University, 2006.

[2]   B. Zang, P. Jaiswal, R. Rai, P. Guerrier and G. Baggs, "Convolutional Neural Network-Based Inspection of Metal Additive Manufacturing Parts," *Rapid Prototyping Journal,* vol. 25, no. 3, pp. 530-540, 8 April 2019.

[3]   X. Li, X. Jia, Q. Yang and J. Lee, "Quality Analysis in Metal Additive Manufacturing with Deep Learning," *Journal of Intelligent Manufacturing,* no. Published online, February 25, 2020.

[4]   I. M. Kamal, R. A. Sutrisnowati, H. Bae and T. Lim, "Gear Classification for Defect Detection in Vision Inspection System Using Deep Convolution Neural Networks," *ICIC Express Letters,* vol. 9, no. 12, December 2018.

[5]   K. Anding, P. Kuritcyn and D. Garten, "Using Artificial Intelligence Strategies for Process-Related Automated Inspection in the Production Environment," *Journal of Physics, Conference Series,* vol. 772, no. 1, 2016.

[6]   L. Song, X. Li, Y. Yang, X. Zhu, Q. Guo and H. Yang, "Detection of Micro-Defects on Metal Screw Surfaces Based on Deep Convolutional Neural Networks," *Sensors,* vol. 18, no. 3709, 31 October 2018.

[7]   D. Masters and C. Luschi, "Revisiting Small Batch Training for Deep Neural Networks," arXiv.org, Cornell University, 2018.

[8]   S. N. Keskar, D. Mudigere, J. Nocedal, M. Smelyanskiy and P. T. P. Tang, "ON LARGE-BATCH TRAINING FOR DEEP LEARNING: GENERALIZATION GAP AND SHARP MINIMA," arXiv.org, Cornell University, 2017.

[9]   D. P. Kingma and J. L. Ba, "Adam: a Method for Stochastic Optimization," ArXiv, 2015.

[10] T. Dozat, "Incorporating Nesterov Momentum into Adam," ICLR, 2016.

[11] S. Ruder, "An Overview of Gradient Descent Optimization Algorithms," ArXiv, 2017.

[12] N. Srivastava, G. Hinton, A. Krizhevsky, I. Sutskever and R. Salakhutdinov, "Dropout: A Simple Way to Prevent Neural Networks from Overfitting," *Journal of Machine Learning Research (JMLR),* vol. 15, pp. 1929-1958, June 2014.

[13] I. Goodfellow, Y. Bengio and A. Courville, "Pooling," in *Deep Learning*, Cambridge MA, MIT Press, 2016, pp. 129, 195, 332-333, 342.

[14] G. E. P. Box and N. R. Draper, Evolutionary Operation: A Statistical Method for Process Improvement, Wiley, 1998.

[15] D. Sussillo and L. Abbot, "Random Walk Initialization for Training Very Deep Feedforward Networks," arXiv, Cornell University, 2014.

[16] G. Klambauer, T. Unterthiner, A. Mayr and S. Hochreiter, "Self-Normalizing Neural Networks," arXiv, Cornell University, 2017.

[17] R. Reed and R. J. Marksll, "Chaper 16.5: Weight Decay," in *Neural Smithing: Supervised Learning in Feedforward Artificial Neural Networks*, Cambridge MA, MIT Press, 1999, pp. 269-271.

[18] B. Jones and C. J. Nachtsheim, "A Class of Three-Level Designs for Definitive Screening in the Presence of Second-Order Effects," *Journal of Quality Technology,* vol. 43, no. 1, pp. 1-15, 2011.

[19] B. Jones and C. J. Nachtsheim, "Definitive Screening Designs with Added Two-Level Categorical Factors," *Journal of Quality Technology,* vol. 45, no. 2, pp. 121-129, 2013.

[20] G. E. P. Box and K. B. Wilson, "On the Experimental Attainment of Optimal Conditions," *Journal of the Royal Statistical Society,* vol. XIII, no. 1 Series B (Methodological), 1951.

[21] NIST National Institute of Standards and Technology, U.S. Department of Commerce, "5.3.3.6.1. Central Composite Designs (CCD)," 30 October 2013. [Online]. Available: https://www.itl.nist.gov/div898/handbook/pri/section3/pri3361.htm.

[22] G. S. Baggs, "Issue #2: Process Development and Control in Metal Additive Manufacturing," 2017. [Online]. Available: https://www.moog.com/news/blog-new/Issue2_ProcessDevelopmentAndControlInMetalAM.html.

[23] ASTM E1282-97, *Standard Test Methods for Determining Average Grain Size Using Semiautomatic and Automatic Image Analysis,* West Conshohocken, PA: ASTM International, 2010.

[24] R. E. Woods and R. C. Gonzalez, "Histogram Equalization," in *Digital Image Processing, Third Edition*, Prentice Hall, 2008, p. Chapter 3.

[25] M. Turek, "Explainable Artificial Intelligence (XAI)," DARPA (Defense Advanced Research Projects Agency), 26 November 2019. [Online]. Available: https://www.darpa.mil/program/explainable-artificial-intelligence. [Accessed 26 November 2019].

[26] J. Wang and L. Perez, "The Effectiveness of Data Augmentation in Image Classification using Deep Learning," arXiv, 2017.

[27] F. Chollet, "Building Powerful Image Classification Models Using Very Little Data," Keras, 5 June 2016. [Online]. Available: https://blog.keras.io/building-powerful-image-classification-models-using-very-little-data.html. [Accessed 25 October 2017].

[28] D. Falbel, J. Allaire and F. Chollet, "Tutorial: Overfitting and Underfitting," [Online]. Available: https://keras.rstudio.com/articles/tutorial_overfit_underfit.html. [Accessed 7 October 2019].

[29] J. Brownlee, "Deep Learning With Python," Jason Brownlee, 2016.

[30] T. Saito and M. Rehmsmeier, "Basic evaluation measures from the confusion matrix," 2018. [Online]. Available: https://classeval.wordpress.com/introduction/basic-evaluation-measures/.

[31] Keras Documentation, "Docs » Layers » Convolutional Layers," 2019. [Online]. Available: https://keras.io/layers/convolutional/.

[32] Minitab, LLC, "Coefficients table for Fit Regression Model," 2019. [Online]. Available: https://support.minitab.com/en-us/minitab/18/help-and-how-to/modeling-statistics/regression/how-to/fit-regression-model/interpret-the-results/all-statistics-and-graphs/coefficients-table/#vif.

[33] PennState Eberly College of Science, "1.1 - A Quick History of the Design of Experiments (DOE)," 2020. [Online]. Available: https://online.stat.psu.edu/stat503/lesson/1/1.1.

**Automated Copper Alloy Grain Size Evaluation Using a Deep-learning CNN**

# Appendix A: CNN Structure

The structure table of the CNN is provided below as output by the *Keras* API. The text following the CNN structure table provides the k-fold and run (treatment), epoch status and hyperparameter settings. In this case, the CNN was beginning the first epoch in the first fold of the expanded dataset 10-fold cross-validation of Section 5.6.

| Layer (type) | Output Shape | Param # |
|---|---|---|
| conv2d_1 (Conv2D) | (None, 32, 70, 70) | 896 |
| max_pooling2d_1 (MaxPooling2 | (None, 32, 35, 35) | 0 |
| dropout_1 (Dropout) | (None, 32, 35, 35) | 0 |
| conv2d_2 (Conv2D) | (None, 32, 35, 35) | 9248 |
| max_pooling2d_2 (MaxPooling2 | (None, 32, 17, 17) | 0 |
| dropout_2 (Dropout) | (None, 32, 17, 17) | 0 |
| conv2d_3 (Conv2D) | (None, 64, 17, 17) | 18496 |
| max_pooling2d_3 (MaxPooling2 | (None, 64, 8, 8) | 0 |
| dropout_3 (Dropout) | (None, 64, 8, 8) | 0 |
| flatten_1 (Flatten) | (None, 4096) | 0 |
| dense_1 (Dense) | (None, 64) | 262208 |
| dropout_4 (Dropout) | (None, 64) | 0 |
| dense_2 (Dense) | (None, 1) | 65 |
| activation_1 (Activation) | (None, 1) | 0 |

Total params: 290,913
Trainable params: 290,913
Non-trainable params: 0

None
K-fold: 1 Treatment: 1 , Batch: 240 , Kernel Constraint: 7.5 , Optimizer: Adam , DropCONV1: 0 , DropCONV2: 0 , DropCONV3: 0 , DropDNSE1: 0 , MaxPCONV1: 2 , MaxPCONV2: 2 , MaxPCONV3: 2 , FiltCONV1: 3 , FiltCONV2: 3 , FiltCONV3: 3 , Padding: same , StrideCONV1: 2 , Activator: Relu , C1_L1: 0 , C1_L2: 0.001 , C2_L1: 0, C2_L2: 0.0000001 , C3_L1: 0 , C3_L2: 0.0000001 , D1_L1: 0 , D1_L2: 0.0000001
Found 8700 images belonging to 2 classes.
Found 2400 images belonging to 2 classes.
Epoch 1/250

# Appendix B: k-Fold Fine Tuning

The 10-fold cross-validation on the expanded dataset was initially performed using fine tuning, where only the last convolutional feature extraction layer and the dense classification layers were allowed to train.

The initial dataset was expanded as described in Section 3.7.2, and the 10-fold cross-validation of Section 5.4.1 was repeated on the expanded dataset discussed in Section 3.7.2 using the CCD-optimized CNN. The $L2$ weight regularization settings described in Section 5.5 were also applied to the CNN.

The CNN successfully completed all 50 runs and achieved a test accuracy of 86.7 ±1.3% for the expanded 10-fold dataset, and required 70 hours and 59 minutes to complete. The CNN remained biased towards the classification of good grains over bad grains, with sensitivity exceeding specificity at 88.8 ±3.0% and 84.6 ±4.0% respectively.

The 10-fold cross-validation results appear in the GLM ANOVA of Table 36, Figure 56, Figure 57 and Figure 58, as well as Table 37. More variation came from the k-folds, and only the orange-highlighted *k-Fold* factor was statistically significant with a P value < 0.05 at 95% confidence. Figure 56 provides a graphical representation of the ANOVA results in Table 36, showing the variation between k-folds and the variation between test runs within the k-folds.

**Analysis of Variance (General Linear Model)**

**Test Accuracy**

| Source | DF | Adj. SS | Contribution | Adj. MS | F-Value | P-Value | VIF |
|--------|----|---------|--------------|---------|---------|---------|-----|
| k-Fold | 9 | 0.004582 | 56.81% | 0.000509 | 5.71 | 0.000 | 1.80 |
| Run | 4 | 0.000274 | 3.40% | 0.000068 | 0.77 | 0.553 | 1.60 |
| Error | 36 | 0.003211 | 39.81% | 0.000089 | | | |
| Total | 49 | 0.008066 | 100.00% | | | | |

**Model Summary**

| S | R-sq | R-sq(adj) | R-sq(pred) |
|---|------|-----------|------------|
| 0.009444 | 60.20% | 45.82% | 23.22% |

Table 36 – Expanded Dataset 10-Fold Cross-Validation (with Fine Tuning) Test Accuracy ANOVA

The horizontal axis of the histogram for the Figure 59 expanded dataset (F) 10-fold cross-validation using fine tuning is identically scaled to that of the initial dataset 10-fold cross-validation shown in Figure 54 and the expanded dataset 10-fold cross-validation shown in Figure 55.

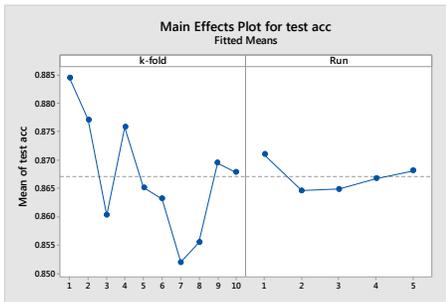

Figure 56 – Expanded Dataset 10-Fold Cross-Validation (with Fine Tuning) Mean Test Accuracies Partitioned between Fold and Run

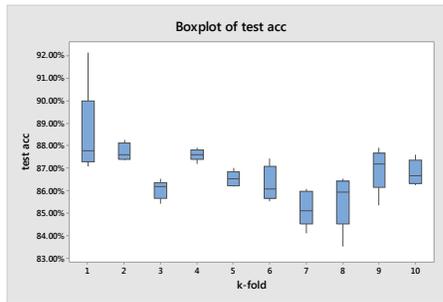

Figure 57 – Expanded Dataset 10-Fold Cross-Validation Boxplot (with Fine Tuning) of Test Accuracies for each Dataset Fold

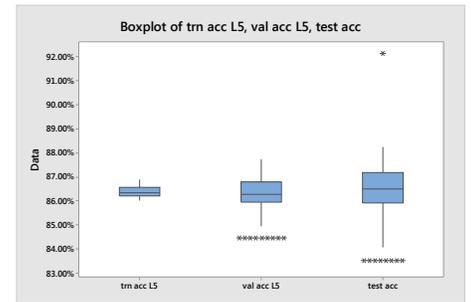

Figure 58 – Expanded Dataset 10-Fold Cross-Validation Boxplots (with Fine tuning) of Accuracies for Training, Validation and Test Data

**Automated Copper Alloy Grain Size Evaluation Using a Deep-learning CNN**

|  | Run Time (minutes) | Training Accuracy | Validation Accuracy | Test Accuracy | Specificity | Sensitivity | FPR | Precision |
|---|---|---|---|---|---|---|---|---|
| X-bar | 85.184 | 86.414% | 86.359% | 86.711% | 88.811% | 84.602% | 11.189% | 88.439% |
| S | 0.208 | 0.202% | 0.666% | 1.283% | 2.997% | 3.988% | 2.997% | 2.368% |

Table 37 – Expanded Dataset 10-Fold Cross-Validation (with Fine tuning) Test Accuracy Summary Statistics

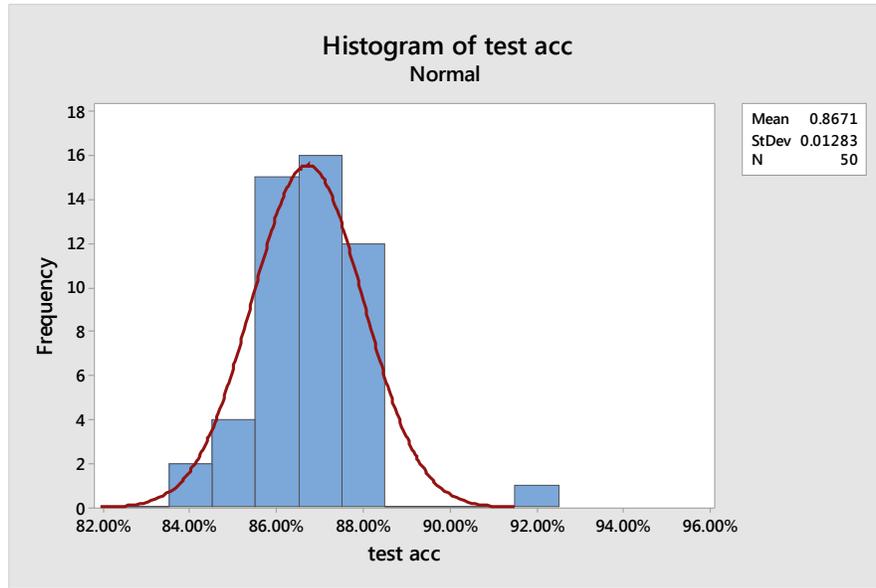

Figure 59 – Histogram for 10-Fold Cross-Validation (with Fine Tuning) on Expanded (F) Dataset